\def\year{2022}\relax
\documentclass[letterpaper]{article} 
\usepackage{aaai22}  
\usepackage{times}  
\usepackage{helvet}  
\usepackage{courier}  
\usepackage[hyphens]{url}  
\usepackage{graphicx} 
\usepackage{natbib}  
\usepackage{caption} 
\DeclareCaptionStyle{ruled}{labelfont=normalfont,labelsep=colon,strut=off} 
\usepackage{algorithm}
\usepackage{algorithmic}

\usepackage{amsmath}
\usepackage{amssymb}
\usepackage{mathtools}
\usepackage{siunitx}
\usepackage{array}
\usepackage{multirow}
\usepackage{booktabs}

\DeclareMathOperator{\Beta}{Beta}
\DeclareMathOperator{\Bernoulli}{Ber}
\DeclareMathOperator{\Categorical}{Cat}
\DeclareMathOperator{\Normal}{\mathcal{N}}
\DeclareMathOperator{\diag}{diag}
\DeclareMathOperator{\sigmoid}{sigmoid}
\DeclareMathOperator{\softmax}{softmax}
\DeclareMathOperator{\mean}{mean}
\DeclareMathOperator{\MI}{MI}
\DeclareMathOperator{\entropy}{H}

\DeclareMathOperator*{\argmax}{arg\,max}

\newcolumntype{C}[1]{>{\centering\let\newline\\\arraybackslash\hspace{0pt}}m{#1}}

\allowdisplaybreaks

\usepackage{newfloat}
\usepackage{listings}
\floatstyle{ruled}
\newfloat{listing}{tb}{lst}{}
\floatname{listing}{Listing}

\title{Unsupervised Learning of Compositional Scene Representations from \\Multiple Unspecified Viewpoints}
\author{
    Jinyang Yuan,
    Bin Li\protect\thanks{Corresponding author},
    Xiangyang Xue\footnotemark[1] \\
}
\affiliations{
    Shanghai Key Laboratory of Intelligent Information Processing, School of Computer Science, Fudan University \\
    \{yuanjinyang, libin, xyxue\}@fudan.edu.cn
}

\begin{document}

\maketitle

\begin{abstract}
Visual scenes are extremely rich in diversity, not only because there are infinite combinations of objects and background, but also because the observations of the same scene may vary greatly with the change of viewpoints. When observing a visual scene that contains multiple objects from multiple viewpoints, humans are able to perceive the scene in a compositional way from each viewpoint, while achieving the so-called ``object constancy'' across different viewpoints, even though the exact viewpoints are untold. This ability is essential for humans to identify the same object while moving and to learn from vision efficiently. It is intriguing to design models that have the similar ability. In this paper, we consider a novel problem of learning compositional scene representations from multiple unspecified viewpoints without using any supervision, and propose a deep generative model which separates latent representations into a viewpoint-independent part and a viewpoint-dependent part to solve this problem. To infer latent representations, the information contained in different viewpoints is iteratively integrated by neural networks. Experiments on several specifically designed synthetic datasets have shown that the proposed method is able to effectively learn from multiple unspecified viewpoints.
\end{abstract}

\section{Introduction}

Vision is an important way for humans to acquire knowledge about the world. Due to the diverse combinations of objects and background that constitute visual scenes, it is hard to model the whole scene directly. In the process of learning from the world, humans are able to develop the concept of object \cite{Johnson2010HowIL}, and is thus capable of perceiving visual scenes compositionally, which in turn leads to more efficient learning compared with perceiving the entire scene as a whole \cite{Fodor1988ConnectionismAC}. Compositionality is one of the fundamental ingredients for building artificial intelligence systems that learn efficiently and effectively like humans \cite{Lake2017BuildingMT}. Therefore, instead of learning a single representation for the entire visual scene, it is desirable to build compositional scene representation models which learn \emph{object-centric representations} (i.e., learn separate representations for different objects and background), so that the combinational property can be better captured.

\begin{figure}[t]
\centering
\includegraphics[width=0.96\columnwidth]{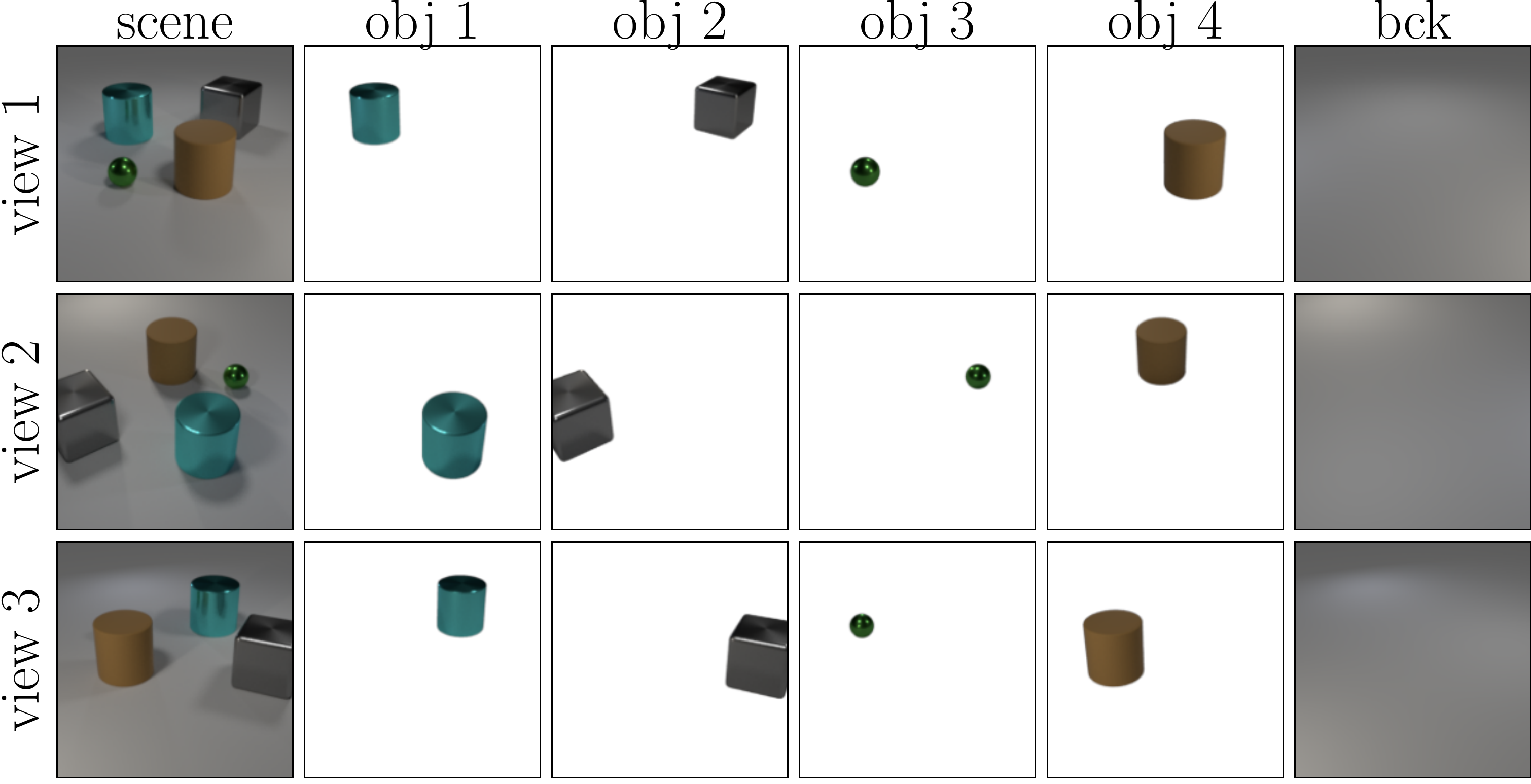}
\caption{Humans are able to perceive visual scenes compositionally, while maintaining object constancy across different viewpoints (indexes of objects are arbitrarily chosen).}
\label{fig:intro}
\end{figure}

In addition, humans have the ability to achieve the so-called ``object constancy'' in visual perception, i.e., recognizing the same object from different viewpoints \cite{Turnbull1997TheNO}, possibly because of the mechanisms such as performing mental rotation \cite{Shepard1971MentalRO} or representing objects in a viewpoint-independent way \cite{Marr1982Vision}. When observing a multi-object scene from multiple viewpoints, humans are able to separate different objects from one another, and identify the same one from different viewpoints. As shown in Figure \ref{fig:intro}, given three images of the same visual scene observed from different viewpoints (column 1), humans are capable of decomposing each image into \emph{complete} objects (columns 2-5) and background (column 6) that are \emph{consistent} across viewpoints, even though the viewpoints are \emph{unknown}, the poses of the same object may be significantly \emph{different} across viewpoints, and some objects may be partially (object 2 in viewpoint 1) or even completely (object 3 in viewpoint 3) \emph{occluded}. Observing visual scenes from multiple viewpoints gives humans a better understanding of the scenes, and it is intriguing to design compositional scene representation methods that are able to achieve object constancy and effectively learn from multiple viewpoints like humans.

In recent years, a variety of deep generative models have been proposed to learn compositional representations without object-level supervision. Most methods, such as AIR \cite{Eslami2016AttendIR}, N-EM \cite{Greff2017NeuralEM}, MONet \cite{Burgess2019MONetUS}, IODINE \cite{Greff2019MultiObjectRL}, and Slot Attention \cite{Locatello2020ObjectCentricLW}, however, are unsupervised methods that learn from only a \emph{single} viewpoint. Only few methods, including MulMON \cite{Li2020LearningOR} and ROOTS \cite{Chen2020ObjectCentricRA}, have considered the problem of learning from multiple viewpoints. These methods assume that the viewpoint annotations (under a certain global coordinate system) are given, and aim to learn viewpoint-independent object-centric representations \emph{conditioned on} these annotations. Viewpoint annotations play fundamental roles in the initialization and updates of object-centric representations in MulMON, and in the computations of perspective projections in ROOTS. Therefore, without nontrivial modifications, existing methods \emph{cannot} be applied to the novel problem of learning compositional scene representations from multiple unspecified viewpoints \emph{without} any supervision.

The problem setting considered in this paper is very challenging, as the object-centric representations that are shared across viewpoints and the viewpoint representations that are shared across objects both need to be learned. More specifically, there are two major reasons. \emph{Firstly}, the object constancy needs to be achieved \emph{without the guidance} of viewpoint annotations, which are the only variable among images observed from different viewpoints and can be exploited to reduce the difficulty of learning the common factors. \emph{Secondly}, the representations of images need to be disentangled into object-centric representations and viewpoint representations, even though there are \emph{infinitely many} possible solutions, e.g., due to the change of global coordinate system.

In this paper, we propose a deep generative model called \textbf{O}bject-\textbf{C}entric \textbf{L}earning with \textbf{O}bject \textbf{C}onstancy (OCLOC) to learn object-centric representations from multiple viewpoints \emph{without any supervision} (including viewpoint annotations), under the assumptions that 1) objects in the visual scenes are \emph{static}, and 2) different visual scenes may be observed from \emph{different} sets of \emph{unordered} viewpoints.
The proposed method models viewpoint-independent attributes of objects/background (e.g., 3D shapes and appearances in the global coordinate system) and viewpoints with separate latent variables, and adopts an amortized variational inference method that iteratively updates parameters of the approximated posteriors by integrating information of different viewpoints with inference neural networks.

To the best of the authors' knowledge, no existing object-centric learning method can learn from multiple unspecified viewpoints without viewpoint annotations. Thus, the proposed OCLOC cannot be directly compared with existing ones in the considered problem setting. Experiments on several specifically designed synthetic datasets have shown that OCLOC can effectively learn from multiple unspecific viewpoints without supervision, and \emph{competes with} or \emph{slightly outperforms} a state-of-the-art method that uses viewpoint annotations in the learning. Under an extreme condition that visual scenes are observed from one viewpoint, the proposed OCLOC is also comparable with the state-of-the-arts.

\section{Related Work}

Object-centric representations are compositional scene representations that treat object or background as the basic entity of the visual scene and represent different objects or background separately. In recent years, various methods have been proposed to learn object-centric representations in an unsupervised manner, or using only scene-level annotations. Based on whether learning from multiple viewpoints and whether considering the movements of objects, these methods can be roughly divided into three categories.

\textbf{Single-Viewpoint Static Scenes:}
CST-VAE \cite{Huang2015EfficientII}, AIR \cite{Eslami2016AttendIR}, and MONet \cite{Burgess2019MONetUS} extract the representation of each object sequentially based on the attention mechanism. GMIOO \cite{Yuan2019GenerativeMO} initializes the representation of each object sequentially and iteratively updates the representations, both with attentions on objects. SPAIR \cite{Crawford2019SpatiallyIU} and SPACE \cite{Lin2020SPACEUO} generate object proposals with convolutional neural networks and are applicable to large visual scenes containing a relatively large number of objects. N-EM \cite{Greff2017NeuralEM}, LDP \cite{Yuan2019SpatialMM}, IODINE \cite{Greff2019MultiObjectRL}, Slot Attention \cite{Locatello2020ObjectCentricLW}, and EfficientMORL \cite{Equivariance2021EfficientIA} first initialize representations of all the objects, and then apply some kind of competitions among objects to iteratively update the representations in parallel. GENESIS \cite{Engelcke2020GENESISGS} and GNM \cite{Jiang2020GenerativeNM} consider the structure of visual scene in the generative models in order to generate more coherent samples. ADI \cite{Yuan2021KnowledgeGuidedOD} considers the acquisition and utilization of knowledge. These methods provide mechanisms to separate objects, and form the foundations of learning object-centric representations with the existences of object motions or from multiple viewpoints.

\textbf{Multi-Viewpoint Static Scenes:}
MulMON \cite{Li2020LearningOR} and ROOTS \cite{Chen2020ObjectCentricRA} are two methods proposed to learn from static scenes from multiple viewpoints. MulMON extends the iterative amortized inference \cite{Marino2018IterativeAI} used in IODINE \cite{Greff2019MultiObjectRL} to sequences of images observed from different viewpoints. Object-centric representations are first initialized based on the first pair of image and \emph{viewpoint annotation}, and then iteratively refined by processing the rest pairs of data one by one. At each iteration, the previously estimated posteriors of latent variables are used as the current object-wise priors in order to guide the inference. ROOTS adopts the idea of using grid cells like SPAIR \cite{Crawford2019SpatiallyIU} and SPACE \cite{Lin2020SPACEUO}, and generates object proposals in a bounded 3D region. The 3D center position of each object proposal is estimated and projected into different images with transformations that are computed based on the \emph{annotated viewpoints}. After extracting crops of images corresponding to each object proposal, a type of GQN \cite{Eslami2018NeuralSR} is applied to infer object-centric representations. As with our problem setting, different visual scenes are not assumed to be observed from the same set of viewpoints. However, because both methods heavily rely on the viewpoint annotations, they cannot be trivially applied to the fully-unsupervised scenario that the viewpoint annotations are unknown.

\textbf{Dynamic Scenes:}
Inspired by the methods proposed for learning from single-viewpoint static scenes, several methods, such as Relational N-EM \cite{Steenkiste2018RelationalNE}, SQAIR \cite{Kosiorek2018SequentialAI}, R-SQAIR \cite{Stanic2019RSQAIRRS}, TBA \cite{He2019TrackingBA}, SILOT \cite{Crawford2020ExploitingSI}, SCALOR \cite{Jiang2020SCALORGW}, OP3 \cite{Veerapaneni2020EntityAI}, and PROVIDE \cite{Zablotskaia2021PROVIDE}, have been proposed for learning from video sequences. The difficulties of this problem setting include modeling object motions and relationships, as well as maintaining the identities of objects even if objects disappear and reappear after full occlusion \cite{Weis2020BenchmarkingUO}. Although these methods are able to identify the same object across adjacent frames, they cannot be directly applied to the problem setting considered in this paper for two major reasons: 1) images observed from different viewpoints are assumed to be unordered, and the positions of the same object may differ significantly in different images; and 2) viewpoints are shared among objects in the same visual scene, while object motions in videos do not have such a property.

\section{Generative Modeling}

Visual scenes are assumed to be independent and identically distributed. For simplicity, the index of visual scene is omitted, and the procedure to generate images of a single visual scene is described. Let $M$ denote the number of images observed from different viewpoints (\emph{may vary} in different visual scenes), $N$ and $C$ denote the respective numbers of pixels and channels in each image, and $K$ denote the maximum number of objects that may appear in the visual scene. The image of the $m$th viewpoint $\boldsymbol{x}_{m} \in \mathbb{R}^{N \times C}$ is assumed to be generated via a pixel-wise weighted summation of $K + 1$ layers, with $K$ layers ($1 \!\leq\! k \!\leq\! K$) describing the objects and $1$ layer ($k \!=\! 0$) describing the background. The pixel-wise weights $\boldsymbol{s}_{m,0:K} \!\in\! [0, 1]^{(K + 1) \times N}$ as well as the images of layers $\boldsymbol{a}_{m,0:K} \!\in\! \mathbb{R}^{(K + 1) \times N \times C}$ are computed based on latent variables. In the following, we first describe the latent variables and the likelihood function, and then express the generative model in the mathematical form.

\subsection{Viewpoint-Independent Latent Variables}

Viewpoint-independent latent variables are the ones that are shared across different viewpoints, and are introduced in the generative model to achieve object constancy. These latent variables include $\boldsymbol{z}^{\text{attr}}$, $\boldsymbol{\rho}$, and $\boldsymbol{z}^{\text{prs}}$.
\begin{itemize}
	\item $\boldsymbol{z}^{\text{attr}}_{0:K}$ characterize the viewpoint-independent attributes of objects ($1 \!\leq\! k \!\leq\! K$) and background ($k \!=\! 0$). These attributes include the 3D shapes and appearances of objects and background in an automatically chosen global coordinate system. The dimensionalities of all the $\boldsymbol{z}^{\text{attr}}_{k}$ with $1 \!\leq\! k \!\leq\! K$ are identical, and are in general different from the dimensionality of $\boldsymbol{z}^{\text{attr}}_{0}$. For notational simplicity, this difference is not reflected in the expressions of the generative model. The priors of all the $\boldsymbol{z}^{\text{attr}}_{k}$ with $0 \!\leq\! k \!\leq\! K$ are standard normal distributions.
	\item $\boldsymbol{\rho}_{1:K}$ and $\boldsymbol{z}^{\text{prs}}_{1:K}$ are used to model the number of objects in the visual scene, considering that different visual scenes may contain different numbers of objects. The binary latent variable $\boldsymbol{z}^{\text{prs}}_{k} \in \{0, 1\}$ indicates whether the $k$th object is included in the visual scene (i.e., the number of objects is $\sum_{k=1}^{K}{\boldsymbol{z}^{\text{prs}}_{k}}$), and is sampled from a Bernoulli distribution with the latent variable $\boldsymbol{\rho}_{k}$ as its parameter. The priors of all the $\boldsymbol{\rho}_{k}$ with $1 \!\leq\! k \!\leq\! K$ are beta distributions parameterized by hyperparameters $\alpha$ and $K$.
\end{itemize}

\subsection{Viewpoint-Dependent Latent Variables}

Viewpoint-dependent latent variables may vary as the viewpoint changes. These latent variables include $\boldsymbol{z}^{\text{view}}$ and $\boldsymbol{z}^{\text{shp}}$.
\begin{itemize}
	\item $\boldsymbol{z}_{m}^{\text{view}}$ determines the viewpoint (in an automatically chosen global coordinate system) of the $m$th image, and is drawn from a standard normal prior distribution.
	\item $\boldsymbol{z}_{m,1:K,1:N}^{\text{shp}} \in \{0, 1\}^{K \times N}$ consist of binary latent variables that indicate the complete shapes of objects in the image coordinate system determined by the $m$th viewpoint. Each element of $\boldsymbol{z}_{m,1:K,1:N}^{\text{shp}}$ is sampled independently from a Bernoulli distribution, whose parameter is computed by transforming latent variables $\boldsymbol{z}_{m}^{\text{view}}$ and $\boldsymbol{z}_{k}^{\text{attr}}$ ($1 \!\leq\! k \!\leq\! K$) with a neural network $f_{\text{shp}}$ that captures the spatial dependencies among pixels. The sigmoid activation function in the last layer of $f_{\text{shp}}$ is explicitly expressed to clarify the output range of the neural network.
\end{itemize}

\subsection{Likelihood Function}

All the pixels of the images $\boldsymbol{x}_{1:M,1:N}$ are assumed to be conditional independent of each other given all the latent variables $\boldsymbol{\Omega}$, and the likelihood function $p(\boldsymbol{x}|\boldsymbol{\Omega})$ is assumed to be factorized as the product of several normal distributions with varying mean vectors and constant covariance matrices, i.e., $\prod_{m=1}^{M}{\prod_{n=1}^{N}{\mathcal{N}(\sum_{k=0}^{K}{s_{m,k,n} \, \boldsymbol{a}_{m,k,n}}, \, \sigma_{\text{x}}^2 \boldsymbol{I})}}$. To compute the mean vectors, intermediate variables $\boldsymbol{o}$, $\boldsymbol{s}$, and $\boldsymbol{a}$ need to be computed by transforming the sampled latent variables with deterministic functions.
\begin{itemize}
	\item $\boldsymbol{o}_{m,1:K}$ characterize the depth ordering of objects in the image observed from the $m$th viewpoint. If multiple objects overlap, the object with the largest value of $o_{m,k}$ is assumed to occlude the others in a soft and differentiable way. To compute $o_{m,k}$, latent variables $\boldsymbol{z}_{m}^{\text{view}}$ and $\boldsymbol{z}_{k}^{\text{attr}}$ are first transformed by a neural network $f_{\text{ord}}$, and then the exponential function is applied to the output of $f_{\text{ord}}$ divided by $\lambda$. The exponential function ensures that the value of $o_{m,k}$ is greater than $0$, and the hyperparameter $\lambda$ controls the softness of object occlusions.
	\item $\boldsymbol{s}_{m,0:K,1:N}$ indicate the perceived shapes of objects ($1 \!\leq\! k \!\leq\! K$) and background ($k \!=\! 0$) in the $m$th image, and satisfy the constraints that $(\forall m, k, n) \, 0 \!\leq\! s_{m,k,n} \!\leq\! 1$ and $(\forall m, n) \sum_{k=0}^{K}{s_{m,k,n}} \!=\! 1$. These latent variables are computed based on $\boldsymbol{z}_{1:K}^{\text{prs}}$, $\boldsymbol{z}_{m,1:K,1:N}^{\text{shp}}$, and $\boldsymbol{o}_{m,1:K}$. Because $\boldsymbol{z}^{\text{prs}}$ and $\boldsymbol{z}^{\text{shp}}$ are binary variables, the perceived shape $\boldsymbol{s}_{m,0,1:N}$ of background is also binary, and equals $1$ at the pixels that are not covered by any object. The computation of perceived shapes $\boldsymbol{s}_{m,1:K,n}$ of objects at each pixel can be interpreted as a masked softmax operation that only considers the objects covering that pixel. As the hyperparameter $\lambda$ in the computation of $\boldsymbol{o}$ approaches $0$, the perceived shapes $\boldsymbol{s}_{m,0:K,n}$ of all the objects and background at each pixel approach a one-hot vector.
	\item $\boldsymbol{a}_{m,0:K,1:N}$ contain information about the complete appearances of objects ($1 \!\leq\! k \!\leq\! K$) and the background image ($k \!=\! 0$) in the $m$th image, and are computed by transforming latent variables $\boldsymbol{z}_{m}^{\text{view}}$ and $\boldsymbol{z}_{k}^{\text{attr}}$ with neural networks $f_{\text{bck}}$ (for $ k \!=\! 0$) and $f_{\text{apc}}$ (for $1 \!\leq\! k \!\leq\! K$). Appearances of objects and the background image are computed differently because the dimensionality of $\boldsymbol{z}_{0}^{\text{attr}}$ is in general different from $\boldsymbol{z}_{k}^{\text{attr}}$ with $1 \!\leq\! k \!\leq\! K$.
\end{itemize}

\subsection{Generative Model}

The mathematical expressions of the generative model are
\begin{align*}
	\boldsymbol{z}_{m}^{\text{view}} & \sim \Normal\big(\boldsymbol{0}, \boldsymbol{I}\big); \qquad\qquad\qquad\mkern4mu \boldsymbol{z}_{k}^{\text{attr}} \sim \Normal\big(\boldsymbol{0}, \boldsymbol{I}\big) \\
	\rho_{k} & \sim \Beta\big(\alpha / K, 1\big); \qquad\qquad z_{k}^{\text{prs}} \sim \Bernoulli\big(\rho_{k}\big) \\
	z_{m,k,n}^{\text{shp}} & \sim \Bernoulli\big(\sigmoid(f_{\text{shp}}(\boldsymbol{z}_{m}^{\text{view}}, \boldsymbol{z}_{k}^{\text{attr}})_n)\big) \\
	o_{m,k} & = \exp\big(f_{\text{ord}}(\boldsymbol{z}_{m}^{\text{view}}, \boldsymbol{z}_{k}^{\text{attr}}) / \lambda\big) \\
	s_{m,k,n} & =
	\begin{dcases}
		\prod\nolimits_{k'=1}^{K}(1 - z_{k'}^{\text{prs}} \, z_{m,k',n}^{\text{shp}}), & k = 0 \\
		\frac{(1 - s_{m,0,n}) \, z_{k}^{\text{prs}} \, z_{m,k,n}^{\text{shp}} \, o_{m,k}}{\sum_{k'=1}^{K}{z_{k'}^{\text{prs}} \, z_{m,k',n}^{\text{shp}} \, o_{m,k'}}}, & 1 \leq k \leq K
	\end{dcases} \\
	\boldsymbol{a}_{m,k,n} & =
	\begin{cases}
		f_{\text{bck}}\big(\boldsymbol{z}_{m}^{\text{view}}, \boldsymbol{z}_{k}^{\text{attr}}\big)_{n}, & \qquad\qquad\quad k = 0 \\
		f_{\text{apc}}\big(\boldsymbol{z}_{m}^{\text{view}}, \boldsymbol{z}_{k}^{\text{attr}}\big)_{n}, & \qquad\qquad\quad 1 \leq k \leq K
	\end{cases} \\
	\boldsymbol{x}_{m,n} & \sim \mathcal{N}\Big(\sum\nolimits_{k=0}^{K}{s_{m,k,n} \, \boldsymbol{a}_{m,k,n}}, \, \sigma_{\text{x}}^2 \boldsymbol{I}\Big)
\end{align*}
In the above expressions, some of the ranges of indexes $m$ ($1 \!\leq\! m \!\leq\! M$), $n$ ($1 \!\leq\! n \!\leq\! N$), and $k$ ($0 \!\leq\! k \!\leq\! K$ for $\boldsymbol{z}^{\text{attr}}$, and $1 \!\leq\! k \!\leq\! K$ for $\boldsymbol{\rho}$, $\boldsymbol{z}^{\text{prs}}$, $\boldsymbol{z}^{\text{shp}}$, $\boldsymbol{o}$) are omitted for notational simplicity. $\alpha$, $\lambda$, and $\sigma_{\text{x}}$ are tunable hyperparameters. Let $\boldsymbol{\Omega} = \{$$\boldsymbol{z}^{\text{view}}$, $\boldsymbol{z}^{\text{attr}}$, $\boldsymbol{\rho}$, $\boldsymbol{z}^{\text{prs}}$, $\boldsymbol{z}^{\text{shp}}$$\}$ be the collection of all latent variables. The joint probability of $\boldsymbol{x}$ and $\boldsymbol{\Omega}$ is
\begin{align}
	\label{equ:generative}
	& p(\boldsymbol{x}, \boldsymbol{\Omega}) = \prod\nolimits_{k=0}^{K}{\!p(\boldsymbol{z}_{k}^{\text{attr}})} \prod\nolimits_{k=1}^{K}{\!p(\rho_{k}) p(z_{k}^{\text{prs}}|\rho_{k})} \\
	& \qquad \prod\nolimits_{m=1}^{M}{\!p(\boldsymbol{z}_{m}^{\text{view}}) \prod\nolimits_{k=1}^{K}{\!\prod\nolimits_{n=1}^{N}{\!p(z_{m,k,n}^{\text{shp}}|\boldsymbol{z}_{m}^{\text{view}}, \boldsymbol{z}_{k}^{\text{attr}})}}} \nonumber\\
	& \qquad \prod\nolimits_{m=1}^{M}{\!\prod\nolimits_{n=1}^{N}{\!p(\boldsymbol{x}_{m,n}|\boldsymbol{z}_{m}^{\text{view}}, \boldsymbol{z}_{0:K}^{\text{attr}}, \boldsymbol{z}_{1:K}^{\text{prs}}, \boldsymbol{z}_{m,1:K,n}^{\text{shp}})}} \nonumber
\end{align}

\section{Inference and Learning}

The exact posterior distribution of latent variables $p(\boldsymbol{\Omega}|\boldsymbol{x})$ is intractable to compute. Therefore, we adopt amortized variational inference, which approximates the complex posterior distribution with a tractable variational distribution $q(\boldsymbol{\Omega}|\boldsymbol{x})$, and apply neural networks to transform the images $\boldsymbol{x}$ into parameters of the variational distribution. The neural networks $f_{\text{shp}}$, $f_{\text{ord}}$, $f_{\text{bck}}$, and $f_{\text{apc}}$ in the generative model, as well as the inference networks, are jointly optimized with the goal of maximizing the evidence lower bound (ELBO). Details of the inference and learning are described below.

\begin{algorithm}[tb]
	\caption{Inference of latent variables}
	\label{alg:infer}
	\textbf{Input}: Images of $M$ viewpoints $\boldsymbol{x}_{1:M}$ \\
	\textbf{Output}: Parameters of $q(\boldsymbol{\Omega}|\boldsymbol{x})$
	\begin{algorithmic}[1] 
		\STATE // Extract features and initialize intermediate variables
		\STATE $\boldsymbol{y}_{m}^{\text{feat}} \gets g_{\text{feat}}(\boldsymbol{x}_m), \quad \forall \, 1 \!\leq\! m \!\leq\! M$
		\STATE $\boldsymbol{y}_{m}^{\text{view}} \sim \mathcal{N}(\hat{\boldsymbol{\mu}}^{\text{view}}, \diag(\hat{\boldsymbol{\sigma}}^{\text{view}})), \quad \forall \, 1 \!\leq\! m \!\leq\! M$
		\STATE $\boldsymbol{y}_{k}^{\text{attr}} \sim \mathcal{N}(\hat{\boldsymbol{\mu}}^{\text{attr}}, \diag(\hat{\boldsymbol{\sigma}}^{\text{attr}})), \quad \forall \, 0 \!\leq\! k \!\leq\! K$
		\STATE // Update intermediate variables $\boldsymbol{y}_{1:M}^{\text{view}}$ and $\boldsymbol{y}_{0:K}^{\text{attr}}$
		\FOR[$\forall \, 1 \mkern-4mu\leq\mkern-4mu m \mkern-4mu\leq\mkern-4mu M, 0 \mkern-4mu\leq\mkern-4mu k \mkern-4mu\leq\mkern-4mu K$ in the loop]{$t \gets 1$ to $T$}
			\STATE $\boldsymbol{y}_{m,k}^{\text{full}} \!\gets [\boldsymbol{y}_{m}^{\text{view}}, \boldsymbol{y}_{k}^{\text{attr}}]$
			\STATE $\boldsymbol{a}_{m,k} \!\gets \softmax_K\!\big(g_{\text{key}}(\boldsymbol{y}_{m}^{\text{feat}}) g_{\text{qry}}(\boldsymbol{y}_{m,0:K}^{\text{full}}) / \!\sqrt{\!D_{\text{key}}}\big)$
			\STATE $\boldsymbol{u}_{m,k} \!\gets \sum_{N}{\softmax_N(\log{\boldsymbol{a}_{m,k}}) \, g_{\text{val}}(\boldsymbol{y}_{m}^{\text{feat}})}$
			\STATE $[\boldsymbol{v}_{1:M,0:K}^{\text{view}}, \boldsymbol{v}_{1:M,0:K}^{\text{attr}}] \gets g_{\text{upd}}(\boldsymbol{y}_{1:M,0:K}^{\text{full}}, \boldsymbol{u}_{1:M,0:K})$
			\STATE $\boldsymbol{y}_{m}^{\text{view}} \gets \mean_{K}(\boldsymbol{v}_{m,k}^{\text{view}})$
			\STATE $\boldsymbol{y}_{k}^{\text{attr}} \gets \mean_{M}(\boldsymbol{v}_{m,k}^{\text{attr}})$
		\ENDFOR
		\STATE // Sample the background index and rearrange $\boldsymbol{y}_{0:K}^{\text{attr}}$
		\STATE $\pi_{k} = \softmax_{K}\!\big(g_{\text{sel}}(\boldsymbol{y}_{0:K}^{\text{attr}})\big), \quad \forall \, 0 \!\leq\! k \!\leq\! K$
		\STATE $k^* \sim \Categorical(\pi_{0}, \dots, \pi_{K}); \quad \boldsymbol{y}_{0:K}^{\text{attr}} \gets [\boldsymbol{y}_{k^*}^{\text{attr}}, \boldsymbol{y}_{0:K \setminus k^*}^{\text{attr}}]$
		\STATE // Convert $\boldsymbol{y}_{1:M}^{\text{view}}$ and $\boldsymbol{y}_{0:K}^{\text{attr}}$ to parameters of $q(\boldsymbol{\Omega}|\boldsymbol{x})$
		\STATE $\boldsymbol{\mu}_{0}^{\text{attr}}, \boldsymbol{\sigma}_{0}^{\text{attr}} \gets g_{\text{bck}}(\boldsymbol{y}_{0}^{\text{attr}})$
		\STATE $\boldsymbol{\mu}_{k}^{\text{attr}}, \boldsymbol{\sigma}_{k}^{\text{attr}}, \boldsymbol{\tau}_{k}, \kappa_{k} \gets g_{\text{obj}}(\boldsymbol{y}_{k}^{\text{attr}}), \quad \forall \, 1 \!\leq\! k \!\leq\! K$
		\STATE $\boldsymbol{\mu}_{m}^{\text{view}}, \boldsymbol{\sigma}_{m}^{\text{view}} \gets g_{\text{view}}(\boldsymbol{y}_{m}^{\text{view}}), \quad \forall \, 1 \!\leq\! m \!\leq\! M$
		\STATE \textbf{return} $\boldsymbol{\mu}_{0:K}^{\text{attr}}, \boldsymbol{\sigma}_{0:K}^{\text{attr}}, \boldsymbol{\tau}_{1:K}, \boldsymbol{\kappa}_{1:K}, \boldsymbol{\mu}_{1:M}^{\text{view}}, \boldsymbol{\sigma}_{1:M}^{\text{view}}$
	\end{algorithmic}
\end{algorithm}

\subsection{Inference of Latent Variables}

The variational distribution $q(\boldsymbol{\Omega}|\boldsymbol{x})$ is factorized as
\begin{align}
	\label{equ:inference}
	& q(\boldsymbol{\Omega}|\boldsymbol{x}) = \prod\nolimits_{k}{\!q(\boldsymbol{z}_{k}^{\text{attr}}|\boldsymbol{x})} \prod\nolimits_{k}{\!q(\rho_{k}|\boldsymbol{x}) q(z_{k}^{\text{prs}}|\boldsymbol{x})} \\
	& \qquad \prod\nolimits_{m}{\!q(\boldsymbol{z}_{m}^{\text{view}}|\boldsymbol{x}) \prod\nolimits_{k}{\!\prod\nolimits_{n}{\!q(z_{m,k,n}^{\text{shp}}|\boldsymbol{z}_{m}^{\text{view}}, \boldsymbol{z}_{k}^{\text{attr}}, \boldsymbol{x})}}} \nonumber
\end{align}
The ranges of indexes in Eq. \eqref{equ:inference} are identical to the ones in Eq. \eqref{equ:generative}, and are omitted for simplicity. The choices of terms on the right-hand side of Eq. \eqref{equ:inference} are
\begin{align*}
	q(\boldsymbol{z}_{k}^{\text{attr}}|\boldsymbol{x}) & = \Normal\big(\boldsymbol{z}_{k}^{\text{attr}}; \boldsymbol{\mu}_{k}^{\text{attr}}, \diag(\boldsymbol{\sigma}_{k}^{\text{attr}})^2\big) \\
	q(\rho_{k}|\boldsymbol{x}) & = \Beta\big(\rho_{k}; \tau_{k,1}, \tau_{k,2}\big) \\
	q(z_{k}^{\text{prs}}|\boldsymbol{x}) & = \Bernoulli\big(z_{k}^{\text{prs}}; \kappa_{k}\big) \\
	q(\boldsymbol{z}_{m}^{\text{view}}|\boldsymbol{x}) & = \Normal\big(\boldsymbol{z}_{m}^{\text{view}}; \boldsymbol{\mu}_{m}^{\text{view}}, \diag(\boldsymbol{\sigma}_{k}^{\text{attr}})^2\big) \\
	q(z_{m,k,n}^{\text{shp}}|\boldsymbol{z}_{m}^{\text{view}}, \boldsymbol{z}_{k}^{\text{attr}}, \boldsymbol{x}) & = p(z_{m,k,n}^{\text{shp}}|\boldsymbol{z}_{m}^{\text{view}}, \boldsymbol{z}_{k}^{\text{attr}})
\end{align*}
In the variational distribution, $q(\boldsymbol{z}_{k}^{\text{attr}}|\boldsymbol{x})$ and $q(\boldsymbol{z}_{m}^{\text{view}}|\boldsymbol{x})$ are normal distributions with diagonal covariance matrices. $z_{k}^{\text{prs}}$ is assumed to be independent of $\rho_{k}$ given $\boldsymbol{x}$, and $q(\rho_{k}|\boldsymbol{x})$ and $q(z_{k}^{\text{prs}}|\boldsymbol{x})$ are chosen to be a beta distribution and a Bernoulli distribution, respectively. The advantage of this formulation is that the Kullback-Leibler (KL) divergence between $q(\rho_{k}|\boldsymbol{x}) q(z_{k}^{\text{prs}}|\boldsymbol{x})$ and $p(\rho_{k}) p(z_{k}^{\text{prs}}|\rho_{k})$ has a closed-form solution. For simplicity, $q(z_{m,k,n}^{\text{shp}}|\boldsymbol{z}_{m}^{\text{view}}, \boldsymbol{z}_{k}^{\text{attr}}, \boldsymbol{x})$ is assumed to be identical to $p(z_{m,k,n}^{\text{shp}}|\boldsymbol{z}_{m}^{\text{view}}, \boldsymbol{z}_{k}^{\text{attr}})$ in the generative model, so that no extra inference network is needed for $z_{m,k,n}^{\text{shp}}$. The procedure to compute the parameters $\boldsymbol{\mu}^{\text{attr}}$, $\boldsymbol{\sigma}^{\text{attr}}$, $\boldsymbol{\tau}$, $\boldsymbol{\kappa}$, $\boldsymbol{\mu}^{\text{view}}$, and $\boldsymbol{\sigma}^{\text{view}}$ of these distributions is presented in Algorithm \ref{alg:infer}, and the brief explanations are given below.

First, the feature maps $\boldsymbol{y}_{m}^{\text{feat}}$ of each image $\boldsymbol{x}_{m}$ are extracted by a neural network $g_{\text{feat}}$. Next, intermediate variables $\boldsymbol{y}^{\text{view}}$ and $\boldsymbol{y}^{\text{attr}}$ which fully characterize parameters of the viewpoint-dependent ($\boldsymbol{\mu}^{\text{view}}$ and $\boldsymbol{\sigma}^{\text{view}}$) and viewpoint-independent ($\boldsymbol{\mu}^{\text{attr}}$, $\boldsymbol{\sigma}^{\text{attr}}$, $\boldsymbol{\tau}$, and $\boldsymbol{\kappa}$) latent variables are not directly estimated, but instead randomly initialized from normal distributions with learnable parameters ($\hat{\boldsymbol{\mu}}^{\text{view}}$, $\hat{\boldsymbol{\sigma}}^{\text{view}}$, $\hat{\boldsymbol{\mu}}^{\text{attr}}$, and $\hat{\boldsymbol{\sigma}}^{\text{attr}}$) and then iteratively updated, considering that there are infinitely many possible solutions (e.g., due to the change of global coordinate system) to disentangle the image representations into a viewpoint-dependent part and a viewpoint-independent part. In each step of the iterative updates, information of images observed from different viewpoints are integrated using neural networks $g_{\text{key}}$, $g_{\text{qry}}$, $g_{\text{val}}$, and $g_{\text{upd}}$, based on attentions between feature maps $\boldsymbol{y}^{\text{feat}}$ and intermediate variables $\boldsymbol{y}^{\text{view}}$ and $\boldsymbol{y}^{\text{attr}}$. To achieve permutation equivariance, which has been considered as an important property in object-centric learning \cite{Equivariance2021EfficientIA}, objects and background are not distinguished in the initialization and updates of $\boldsymbol{y}^{\text{attr}}$, and the index $k^*$ that corresponds to background is determined after the iterative updates, by applying a neural network $g_{\text{sel}}$ to transform $\boldsymbol{y}^{\text{attr}}$ into parameters $\boldsymbol{\pi}$ of a categorical distribution and sampling from the distribution. After rearranging $\boldsymbol{y}_{0:K}^{\text{attr}}$ based on $k^*$, parameters of the variational distribution are computed by transforming $\boldsymbol{y}_{0}^{\text{attr}}$, $\boldsymbol{y}_{1:K}^{\text{attr}}$, and $\boldsymbol{y}_{1:M}^{\text{view}}$ with neural networks $g_{\text{bck}}$, $g_{\text{obj}}$, and $g_{\text{view}}$, respectively. For further details, please refer to the Supplementary Material.

\subsection{Learning of Neural Networks}

The neural networks used in both the generative model and the amortized variational inference (including learnable parameters $\hat{\boldsymbol{\mu}}^{\text{view}}$, $\hat{\boldsymbol{\sigma}}^{\text{view}}$, $\hat{\boldsymbol{\mu}}^{\text{attr}}$, and $\hat{\boldsymbol{\sigma}}^{\text{attr}}$), are jointly optimized by minimizing the negative value of evidence lower bound (ELBO) that serves as the loss function $\mathcal{L}$. The expression of $\mathcal{L}$ is briefly given below, and a more detailed version is included in the Supplementary Material.
\begin{align}
\label{equ:loss}
\mathcal{L} \!=\! & - \!\! \sum\nolimits_{m}\!{\sum\nolimits_{n}\!{\mathbb{E}_{q(\boldsymbol{\Omega}|\boldsymbol{x})}\big[\log{p(\boldsymbol{x}_{m,n}|\boldsymbol{z}^{\text{view}}\!, \boldsymbol{z}^{\text{attr}}\!, \boldsymbol{z}^{\text{prs}}\!, \boldsymbol{z}^{\text{shp}})}\big]}} \nonumber\\
& + \!\! \sum\nolimits_{m}\!{D_{\text{KL}}\big(q(\boldsymbol{z}_{m}^{\text{view}}|\boldsymbol{x})||p(\boldsymbol{z}_{m}^{\text{view}})\big)} \nonumber\\
& + \!\! \sum\nolimits_{k}\!{D_{\text{KL}}\big(q(\boldsymbol{z}_{k}^{\text{attr}}|\boldsymbol{x})||p(\boldsymbol{z}_{k}^{\text{attr}})\big)} \nonumber\\
& + \!\! \sum\nolimits_{k}\!{D_{\text{KL}}\big(q(\rho_{k}|\boldsymbol{x})||p(\rho_{k})\big)} \nonumber\\
& + \!\! \sum\nolimits_{k}\!{\mathbb{E}_{q(\rho_{k}|\boldsymbol{x})}\big[D_{\text{KL}}\big(q(z_{k}^{\text{prs}}|\boldsymbol{x})||p(z_{k}^{\text{prs}}|\rho_{k})\big)\big]}
\end{align}
In Eq. \eqref{equ:loss}, the first term is negative log-likelihood, and the rest four terms are Kullback-Leibler (KL) divergences that are computed by $D_{\text{KL}}(q||p) \!=\! \mathbb{E}_{q}[\log{q} - \log{p}]$. The loss function is optimized using the gradient-based method. All the KL divergences have closed-form solutions, and the gradients of these terms can be easily computed. The negative log-likelihood cannot be computed analytically, and the gradients of this term is approximated by sampling latent variables $\boldsymbol{z}^{\text{view}}$, $\boldsymbol{z}^{\text{attr}}$, $\boldsymbol{z}^{\text{prs}}$, and $\boldsymbol{z}^{\text{shp}}$ from the variational distribution $q(\boldsymbol{\Omega}|\boldsymbol{x})$. To reduce the variances of gradients, the continuous variables $\boldsymbol{z}^{\text{view}}$ and $\boldsymbol{z}^{\text{attr}}$ are sampled using the reparameterization trick \cite{Salimans2012FixedFormVP,Kingma2014AutoEncodingVB}, and the discrete variables $\boldsymbol{z}^{\text{prs}}$ and $\boldsymbol{z}^{\text{shp}}$ are approximated using a continuous relaxation \cite{Maddison2017TheCD,Jang2017CategoricalRW}. To learn the neural network $g_{\text{sel}}$ that computes parameters of the categorical distribution from which the background index $k^*$ is sampled, NVIL \cite{Mnih2014NeuralVI} is applied to obtain low-variance and unbiased estimates of gradients.

\section{Experiments}

In this section, we aim to verify that the proposed method\footnote{Code is available at \url{https://git.io/JDnne}.}:
\begin{itemize}
	\item is able to learn from multiple viewpoints \emph{without any supervision}, which \emph{cannot} be solved by existing methods;
	\item competes with existing state-of-the-art methods that use \emph{viewpoint annotations} in the learning;
	\item is comparable to the state-of-the-arts under an \emph{extreme condition} that scenes are observed from one viewpoint.
\end{itemize}

\noindent\textbf{Evaluation Metrics:} Several metrics are used to evaluate the performance from four aspects. 1) \emph{Adjusted Rand Index} (ARI) \cite{Hubert1985ComparingP} and \emph{Adjusted Mutual Information} (AMI) \cite{Nguyen2010InformationTM} assess the quality of segmentation, i.e., how accurately images are partitioned into different objects and background. Previous work usually evaluates ARI and AMI only at pixels belong to objects, and how accurately background is separated from objects is unclear. We evaluate ARI and AMI under two conditions. ARI-A and AMI-A are computed considering both objects and background, while ARI-O and AMI-O are computed considering only objects. 2) \emph{Intersection over Union} (IoU) and \emph{$F_1$ score} (F1) assess the quality of amodal segmentation, i.e., how accurately complete shapes of objects are estimated. 3) \emph{Object Counting Accuracy} (OCA) assesses the accuracy of the estimated number of objects. 4) \emph{Object Ordering Accuracy} (OOA) as used in \cite{Yuan2019GenerativeMO} assesses the accuracy of the estimated pairwise ordering of objects. Formal definitions of these metrics are included in the Supplementary Material.

\subsection{Multi-Viewpoint Learning}

\noindent\textbf{Datasets:} The experiments are performed on four multi-viewpoint variants (referred to as CLEVR-M1 to CLEVR-M4) of the commonly used CLEVR dataset that differ in the ranges to sample viewpoints and in the attributes of objects. CLEVR-M3/CLEVR-M4 is harder than CLEVR-M1/CLEVR-M2 in that the poses of objects are more dissimilar in different images of the same visual scene because viewpoints are sampled from a larger range. CLEVR-M2/CLEVR-M4 is harder than CLEVR-M1/CLEVR-M3 in that there are fewer visual cues to distinguish objects from one another because all the objects in the same visual scene share the same colors, shapes, and materials. Further details are described in the Supplementary Material.

\noindent\textbf{Comparison Methods:} It is worth noting that the proposed method cannot be directly compared with existing methods in the novel problem setting considered in this paper. To verify that the proposed method can effectively achieve object constancy, a baseline method that does not maintain the identities of objects across viewpoints is compared with. This baseline method is derived from the proposed method by assigning each viewpoint a separate set of latent variables $\boldsymbol{z}^{\text{attr}}$, $\boldsymbol{\rho}$, and $\boldsymbol{z}^{\text{prs}}$ (all latent variables are viewpoint-dependent). To verify that the proposed method can effectively learn without supervision, we compare it with MulMON \cite{Li2020LearningOR}, which solves a \emph{simpler} problem by using viewpoint annotations in both learning and testing. Another representative partially supervised method ROOTS \cite{Chen2020ObjectCentricRA} is not compared with because the official code is not publicly available.

\begin{figure}[t]
	\centering
	\includegraphics[width=0.932\columnwidth]{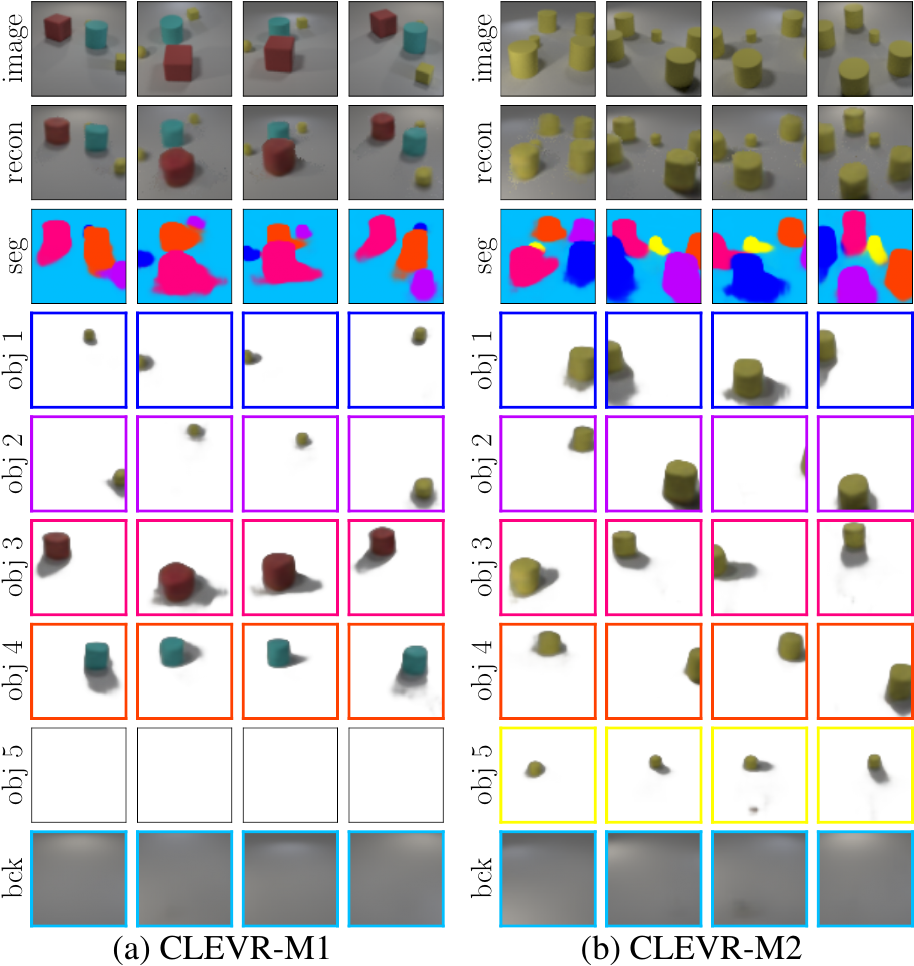}
	\caption{Scene decomposition results of the proposed method in the multi-viewpoint learning setting. Objects are sorted based on the estimated $\boldsymbol{z}^{\text{prs}}$. Models are tested with $K \!=\! 7$, and the last two objects with $z_{k}^{\text{prs}} \!=\! 0$ are not shown.}
	\label{fig:proposed_test_4_2}
\end{figure}

\noindent\textbf{Scene Decomposition:}
Qualitative results of the proposed method evaluated on the CLEVR-M1 and CLEVR-M2 datasets are shown in Figure \ref{fig:proposed_test_4_2}. The proposed method is able to achieve \emph{object constancy} even if objects are fully occluded (object 1 in columns 1 and 4 of sub-figure (a)). In addition, under the circumstances that objects are less identifiable and the poses of objects vary significantly across different viewpoints (objects 1 $\!\sim\!$ 4 in sub-figure (b)), the proposed method can also correctly identify the same objects across viewpoints. The proposed method tends to treat \emph{shadows} as parts of objects instead of background, which is desirable because lighting effects are not explicitly modeled and the shadows will change accordingly as the objects move. More results can be found in the Supplementary Material.

Quantitative comparison of scene decomposition performance on all the datasets is presented in Table \ref{tab:multi}. The proposed method achieves high ARI-O, AMI-O, and OOA scores. As for ARI-A, AMI-A, IoU, and F1, the achieved performance is not so well. The major reason is that the proposed method tends to treat regions of shadows as objects, while they are considered as background in the ground truth annotations. MulMON also tends to incorrectly estimate shadows as objects, but slightly outperforms the proposed method in terms of ARI-A and AMI-A on all the datasets, possibly because MulMON does not explicitly model the complete shapes, the number, and the depth ordering of objects, but directly computes the perceived shapes using the softmax function, which makes it easier to learn the boundary regions of objects. For the similar reason, the IoU, F1, and OOA scores which require the estimations of complete shapes and depth ordering are not evaluated for MulMON. The OCA scores are computed based on the heuristically estimated number of objects (details in the Supplementary Material). The \emph{unsupervised} proposed method achieves competitive or slightly better results compared to the \emph{partially supervised} MulMON, which has validated the motivation of the proposed method.

\begin{figure}[t]
	\centering
	\includegraphics[width=0.94\columnwidth]{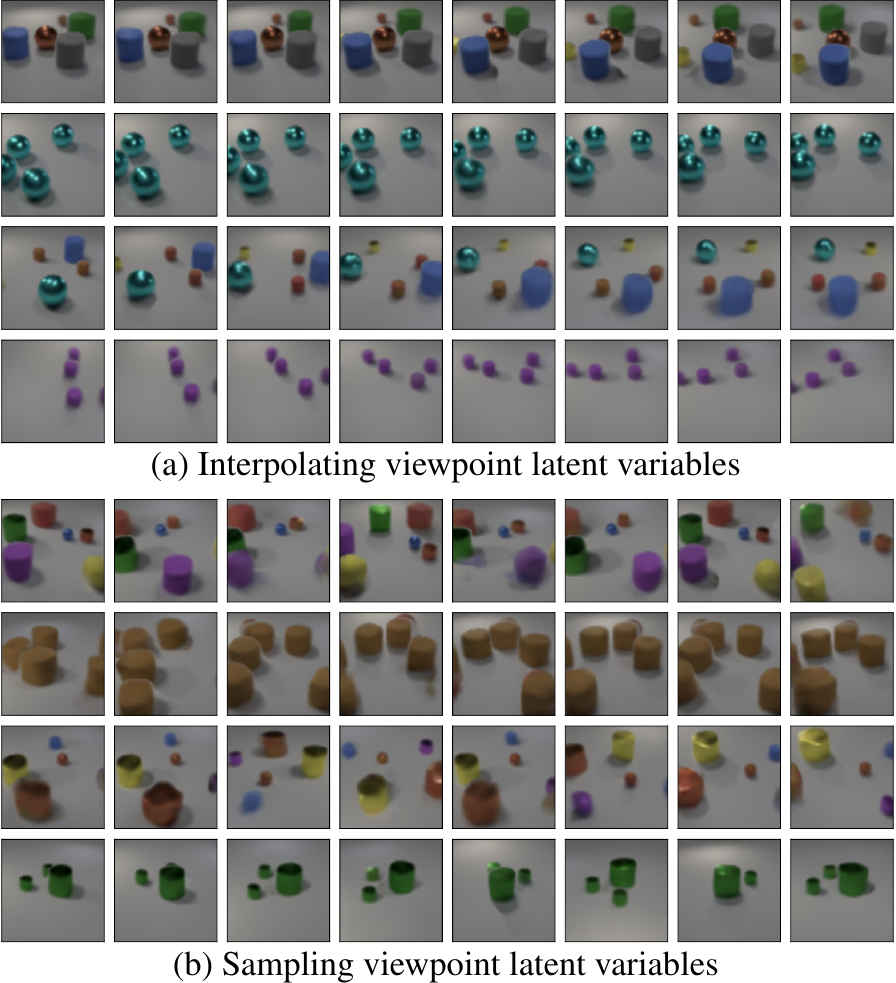}
	\caption{Results of interpolating and sampling viewpoints in latent space. The $i$th row of each sub-figure corresponds to the results evaluated on the CLEVR-M$\{i\}$ dataset.}
	\label{fig:viewpoint_test_2}
\end{figure}

\noindent\textbf{Generalizability:}
Because visual scenes are modeled compositionally by the proposed method, the trained models are generalizable to novel scenes containing more numbers of objects than the ones used for training. Evaluations of generalizability are included in the Supplementary Material. Although the increased number of objects makes it more difficult to extract compositional scene representations, the proposed method performs reasonably well.

\begin{table*}[ht]
	\centering
	\begin{small}
		\addtolength{\tabcolsep}{-2.8pt}
		\begin{tabular}{c|c|C{0.62in}C{0.62in}C{0.62in}C{0.62in}C{0.62in}C{0.62in}C{0.62in}C{0.62in}}
			\toprule
			Dataset          &  Method  &          ARI-A          &          AMI-A          &          ARI-O          &          AMI-O          &           IoU           &           F1            &           OCA           &           OOA           \\ \midrule
			\multirow{3}{*}{CLEVR-M1} & Baseline &     0.512$\pm$9e-4      &     0.361$\pm$3e-3      &     0.269$\pm$1e-2      &     0.418$\pm$1e-2      &     0.171$\pm$3e-3      &     0.279$\pm$4e-3      &     0.004$\pm$5e-3      &     0.628$\pm$3e-2      \\
			&  MulMON  & \textbf{0.615}$\pm$2e-3 & \textbf{0.560}$\pm$2e-3 &     0.927$\pm$5e-3      &     0.917$\pm$2e-3      &           N/A           &           N/A           &     0.446$\pm$5e-2      &           N/A           \\
			& Proposed &     0.507$\pm$2e-3      &     0.486$\pm$2e-3      & \textbf{0.948}$\pm$3e-3 & \textbf{0.934}$\pm$2e-3 & \textbf{0.442}$\pm$3e-3 & \textbf{0.603}$\pm$3e-3 & \textbf{0.730}$\pm$5e-2 & \textbf{0.970}$\pm$1e-2 \\ \midrule
			\multirow{3}{*}{CLEVR-M2} & Baseline &     0.505$\pm$1e-3      &     0.356$\pm$3e-3      &     0.274$\pm$1e-2      &     0.422$\pm$9e-3      &     0.167$\pm$4e-3      &     0.273$\pm$5e-3      &     0.004$\pm$5e-3      &     0.682$\pm$2e-2      \\
			&  MulMON  & \textbf{0.602}$\pm$7e-4 & \textbf{0.550}$\pm$4e-4 &     0.939$\pm$3e-3      &     0.926$\pm$2e-3      &           N/A           &           N/A           &     0.570$\pm$5e-2      &           N/A           \\
			& Proposed &     0.507$\pm$3e-3      &     0.479$\pm$2e-3      & \textbf{0.941}$\pm$3e-3 & \textbf{0.933}$\pm$2e-3 & \textbf{0.428}$\pm$3e-3 & \textbf{0.587}$\pm$4e-3 & \textbf{0.686}$\pm$3e-2 & \textbf{0.939}$\pm$2e-2 \\ \midrule
			\multirow{3}{*}{CLEVR-M3} & Baseline &     0.531$\pm$1e-3      &     0.372$\pm$3e-3      &     0.278$\pm$1e-2      &     0.425$\pm$1e-2      &     0.173$\pm$5e-3      &     0.283$\pm$7e-3      &     0.000$\pm$0e-0      &     0.600$\pm$5e-2      \\
			&  MulMON  & \textbf{0.591}$\pm$7e-3 & \textbf{0.552}$\pm$3e-3 &     0.938$\pm$2e-3      &     0.923$\pm$2e-3      &           N/A           &           N/A           &     0.424$\pm$5e-2      &           N/A           \\
			& Proposed &     0.534$\pm$2e-3      &     0.498$\pm$2e-3      & \textbf{0.939}$\pm$5e-3 & \textbf{0.929}$\pm$3e-3 & \textbf{0.453}$\pm$3e-3 & \textbf{0.610}$\pm$4e-3 & \textbf{0.632}$\pm$3e-2 & \textbf{0.974}$\pm$8e-3 \\ \midrule
			\multirow{3}{*}{CLEVR-M4} & Baseline &     0.519$\pm$1e-3      &     0.365$\pm$1e-3      &     0.280$\pm$6e-3      &     0.428$\pm$6e-3      &     0.170$\pm$4e-3      &     0.278$\pm$5e-3      &     0.000$\pm$0e-0      &     0.633$\pm$5e-2      \\
			&  MulMON  & \textbf{0.640}$\pm$4e-4 & \textbf{0.578}$\pm$6e-4 & \textbf{0.936}$\pm$3e-3 & \textbf{0.927}$\pm$2e-3 &           N/A           &           N/A           &     0.490$\pm$3e-2      &           N/A           \\
			& Proposed &     0.473$\pm$2e-3      &     0.452$\pm$2e-3      &     0.923$\pm$4e-3      &     0.922$\pm$2e-3      & \textbf{0.401}$\pm$1e-3 & \textbf{0.558}$\pm$1e-3 & \textbf{0.606}$\pm$8e-3 & \textbf{0.853}$\pm$2e-2 \\ \bottomrule
		\end{tabular}
	\end{small}
	\caption{Comparison of scene decomposition performance when learning from multiple viewpoints. All the methods are trained and tested with $M \!=\! 4$ and $K \!=\! 7$. The proposed \emph{fully unsupervised} method achieves \emph{competitive} or \emph{slightly better} results compared with MulMON with \emph{viewpoint supervision}.}
	\label{tab:multi}
\end{table*}

\begin{table*}[ht]
	\centering
	\begin{small}
		\addtolength{\tabcolsep}{-2.2pt}
		\begin{tabular}{c|c|C{0.62in}C{0.62in}C{0.62in}C{0.62in}C{0.62in}C{0.62in}C{0.62in}C{0.62in}}
			\toprule
			Dataset          &  Method   &          ARI-A          &          AMI-A          &          ARI-O          &          AMI-O          &           IoU           &           F1            &           OCA           &           OOA           \\ \midrule
			\multirow{4}{*}{dSprites} & Slot Attn &     0.142$\pm$3e-3      &     0.286$\pm$2e-3      &     0.935$\pm$1e-3      &     0.917$\pm$1e-3      &           N/A           &           N/A           &     0.000$\pm$0e-0      &           N/A           \\
			&   GMIOO   & \textbf{0.969}$\pm$2e-4 & \textbf{0.922}$\pm$5e-4 & \textbf{0.956}$\pm$1e-3 & \textbf{0.950}$\pm$9e-4 & \textit{0.860}$\pm$8e-4 & \textit{0.911}$\pm$7e-4 & \textbf{0.874}$\pm$2e-3 & \textit{0.891}$\pm$5e-3 \\
			&   SPACE   &     0.946$\pm$7e-4      &     0.874$\pm$6e-4      &     0.858$\pm$1e-3      &     0.870$\pm$5e-4      &     0.729$\pm$7e-4      &     0.805$\pm$5e-4      &     0.587$\pm$4e-3      &     0.624$\pm$1e-2      \\
			& Proposed  & \textit{0.959}$\pm$2e-4 & \textit{0.907}$\pm$4e-4 & \textit{0.939}$\pm$7e-4 & \textit{0.922}$\pm$7e-4 & \textbf{0.861}$\pm$8e-4 & \textbf{0.912}$\pm$8e-4 & \textit{0.813}$\pm$4e-3 & \textbf{0.899}$\pm$8e-3 \\ \midrule
			\multirow{4}{*}{Abstract} & Slot Attn & \textbf{0.940}$\pm$5e-4 & \textbf{0.877}$\pm$3e-4 &     0.935$\pm$8e-4      &     0.903$\pm$7e-4      &           N/A           &           N/A           &     0.888$\pm$5e-3      &           N/A           \\
			&   GMIOO   &     0.832$\pm$2e-4      &     0.751$\pm$3e-4      & \textit{0.941}$\pm$2e-3 & \textit{0.927}$\pm$1e-3 & \textit{0.750}$\pm$8e-4 & \textit{0.848}$\pm$8e-4 & \textbf{0.955}$\pm$2e-3 & \textit{0.940}$\pm$3e-3 \\
			&   SPACE   & \textit{0.888}$\pm$6e-4 &     0.797$\pm$6e-4      &     0.816$\pm$1e-3      &     0.817$\pm$2e-3      &     0.722$\pm$7e-4      &     0.798$\pm$8e-4      &     0.685$\pm$2e-3      &     0.799$\pm$5e-3      \\
			& Proposed  &     0.887$\pm$3e-4      & \textit{0.812}$\pm$4e-4 & \textbf{0.947}$\pm$9e-4 & \textbf{0.933}$\pm$9e-4 & \textbf{0.801}$\pm$3e-4 & \textbf{0.883}$\pm$2e-4 & \textit{0.940}$\pm$6e-3 & \textbf{0.962}$\pm$2e-3 \\ \midrule
			\multirow{4}{*}{CLEVR}   & Slot Attn &     0.026$\pm$2e-4      &     0.240$\pm$3e-4      & \textbf{0.985}$\pm$6e-4 & \textbf{0.983}$\pm$3e-4 &           N/A           &           N/A           &     0.002$\pm$1e-3      &           N/A           \\
			&   GMIOO   & \textit{0.716}$\pm$5e-4 & \textit{0.665}$\pm$4e-4 &     0.943$\pm$1e-3      &     0.955$\pm$8e-4      & \textit{0.605}$\pm$2e-3 &     0.725$\pm$2e-3      &     0.683$\pm$2e-3      &     0.906$\pm$4e-3      \\
			&   SPACE   & \textbf{0.860}$\pm$3e-4 & \textbf{0.796}$\pm$3e-4 &     0.976$\pm$3e-4      &     0.973$\pm$1e-4      & \textbf{0.776}$\pm$7e-4 & \textbf{0.863}$\pm$7e-4 & \textit{0.711}$\pm$2e-3 & \textit{0.936}$\pm$7e-3 \\
			& Proposed  &     0.649$\pm$1e-4      &     0.614$\pm$2e-4      & \textit{0.982}$\pm$9e-4 & \textit{0.978}$\pm$5e-4 &     0.591$\pm$6e-4      & \textit{0.736}$\pm$7e-4 & \textbf{0.875}$\pm$8e-3 & \textbf{0.952}$\pm$5e-3 \\ \bottomrule
		\end{tabular}
	\end{small}
	\caption{Comparison of scene decomposition performance when learning from a single viewpoint. All the methods are trained and tested with $K \!=\! 6$, $K \!=\! 5$, and $K \!=\! 7$ on the dSprites, Abstract, and CLEVR datasets, respectively. The proposed method is \emph{comparable} with the state-of-the-arts under the extreme condition that visual scenes are observed from one viewpoint.}
	\label{tab:single}
\end{table*}

\noindent\textbf{Viewpoint Estimation:} The proposed method is able to estimate the viewpoints of images, under the condition that the viewpoint-independent attributes of objects and background are known. More specifically, given the approximate posteriors of object-centric representations, the proposed method is able to infer the corresponding viewpoint representations of different observations of the same visual scene. Please refer to the Supplementary Material for more details.

\noindent\textbf{Viewpoint Modification:} Multi-viewpoint images of the same visual scene can be generated by first inferring compositional scene representations and then modifying viewpoint latent variables. Results of interpolating and sampling viewpoint latent variables are illustrated in Figure \ref{fig:viewpoint_test_2}. The proposed method is able to appropriately modify viewpoints.

\subsection{Single-Viewpoint Learning}

\noindent\textbf{Datasets:}
Three datasets are constructed based on the dSprites \cite{dsprites17}, Abstract Scene \cite{Zitnick2013BringingSI}, and CLEVR \cite{Johnson2017CLEVRAD} datasets, in a way similar to the Multi-Objects Datasets \cite{multiobjectdatasets19} but provides extra annotations (for evaluation only) of complete shapes of objects. These datasets are referred to as \emph{dSprites}, \emph{Abstract}, and \emph{CLEVER} for simplicity. More details are provided in the Supplementary Material.

\noindent\textbf{Comparison Methods:} The proposed method is compared with three state-of-the-art compositional scene representation methods. Slot Attention \cite{Locatello2020ObjectCentricLW} is chosen because the proposed method adopts a similar attention mechanism in the inference. GMIOO \cite{Yuan2019GenerativeMO} and SPACE \cite{Lin2020SPACEUO} are chosen because they are two representative methods that also explicitly model the varying number of objects, and can distinguish background from objects and determine the depth ordering of objects.

\noindent\textbf{Experimental Results:} Comparison of scene decomposition performance under the extreme condition that each visual scene is only observed from one viewpoint is shown in Table \ref{tab:single}. The proposed method is competitive with the state-of-the-arts and achieves the best or the second-best scores in almost all the cases. The Supplementary Material includes discussions and further quantitative and qualitative results.

\section{Conclusions}

In this paper, we have considered a novel problem of learning compositional scene representations from multiple unspecified viewpoints in a fully unsupervised way, and proposed a deep generative model called OCLOC to solve this problem. On several specifically designed synthesized datasets, the proposed fully unsupervised method achieves competitive or slightly better results compared with a state-of-the-art method with viewpoint supervision, which has validated the effectiveness of the proposed method.

\section{Acknowledgments}

This work was supported in part by the National Natural Science Foundation of China (No.62176060), STCSM project (No.20511100400), Shanghai Municipal Science and Technology Major Project (No.2021SHZDZX0103), Shanghai Research and Innovation Functional Program (No.17DZ2260900), and the Program for Professor of Special Appointment (Eastern Scholar) at Shanghai Institutions of Higher Learning.

\bibliography{reference}

\clearpage
\appendix

\section{Details of Algorithm 1}

The feature maps $\boldsymbol{y}_{m}^{\text{feat}} \!\in\! \mathbb{R}^{N \times D_{\text{ft}}}$ summarize the information of nearby region at each pixel in the $m$th image $\boldsymbol{x}_{m} \!\in\! \mathbb{R}^{N \times C}$, and are extracted by transforming $\boldsymbol{x}_{m}$ with a neural network $g_{\text{feat}}$. Considering that there are infinitely many possible solutions (e.g., due to the change of global coordinate system) to disentangle the image representations into a viewpoint-dependent part and a viewpoint-independent part, parameters of the variational distribution are first randomly initialized and then iteratively updated. To simplify the updates of parameters of the variational distribution, we use intermediate variables $\boldsymbol{y}^{\text{view}} \!\in\! \mathbb{R}^{M \!\times\! D_{\text{vw}}}$ and $\boldsymbol{y}^{\text{attr}} \!\in\! \mathbb{R}^{(K \!+\! 1) \!\times\! D_{\text{at}}}$ to represent parameters of the viewpoint-dependent part ($\boldsymbol{\mu}^{\text{view}}$ and $\boldsymbol{\sigma}^{\text{view}}$) and the viewpoint-independent part ($\boldsymbol{\mu}^{\text{attr}}$, $\boldsymbol{\sigma}^{\text{attr}}$, $\boldsymbol{\tau}$, and $\boldsymbol{\kappa}$), respectively. These intermediate variables are sampled independently from normal distributions with learnable parameters $\hat{\boldsymbol{\mu}}^{\text{view}}$, $\hat{\boldsymbol{\sigma}}^{\text{view}}$, $\hat{\boldsymbol{\mu}}^{\text{attr}}$, and $\hat{\boldsymbol{\sigma}}^{\text{attr}}$. To achieve permutation equivariance, which has been considered as an important property in object-centric learning \cite{Equivariance2021EfficientIA}, objects and background are not distinguished in the initialization and updates of $\boldsymbol{y}^{\text{attr}}$, and the index that corresponds to background is determined after the iterative updates.

In each step of the iterative updates, intermediate variables $\boldsymbol{y}^{\text{view}} \!\in\! \mathbb{R}^{M \!\times\! D_{\text{vw}}}$ and $\boldsymbol{y}^{\text{attr}} \!\in\! \mathbb{R}^{(K \!+\! 1) \!\times\! D_{\text{at}}}$ are first broadcasted and concatenated to form $\boldsymbol{y}^{\text{full}} \!\in\! \mathbb{R}^{M \!\times\! (K \!+\! 1) \!\times\! (D_{\text{vw}} \!+\! D_{\text{at}})}$. Next, the attention maps $\boldsymbol{a}_{m} \!\in\! [0, 1]^{(K \!+\! 1) \!\times\! N}$ ($1 \!\leq\! m \!\leq\! M$) are computed separately for each viewpoint, by normalizing the similarities between the \emph{keys} $g_{\text{key}}(\boldsymbol{y}_{m}^{\text{feat}}) \!\in\! \mathbb{R}^{N \!\times\! D_{\text{key}}}$ and the \emph{queries} $g_{\text{qry}}(\boldsymbol{y}_{m}^{\text{full}}) \!\in\! \mathbb{R}^{(K \!+\! 1) \!\times\! D_{\text{key}}}$ across different objects and background (i.e., $(\forall m, n) \sum_{k=0}^{K}{a_{m,k,n}} \!=\! 1$) with temperature $\sqrt{\!D_{\text{key}}}$. Both $g_{\text{key}}$ and $g_{\text{qry}}$ are neural networks, and the similarities are measured by first broadcasting \emph{keys} and \emph{queries} to $\mathbb{R}^{(K \!+\! 1) \!\times\! N \!\times\! D_{\text{key}}}$, and then performing dot product in the last dimension. After that, $\boldsymbol{u} \!\in\! \mathbb{R}^{M \!\times\! (K \!+\! 1) \!\times\! D_{\text{val}}}$ which contains information to update $\boldsymbol{y}^{\text{view}}$ and $\boldsymbol{y}^{\text{attr}}$ is computed as the weighted average of the \emph{values} $g_{\text{val}}(\boldsymbol{y}_{m}^{\text{feat}}) \!\in\! \mathbb{R}^{N \!\times\! D_{\text{val}}}$ across $N$ pixels, with the attention maps $\boldsymbol{a}_{m} \!\in\! [0, 1]^{(K \!+\! 1) \!\times\! N}$ as weights. $g_{\text{val}}$ is a neural network, and the weighted average is computed by first broadcasting \emph{values} and the normalized weights $\softmax_N(\log{\boldsymbol{a}_{m}})$ to $\mathbb{R}^{(K \!+\! 1) \!\times\! N \!\times\! D_{\text{val}}}$, and then performing dot product in the second dimension. Finally, a neural network $g_{\text{upd}}$ is applied to transform $\boldsymbol{y}^{\text{full}}$ and $\boldsymbol{u}$ to $\boldsymbol{v}^{\text{view}} \!\in\! \mathbb{R}^{M \!\times\! (K \!+\! 1) \!\times\! D_{\text{vw}}}$ and $\boldsymbol{v}^{\text{attr}} \!\in\! \mathbb{R}^{M \!\times\! (K \!+\! 1) \!\times\! D_{\text{at}}}$, and intermediate variables $\boldsymbol{y}_{1:M}^{\text{view}}$ and $\boldsymbol{y}_{0:K}^{\text{attr}}$ are updated as the averages of $\boldsymbol{v}_{1:M,0:K}^{\text{view}}$ and $\boldsymbol{v}_{1:M,0:K}^{\text{attr}}$ across the second and the first dimensions, respectively.

After the iterative updates, each $\boldsymbol{y}_{k}^{\text{attr}}$ ($0 \!\leq\! k \!\leq\! K$) is transformed to a scalar by a neural network $g_{\text{sel}}$, which is expected to output high value for the $\boldsymbol{y}_{k}^{\text{attr}}$ that corresponds to background and low values for the rest ones corresponding to objects. After normalizing the $K \!+\! 1$ outputs to form valid parameters $\boldsymbol{\pi}_{0:K}$ of a categorical distribution and sampling the index $k^*$ of background from the distribution, $\boldsymbol{y}_{0:K}^{\text{attr}}$ is rearranged so that $\boldsymbol{y}_{0}^{\text{attr}}$ and $\boldsymbol{y}_{1:K}^{\text{attr}}$ correspond to background and objects, respectively. The final outputs of the inference, i.e., parameters of the variational distribution, are computed by transforming $\boldsymbol{y}_{0}^{\text{attr}}$, $\boldsymbol{y}_{1:K}^{\text{attr}}$, and $\boldsymbol{y}_{1:M}^{\text{view}}$ with neural networks $g_{\text{bck}}$, $g_{\text{obj}}$, and $g_{\text{view}}$, respectively.

\section{Details of Loss Function}

The loss function $\mathcal{L}$ can be decomposed as
\begin{equation*}
\label{equ-supp:loss}
\mathcal{L} = \tfrac{M N C}{2} \log{2 \pi \sigma_{\text{x}}^2} + \mathcal{L}_{\text{nll}} + \mathcal{L}_{\text{view}} + \mathcal{L}_{\text{attr}} + \mathcal{L}_{\rho} + \mathcal{L}_{\text{prs}}
\end{equation*}
The first term on the right-hand side of the above equation is a constant. The rest terms are computed by
\begin{align*}
\mathcal{L}_{\text{nll}} = \, & \frac{1}{2 \sigma_{\text{x}}^2} \! \sum_{m=1}^{M}{\!\sum_{n=1}^{N}{\!\mathbb{E}_{q(\boldsymbol{\Omega}|\boldsymbol{x})}{\!\Bigg[\!\bigg(\boldsymbol{x}_{m,n} \!-\! \sum_{k=0}^{K}{\!s_{m,k,n} \boldsymbol{a}_{m,k,n}}\bigg)^{\!\!2}\Bigg]}}} \\
\mathcal{L}_{\text{view}} = \, & \frac{1}{2} \! \sum_{m=1}^{M}{\!\sum_{i}{\!\big(\mbox{$\mu_{m,i}^{\text{view}}$}^2 + \mbox{$\sigma_{m,i}^{\text{view}}$}^2 - \log{\mbox{$\sigma_{m,i}^{\text{view}}$}^2} - 1\big)}} \\
\mathcal{L}_{\text{attr}} = \, & \frac{1}{2} \sum_{k=0}^{K}{\!\sum_{i}{\!\big(\mbox{$\mu_{k,i}^{\text{attr}}$}^2 + \mbox{$\sigma_{k,i}^{\text{attr}}$}^2 - \log{\mbox{$\sigma_{k,i}^{\text{attr}}$}^2} - 1\big)}} \\
\mathcal{L}_{\rho} = \, & \sum_{k=1}^{K}{\!\bigg(\!\log{\frac{\Gamma(\tau_{k,1} + \tau_{k,2})}{\Gamma(\tau_{k,1}) \Gamma(\tau_{k,2})}} - \log{\frac{\alpha}{K}}\bigg)} + \\
& \sum_{k=1}^{K}{\!\bigg(\!\!\Big(\tau_{k,1} - \frac{\alpha}{K}\Big) \psi(\tau_{k,1}) + (\tau_{k,2} - 1) \psi(\tau_{k,2})\!\bigg)} - \\
& \sum_{k=1}^{K}{\!\bigg(\!\!\Big(\tau_{k,1} + \tau_{k,2} - \frac{\alpha}{K} - 1\Big) \psi(\tau_{k,1} + \tau_{k,2})\!\bigg)} \\
\mathcal{L}_{\text{prs}} = \, & \sum_{k=1}^{K}{\!\Big(\psi(\tau_{k,1} + \tau_{k,2}) + \kappa_{k} \big(\!\log(\kappa_{k}) - \psi(\tau_{k,1})\big)\!\Big)} + \\
& \sum_{k=1}^{K}{\!\Big(\!(1 - \kappa_{k}) \big(\!\log(1 - \kappa_{k}) - \psi(\tau_{k,2})\big)\!\Big)}
\end{align*}
The $\Gamma$ and $\phi$ in the computations of both $\mathcal{L}_{\rho}$ and $\mathcal{L}_{\text{prs}}$ are gamma and digamma functions, respectively.

\section{Details of Evaluation Metrics}

Formal definitions of evaluation metrics are given below. These metrics are only used to assess the performance of the models after training. The best model parameters (i.e., parameters of neural networks) are chosen based on the value of loss function on the validation set. All the models are trained once and tested for five runs. The reported scores included both means and standard deviations.

\subsection{Adjusted Rand Index (ARI)}

$\hat{K}_{i}$ denotes the ground truth number of objects in the $i$th visual scene of the test set, and $\hat{\boldsymbol{r}}^{i} \!\in\! {\{0, 1\}^{M \!\times\! N \!\times\! (\hat{K}_{i} + 1)}}$ is the ground truth pixel-wise partition of objects and background in the $M$ images of this visual scene. $K$ denotes the maximum
number of objects that may appear in the visual scene, and $\boldsymbol{r}^{i} \!\in\! {\{0, 1\}^{M \!\times\! N \!\times\! (K + 1)}}$ is the estimated partition. ARI is compute using the following expressions.
\begin{equation*}
\text{ARI} = \frac{1}{I} \sum_{i=1}^{I}{\frac{b_{\text{all}}^{i} - b_{\text{row}}^{i} \cdot b_{\text{col}}^{i} / c^{i}}{\big(b_{\text{row}}^{i} + b_{\text{col}}^{i}\big) / 2 - b_{\text{row}}^{i} \cdot b_{\text{col}}^{i} / c^{i}}}
\end{equation*}
where
\begin{align*}
C(n, k) & = \frac{n!}{(n - k)! \, k!} \\
a_{\hat{k},k}^{i} & = \sum\nolimits_{m,n \in \mathcal{S}}{\big(\hat{r}_{m,n,\hat{k}}^{i} \cdot r_{m,n,k}^{i}\big)} \\
b_{\text{all}}^{i} & = \sum\nolimits_{\hat{k}=0}^{\hat{K}_{i}}{\sum\nolimits_{k=0}^{K}{C\big(a_{\hat{k},k}^{i}, 2\big)}} \\
b_{\text{row}}^i & = \sum\nolimits_{\hat{k}=0}^{\hat{K}_i}{C\Big(\sum\nolimits_{k=0}^{K}{a_{\hat{k},k}^{i}}, 2\Big)} \\ b_{\text{col}}^i & = \sum\nolimits_{k=0}^{K}{C\Big(\sum\nolimits_{\hat{k}=0}^{\hat{K}_i}{a_{\hat{k},k}^{i}}, 2\Big)} \\
c^{i} & = C\Big(\sum\nolimits_{\hat{k}=0}^{\hat{K}_i}{\sum\nolimits_{m,n \in \mathcal{S}}{\hat{r}_{m,n,\hat{k}}^i}}, 2\Big)
\end{align*}
When computing ARI-A, $\mathcal{S}$ is the collection of all pixels in the $M$ images, i.e., $\mathcal{S} \!=\! \{1, \dots, M\} \!\times\! \{1, \dots, N\}$. When computing ARI-O, $\mathcal{S}$ corresponds to all pixels belonging to objects in the $M$ images.

\subsection{Adjusted Mutual Information (AMI)}

The meanings of $\hat{K}_{i}$, $\hat{\boldsymbol{r}}^{i}$, $K$, and $\boldsymbol{r}^{i}$ are identical to the ones in the descriptions of ARI. AMI is computed by
\begin{equation*}
\text{AMI} = \frac{1}{I} \sum_{i=1}^{I}{\frac{\MI(\hat{\boldsymbol{l}}^{i}, \boldsymbol{l}^{i}) - \mathbb{E}[\MI(\hat{\boldsymbol{l}}^{i}, \boldsymbol{l}^{i})]}{\big(\!\entropy(\hat{\boldsymbol{l}}^{i}) + \entropy(\boldsymbol{l}^{i})\big) / 2 - \mathbb{E}[\MI(\hat{\boldsymbol{l}}^{i}, \boldsymbol{l}^{i})]}}
\end{equation*}
where
\begin{align*}
\hat{\boldsymbol{l}}^{i} & \in \{0, 1, \dots, \hat{K}_{i} \!+\! 1\}^{|\mathcal{S}|} \\
\hat{\boldsymbol{l}}_{j}^{i} & = \argmax\nolimits_{\hat{k}}{\hat{\boldsymbol{r}}_{m_j,n_j}^{i}}, \qquad (m_j, n_j) = \mathcal{S}_{j} \\
\boldsymbol{l}^{i} & \in \{0, 1, \dots, K \!+\! 1\}^{|\mathcal{S}|} \\
\boldsymbol{l}_{j}^{i} & = \argmax\nolimits_{k}{\boldsymbol{r}_{m_j,n_j}^{i}}, \qquad (m_j, n_j) = \mathcal{S}_{j}
\end{align*}
In the above expressions, $\MI$ denotes mutual information and $\entropy$ denotes entropy. When computing AMI-A/AMI-O, the choice of $\mathcal{S}$ is the same as ARI-A/ARI-O.

\subsection{Intersection over Union (IoU)}

$\hat{\boldsymbol{s}}^{i} \!\in\! [0, 1]^{M \!\times\! N \!\times\! \hat{K}_{i}}$ and $\boldsymbol{s}^{i} \!\in\! [0, 1]^{M \!\times\! N \!\times\! K}$ denote the ground truth and estimated shapes of objects in the $M$ images of the $i$th visual scene of the test set. IoU can be used to evaluate the performance of amodal instance segmentation \cite{Qi2019AmodalIS}. Compared to ARI and AMI, it provides extra information about the estimation of occluded regions of objects because complete shapes instead of perceived shapes of objects are used to compute this metric. Because both the number and the indexes of the estimated objects may be different from the ground truth, $\hat{\boldsymbol{s}}^{i}$ and $\boldsymbol{s}^{i}$ cannot be compared directly. Let $\Xi$ be the set of all the $K!$ possible permutations of the indexes $\{1, 2, \dots, K\}$. $\boldsymbol{\xi}^{i} \!\in\! \Xi$ is a permutation chosen based on the ground truth $\hat{\boldsymbol{r}}^{i}$ and estimated $\boldsymbol{r}^{i}$ partitions of objects and background using the following expression.
\begin{equation*}
\boldsymbol{\xi}^{i} = \max_{\boldsymbol{\xi} \in \Xi}{\sum_{k=1}^{\hat{K}_i}{\sum_{m=1}^{M}{\sum_{n=1}^{N}{\hat{r}_{m,n,k}^{i} \cdot r_{m,n,\xi_k^i}^{i}}}}}
\end{equation*}
IoU is computed by
\begin{equation*}
\text{IoU} = \frac{1}{I} \sum_{i=1}^{I}{\frac{1}{\hat{K}_i} \sum_{k=1}^{\hat{K}_i}{\frac{d_{\text{inter}}}{d_{\text{union}}}}}
\end{equation*}
where
\begin{align*}
d_{\text{inter}} & = \sum\nolimits_{m=1}^{M}{\sum\nolimits_{n=1}^{N}{\min(\hat{s}_{m,n,k}^{i}, s_{m,n,\xi_k^i}^{i})}} \\
d_{\text{union}} & = \sum\nolimits_{m=1}^{M}{\sum\nolimits_{n=1}^{N}{\max(\hat{s}_{m,n,k}^{i}, s_{m,n,\xi_k^i}^{i})}}
\end{align*}
Although the set $\Xi$ contain $K!$ elements, the permutation $\boldsymbol{\xi}^{i}$ can still be computed efficiently by formulating the computation as a linear sum assignment problem.

\subsection{$\boldsymbol{F_1}$ Score (F1)}

$F_1$ score can also be used to assess the performance of amodal segmentation like IoU, and is computed in a similar way. The meanings of $\hat{\boldsymbol{s}}^{i}$, $\boldsymbol{s}^{i}$, $\boldsymbol{\xi}$, and $\Xi$ as well as the computations of $d_{\text{inter}}$ and $d_{\text{union}}$ are identical to the ones in the descriptions of IoU. F1 is computed by
\begin{equation*}
\text{F1} = \frac{1}{I} \sum_{i=1}^{I}{\frac{1}{\hat{K}_i} \sum_{k=1}^{\hat{K}_i}{\frac{2 \cdot d_{\text{inter}}}{d_{\text{inter}} + d_{\text{union}}}}}
\end{equation*}

\begin{table*}[ht]
	\renewcommand*{\arraystretch}{1.5}
	\centering
	\begin{small}
		\addtolength{\tabcolsep}{-3.1pt}
		\begin{tabular}{|c|C{0.31in}|C{0.31in}|C{0.31in}|C{0.31in}|C{0.31in}|C{0.31in}|C{0.31in}|C{0.31in}|C{0.31in}|C{0.31in}|C{0.31in}|C{0.31in}|C{0.31in}|C{0.31in}|C{0.31in}|C{0.31in}|}
			\hline
			Dataset   &                   \multicolumn{4}{c|}{CLEVR-M1}                    &                   \multicolumn{4}{c|}{CLEVR-M2}                    &                   \multicolumn{4}{c|}{CLEVR-M3}                    &                   \multicolumn{4}{c|}{CLEVR-M4}                    \\ \hline
			Split    &     Train      &     Valid      &     Test 1     &     Test 2      &     Train      &     Valid      &     Test 1     &     Test 2      &     Train      &     Valid      &     Test 1     &     Test 2      &     Train      &     Valid      &     Test 1     &     Test 2      \\ \hline
			Scenes   &      5000      &      100       &      100       &       100       &      5000      &      100       &      100       &       100       &      5000      &      100       &      100       &       100       &      5000      &      100       &      100       &       100       \\ \hline
			Objects   & 3 $\!\sim\!$ 6 & 3 $\!\sim\!$ 6 & 3 $\!\sim\!$ 6 & 7 $\!\sim\!$ 10 & 3 $\!\sim\!$ 6 & 3 $\!\sim\!$ 6 & 3 $\!\sim\!$ 6 & 7 $\!\sim\!$ 10 & 3 $\!\sim\!$ 6 & 3 $\!\sim\!$ 6 & 3 $\!\sim\!$ 6 & 7 $\!\sim\!$ 10 & 3 $\!\sim\!$ 6 & 3 $\!\sim\!$ 6 & 3 $\!\sim\!$ 6 & 7 $\!\sim\!$ 10 \\ \hline
			Viewpoints &                      \multicolumn{4}{c|}{10}                       &                      \multicolumn{4}{c|}{10}                       &                      \multicolumn{4}{c|}{10}                       &                      \multicolumn{4}{c|}{10}                       \\ \hline
			Image Size &                \multicolumn{4}{c|}{64 $\times$ 64}                 &                \multicolumn{4}{c|}{64 $\times$ 64}                 &                \multicolumn{4}{c|}{64 $\times$ 64}                 &                \multicolumn{4}{c|}{64 $\times$ 64}                 \\ \hline
			Azimuth   &                  \multicolumn{4}{c|}{$[0, \pi]$}                   &                  \multicolumn{4}{c|}{$[0, \pi]$}                   &                 \multicolumn{4}{c|}{$[0, 2 \pi]$}                  &                 \multicolumn{4}{c|}{$[0, 2 \pi]$}                  \\[-2pt]
			Elevation  &             \multicolumn{4}{c|}{$[0.15\pi, 0.25\pi]$}              &             \multicolumn{4}{c|}{$[0.15\pi, 0.25\pi]$}              &              \multicolumn{4}{c|}{$[0.15\pi, 0.3\pi]$}              &              \multicolumn{4}{c|}{$[0.15\pi, 0.3\pi]$}              \\[-2pt]
			Distance  &               \multicolumn{4}{c|}{$[10.75, 11.75]$}                &               \multicolumn{4}{c|}{$[10.75, 11.75]$}                &                 \multicolumn{4}{c|}{$[10.5, 12]$}                  &                 \multicolumn{4}{c|}{$[10.5, 12]$}                  \\ \hline
			Colors   &                  \multicolumn{4}{c|}{not shared}                   &                    \multicolumn{4}{c|}{shared}                     &                  \multicolumn{4}{c|}{not shared}                   &                    \multicolumn{4}{c|}{shared}                     \\[-2pt]
			Shapes   &                  \multicolumn{4}{c|}{not shared}                   &                    \multicolumn{4}{c|}{shared}                     &                  \multicolumn{4}{c|}{not shared}                   &                    \multicolumn{4}{c|}{shared}                     \\[-2pt]
			Material  &                  \multicolumn{4}{c|}{not shared}                   &                    \multicolumn{4}{c|}{shared}                     &                  \multicolumn{4}{c|}{not shared}                   &                    \multicolumn{4}{c|}{shared}                     \\[-2pt]
			Size    &                  \multicolumn{4}{c|}{not shared}                   &                  \multicolumn{4}{c|}{not shared}                   &                  \multicolumn{4}{c|}{not shared}                   &                  \multicolumn{4}{c|}{not shared}                   \\[-2pt]
			Pose    &                  \multicolumn{4}{c|}{not shared}                   &                  \multicolumn{4}{c|}{not shared}                   &                  \multicolumn{4}{c|}{not shared}                   &                  \multicolumn{4}{c|}{not shared}                   \\ \hline
		\end{tabular}
	\end{small}
	\caption{Configurations of the datasets used in the multi-viewpoint learning setting. Line 1: names of datasets. Line 2: splits of datasets. Line 3: number of visual scenes in each split. Line 4: ranges to sample the number of objects per scene. Line 5: number of images that are observed from different viewpoints per scene. Line 6: width and height of each image. Lines 7-9: ranges to sample viewpoints. Lines 10-14: whether objects in the same visual scene share the same attributes.}
	\label{tab-supp:data_multi}
\end{table*}

\begin{table*}[ht]
	\renewcommand*{\arraystretch}{1.5}
	\centering
	\begin{small}
		\addtolength{\tabcolsep}{-1.8pt}
		\begin{tabular}{|c|C{0.4in}|C{0.4in}|C{0.4in}|C{0.4in}|C{0.4in}|C{0.4in}|C{0.4in}|C{0.4in}|C{0.4in}|C{0.4in}|C{0.4in}|C{0.4in}|}
			\hline
			Dataset   &                   \multicolumn{4}{c|}{dSprites}                   &                   \multicolumn{4}{c|}{Abstract}                   &                     \multicolumn{4}{c|}{CLEVR}                     \\ \hline
			Split    &     Train      &     Valid      &     Test 1     &     Test 2     &     Train      &     Valid      &     Test 1     &     Test 2     &     Train      &     Valid      &     Test 1     &     Test 2      \\ \hline
			Images    &     50000      &      1000      &      1000      &      1000      &     50000      &      1000      &      1000      &      1000      &     50000      &      1000      &      1000      &      1000       \\ \hline
			Objects   & 2 $\!\sim\!$ 5 & 2 $\!\sim\!$ 5 & 2 $\!\sim\!$ 5 & 6 $\!\sim\!$ 8 & 2 $\!\sim\!$ 4 & 2 $\!\sim\!$ 4 & 2 $\!\sim\!$ 4 & 5 $\!\sim\!$ 6 & 3 $\!\sim\!$ 6 & 3 $\!\sim\!$ 6 & 3 $\!\sim\!$ 6 & 7 $\!\sim\!$ 10 \\ \hline
			Image Size  &                \multicolumn{4}{c|}{64 $\times$ 64}                &                \multicolumn{4}{c|}{64 $\times$ 64}                &               \multicolumn{4}{c|}{128 $\times$ 128}                \\ \hline
			Min Visible &                     \multicolumn{4}{c|}{25\%}                     &                     \multicolumn{4}{c|}{25\%}                     &                  \multicolumn{4}{c|}{128 pixels}                   \\ \hline
		\end{tabular}
	\end{small}
	\caption{Configurations of the datasets used in the single-viewpoint learning setting. Line 1: names of datasets. Line 2: splits of datasets. Line 3: number of images in each split. Line 4: ranges to sample the number of objects per image. Line 5: width and height of each image. Lines 6: the minimum visible percentage or number of pixels per object.}
	\label{tab-supp:data_single}
\end{table*}

\subsection{Object Counting Accuracy (OCA)}

$\hat{K}_{i}$ and $\tilde{K}_{i}$ denote the ground truth number and the estimated number of objects in the $i$th visual scene of the test set. Let $\delta$ denote the Kronecker delta function. OCA is computed by
\begin{equation*}
\text{OCA} = \frac{1}{I} \sum\nolimits_{i=1}^{I}{\delta_{\hat{K}_{i}, \tilde{K}_{i}}}
\end{equation*}

\subsection{Object Ordering Accuracy (OOA)}

Let $\hat{t}_{m,k_1, k_2}^{i} \!\in\! \{0, 1\}$ and $t_{m,k_1, k_2}^{i} \!\in\! \{0, 1\}$ denote the ground truth and estimated pairwise orderings of the $k_1$th and $k_2$th objects in the $m$th viewpoint of the $i$th image. The correspondences between the ground truth and estimated indexes of objects are determined based on the permutation of indexes $\boldsymbol{\xi}^{i}$ as described in the computation of IoU. Because the relative ordering of objects is hard to estimate if these objects do not overlap, OOA is computed by
\begin{equation*}
\text{OOA} = \frac{1}{I} \! \sum_{i=1}^{I}{\!\frac{\sum\nolimits_{k_1=1}^{\hat{K}_i - 1}{\!\sum\nolimits_{k_2=k_1 + 1}^{\hat{K}_i}{w_{m\!,k_1\!,k_2}^{i} \!\cdot \delta_{\hat{t}_{m\!,k_1\!,k_2}^{i}\!, t_{m\!,k_1\!,k_2}^{i}}}}}{\sum\nolimits_{k_1=1}^{\hat{K}_i - 1}{\!\sum\nolimits_{k_2=k_1 + 1}^{\hat{K}_i}{w_{m\!,k_1\!,k_2}^{i}}}}}
\end{equation*}
where $w_{m\!,k_1\!,k_2}^{i}$ is the weight computed based on the ground truth complete shapes of objects $\hat{\boldsymbol{s}}^{i}$
\begin{equation*}
w_{m\!,k_1\!,k_2}^{i} = \sum\nolimits_{n=1}^{N}{\hat{s}_{m,n,k_1}^{i} \cdot \hat{s}_{m,n,k_2}^{i}}
\end{equation*}
$w_{m\!,k_1\!,k_2}^{i}$ measures the overlapped area of the ground truth shapes of the $k_1$th and the $k_2$th objects. The more the two objects overlap, the easier it is to determine the relative ordering of these objects, and thus the more important it is for the model to estimate the relative ordering correctly.

\section{Details of Datasets}

Configurations of the datasets used in the multi-viewpoint and single-viewpoint learning settings are presented in Table \ref{tab-supp:data_multi} and Table \ref{tab-supp:data_single}, respectively. Explanations of the configurations are described in the captions of the tables. The CLEVR-M and CLEVR datasets are generated based on the official code provided by \cite{Johnson2017CLEVRAD}. Images in the multi-viewpoint CLEVR-M dataset are generated with size $108 \!\times\! 80$ and cropped to size $64 \!\times\! 64$ at locations $10$ (up), $74$ (down), $22$ (left), and $86$ (right). Code is modified to skip the check of object visibility because the observations of objects vary as viewpoints change. Images in the single-viewpoint CLEVR dataset are generated with size $214 \!\times\! 160$ and cropped to size $128 \!\times\! 128$ at locations $19$ (up), $147$ (down), $43$ (left), and $171$ (right). Code is modified to ensure that at least $128$ pixels of each object is visible after cropping (instead of before cropping). In images of the dSprites datasets, the colors of backgrounds are sampled uniformly from grayscale RGB colors, and the colors of the objects provided by \cite{dsprites17} are sampled uniformly from RGB colors with the constraint that the $l_2$ distance between the colors of object and background is at least $0.5$ (the range of each channel is $[0, 1]$). As for the Abstract dataset, the background and $10$ objects in the Abstract Scene dataset \cite{Zitnick2013BringingSI} are selected to synthesize images. The colors of objects and background are both randomly permuted in HSV space (the range of each channel is $[0, 1]$). The H channel of all the pixels of the same object or background is added with the same random number sampled from $\mathcal{U}(-0.1, 0.1)$. And the S/V channel of all the pixels of the same object or background is multiplied with the same random number sampled from $\mathcal{U}(0.9, 1)$. If not explicitly mentioned, models are tested on the Test 1 splits and the corresponding experimental results are reported.

\section{Choices of Hyperparameters}

\subsection{Multi-Viewpoint Learning}

\subsubsection{Proposed Method}

In the generative model, the standard deviation $\sigma_{\text{x}}$ of the likelihood function is chosen to be $0.2$. The maximum number $K$ of objects that may appear in the visual scene is set to $7$ during training. The respective dimensionalities of latent variables $\boldsymbol{z}_{m}^{\text{view}}$, $\boldsymbol{z}_{0}^{\text{attr}}$, and $\boldsymbol{z}_{k}^{\text{attr}}$ with $1 \!\leq\! k \!\leq\! K$ are $4$, $8$, and $64$. $\alpha$ is $4.5$ and $\lambda$ is $0.5$. In the inference, the dimensionalities $D_{\text{vw}}$ and $D_{\text{at}}$ of intermediate variables $\boldsymbol{y}_{m}^{\text{view}}$ and $\boldsymbol{y}_{k}^{\text{attr}}$ are $8$ and $128$ respectively. $D_{\text{key}}$ is $64$, $D_{\text{val}}$ is $136$, and $T$ is $3$. In the learning, the batch size is chosen to be $32$. The initial learning rate is $4 \times 10^{-4}$, and is decayed exponentially with a factor $0.5$ every $\num[group-separator={,}]{50000}$ steps. In the first $\num[group-separator={,}]{10000}$ training steps, the learning rate is multiplied by a factor that is increased linearly from $0$ to $1$. We have found that the optimization of neural networks with randomly initialized weights tend to get stuck into undesired local optima. To solve this problem, a better initialization of weights is obtained by using only one viewpoint per visual scene to train neural networks in the first $\num[group-separator={,}]{10000}$ steps. On CLEVR-M1 and CLEVR-M2, the proposed method is trained from scratch for $\num[group-separator={,}]{150000}$ steps, even though the relatively large range of azimuth (i.e., $[0, \pi]$) makes the fully-unsupervised learning difficult. On CLEVR-M3/CLEVR-M4 in which the azimuth is sampled from the full range $[0, 2 \pi]$, we have found it beneficial to adopt a curriculum learning strategy that first pretrains the model on the simpler CLEVR-M1/CLEVR-M2 for $\num[group-separator={,}]{100000}$ steps and then continues to train on CLEVR-M3/CLEVR-M4 for $\num[group-separator={,}]{100000}$ steps. The choices of neural networks in both generative model and variational inference are described below. Instead of adopting a superior but more time-consuming method such as grid search, we manually choose the hyperparameters of neural networks based on experience.

\begin{itemize}
	\item $f_{\text{shp}}$ and $f_{\text{apc}}$ in the generative model are implemented as one convolutional neural network (CNN). The outputs of the CNN are split in the channel dimension into $1$ and $3$ for $f_{\text{shp}}$ and $f_{\text{apc}}$, respectively.
	\begin{itemize}
		\item Fully Connected, 4096 ReLU
		\item Fully Connected, 4096 ReLU
		\item Fully Connected, 8 $\times$ 8 $\times$ 128 ReLU
		\item 2x nearest-neighbor upsample; 5 $\times$ 5 Conv, 128 ReLU
		\item 5 $\times$ 5 Conv, 64 ReLU
		\item 2x nearest-neighbor upsample; 5 $\times$ 5 Conv, 64 ReLU
		\item 5 $\times$ 5 Conv, 32 ReLU
		\item 2x nearest-neighbor upsample; 5 $\times$ 5 Conv, 32 ReLU
		\item 3 $\times$ 3 Conv, 1 + 3 Linear
	\end{itemize}
	\item $f_{\text{bck}}$ in the generative model is a CNN.
	\begin{itemize}
		\item Fully Connected, 512 ReLU
		\item Fully Connected, 512 ReLU
		\item Fully Connected, 4 $\times$ 4 $\times$ 16 ReLU
		\item 4x nearest-neighbor upsample; 5 $\times$ 5 Conv, 16 ReLU
		\item 5 $\times$ 5 Conv, 16 ReLU
		\item 4x nearest-neighbor upsample; 5 $\times$ 5 Conv, 16 ReLU
		\item 3 $\times$ 3 Conv, 3 Linear
	\end{itemize}
	\item $f_{\text{ord}}$ in the generative model is a multi-layer perceptron (MLP).
	\begin{itemize}
		\item Fully Connected, 512 ReLU
		\item Fully Connected, 512 ReLU
		\item Fully Connected, 1 Linear
	\end{itemize}
	\item $g_{\text{feat}}$ in the variational inference is a CNN augmented with positional embedding.
	\begin{itemize}
		\item 5 $\times$ 5 Conv, 64 ReLU
		\item 5 $\times$ 5 Conv, 64 ReLU
		\item 5 $\times$ 5 Conv, 64 ReLU
		\item 5 $\times$ 5 Conv, 64 ReLU; Positional Embedding
		\item Layer Norm; Fully Connected, 64 ReLU
		\item Fully Connected, 64 Linear
	\end{itemize}
	\item $g_{\text{key}}$ in the variational inference is a linear layer with layer Normalization.
	\begin{itemize}
		\item Layer Norm; Fully Connected, 64 Linear
	\end{itemize}
	\item $g_{\text{qry}}$ in the variational inference is a linear layer with layer normalization.
	\begin{itemize}
		\item Layer Norm; Fully Connected, 64 Linear
	\end{itemize}
	\item $g_{\text{val}}$ in the variational inference is a linear layer with layer normalization.
	\begin{itemize}
		\item Layer Norm; Fully Connected, 136 Linear
	\end{itemize}
	\item $g_{\text{upd}}$ in the variational inference is a gated recurrent unit (GRU) followed by a residual MLP with layer normalization, which independently updates $\boldsymbol{y}_{1:M,0:K}^{\text{full}}$ for each $m$ and $k$. Information of different viewpoints are integrated in the two average operations following $g_{\text{upd}}$. It is possible to apply a more complex and powerful neural network such as a graph neural network (GNN) to integrate information of different viewpoints earlier, and we leave the investigation in future work.
	\begin{itemize}
		\item GRU, 136 Tanh
		\item Layer Norm; Fully Connected, 128 ReLU
		\item Fully Connected, 8 + 128 Linear
	\end{itemize}
	\item $g_{\text{bck}}$ in the variational inference is an MLP.
	\begin{itemize}
		\item Fully Connected, 512 ReLU
		\item Fully Connected, 512 ReLU
		\item Fully Connected, 8 + 8 Linear
	\end{itemize}
	\item $g_{\text{obj}}$ in the variational inference is an MLP.
	\begin{itemize}
		\item Fully Connected, 512 ReLU
		\item Fully Connected, 512 ReLU
		\item Fully Connected, 64 + 64 + 2 + 1 Linear
	\end{itemize}
	\item $g_{\text{view}}$ in the variational inference is an MLP.
	\begin{itemize}
		\item Fully Connected, 512 ReLU
		\item Fully Connected, 512 ReLU
		\item Fully Connected, 4 + 4 Linear
	\end{itemize}
	\item The neural network used by NVIL is a CNN.
	\begin{itemize}
		\item 3 $\times$ 3 Conv, 16 ReLU
		\item 3 $\times$ 3 Conv, stride 2, 16 ReLU
		\item 3 $\times$ 3 Conv, 32 ReLU
		\item 3 $\times$ 3 Conv, stride 2, 32 ReLU
		\item 3 $\times$ 3 Conv, 64 ReLU
		\item 3 $\times$ 3 Conv, stride 2, 64 ReLU
		\item Fully Connected, 256 ReLU
		\item Fully Connected, 1 Linear
	\end{itemize}
\end{itemize}

\subsubsection{Baseline Method}

The baseline method which is derived from the proposed method uses the same set of hyperparameters, and differs from the proposed method in two aspects. In the generative model, the viewpoint-independent latent variables $\boldsymbol{z}_{k}^{\text{attr}}$, $\rho_{k}$, and $z_{k}^{\text{prs}}$ are replaced with viewpoint-dependent versions $\boldsymbol{z}_{m,k}^{\text{attr}}$, $\rho_{m,k}$, and $z_{m,k}^{\text{prs}}$. In the variational inference, the lines $3$, $4$, $7$, $11$, $12$, $18$, $19$, $20$ in Algorithm 1 are replaced with the following expressions.
\begin{align*}
\text{line} \, 3\!: \,\, & \boldsymbol{y}_{m,k}^{\text{view}} \sim \mathcal{N}(\hat{\boldsymbol{\mu}}^{\text{view}}\!\!, \diag(\hat{\boldsymbol{\sigma}}^{\text{view}})) \\
\text{line} \, 4\!: \,\, & \boldsymbol{y}_{m,k}^{\text{attr}} \sim \mathcal{N}(\hat{\boldsymbol{\mu}}^{\text{attr}}, \diag(\hat{\boldsymbol{\sigma}}^{\text{attr}})) \\
\text{line} \, 7\!: \,\, & \boldsymbol{y}_{m,k}^{\text{full}} \gets [\boldsymbol{y}_{m,k}^{\text{view}}, \boldsymbol{y}_{m,k}^{\text{attr}}] \\
\text{line} \, 11\!: \,\, & \boldsymbol{y}_{m,k}^{\text{view}} \gets \boldsymbol{v}_{m,k}^{\text{view}} \\
\text{line} \, 12\!: \,\, & \boldsymbol{y}_{m,k}^{\text{attr}} \gets \boldsymbol{v}_{m,k}^{\text{attr}} \\
\text{line} \, 18\!: \,\, & \boldsymbol{\mu}_{m,0}^{\text{attr}}, \boldsymbol{\sigma}_{m,0}^{\text{attr}} \gets g_{\text{bck}}(\boldsymbol{y}_{m,0}^{\text{attr}}) \\
\text{line} \, 19\!: \,\, & \boldsymbol{\mu}_{m,k}^{\text{attr}}, \boldsymbol{\sigma}_{m,k}^{\text{attr}}, \boldsymbol{\tau}_{m,k}, \kappa_{m,k} \gets g_{\text{obj}}(\boldsymbol{y}_{m,k}^{\text{attr}}) \\
\text{line} \, 20\!: \,\, & \boldsymbol{\mu}_{m,k}^{\text{view}}, \boldsymbol{\sigma}_{m,k}^{\text{view}} \gets g_{\text{view}}(\boldsymbol{y}_{m,k}^{\text{view}})
\end{align*}

\subsubsection{MulMON} MulMON is trained with the default hyperparameters described in the ``scripts/train\_clevr\_parallel.sh'' file of the official code repository\footnote{\url{https://github.com/NanboLi/MulMON}}
except: 1) the number of training steps is $\num[group-separator={,}]{600000}$; 2) the number of viewpoints for inference is sampled from $n \!\sim\! \mathcal{U}(1, 3)$ and the number of viewpoints for query is $4 - n$; 3) the number of slots $K \!+\! 1$ is $8$ during training.

\subsection{Single-Viewpoint Learning}

\subsubsection{Proposed Method} In the generative model, the standard deviation $\sigma_{\text{x}}$ of the likelihood function is $0.2$. The maximum number $K$ of objects that may appear in the visual scene is $6$, $5$, and $7$ on the dSprites, Abstract, and CLEVR datasets, respectively. The respective dimensionalities of $\boldsymbol{z}_{m}^{\text{view}}$, $\boldsymbol{z}_{0}^{\text{attr}}$, and $\boldsymbol{z}_{k}^{\text{attr}}$ with $1 \!\leq\! k \!\leq\! K$, as well as the hyperparameter $\alpha$ are $1$/$1$/$1$, $4$/$4$/$8$, $32$/$32$/$64$, and $3.5$/$3.0$/$4.5$ on the dSprites/Abstract/CLEVR dataset. $\lambda$ is chosen to be $0.5$. In the inference, the dimensionalities $D_{\text{vw}}$, $D_{\text{at}}$, $D_{\text{key}}$, and $D_{\text{val}}$ are $1$, $64$, $64$, and $65$, respectively. $T$ is chosen to be $3$. In the learning, the batch size is $64$ and the number of training steps is $\num[group-separator={,}]{500000}$. The initial learning rate is $4 \times 10^{-4}$, and is decayed exponentially with a factor $0.5$ every $\num[group-separator={,}]{100000}$ steps. In the first $\num[group-separator={,}]{10000}$ training steps, the learning rate is multiplied by a factor that is increased linearly from $0$ to $1$. Hyperparameters of neural networks are described below.

\begin{itemize}
	\item $f_{\text{shp}}$ and $f_{\text{apc}}$ on the dSprites and Abstract datasets.
	\begin{itemize}
		\item 64 $\times$ 64 spatial broadcast; Positional Embedding
		\item 5 $\times$ 5 Conv, 32 ReLU
		\item 5 $\times$ 5 Conv, 32 ReLU
		\item 5 $\times$ 5 Conv, 32 ReLU
		\item 3 $\times$ 3 Conv, 1 + 3 Linear
	\end{itemize}
	\item $f_{\text{shp}}$ and $f_{\text{apc}}$ on the CLEVR dataset.
	\begin{itemize}
		\item 8 $\times$ 8 spatial broadcast; Positional Embedding
		\item 5 $\times$ 5 Trans Conv, stride 2, 64 ReLU
		\item 5 $\times$ 5 Trans Conv, stride 2, 64 ReLU
		\item 5 $\times$ 5 Trans Conv, stride 2, 64 ReLU
		\item 5 $\times$ 5 Trans Conv, stride 2, 64 ReLU
		\item 5 $\times$ 5 Conv, 64 ReLU
		\item 3 $\times$ 3 Conv, 1 + 3 Linear
	\end{itemize}
	\item $f_{\text{bck}}$ on the dSprites and Abstract datasets.
	\begin{itemize}
		\item 64 $\times$ 64 spatial broadcast; Positional Embedding
		\item 5 $\times$ 5 Conv, 8 ReLU
		\item 5 $\times$ 5 Conv, 8 ReLU
		\item 3 $\times$ 3 Conv, 3 Linear
	\end{itemize}
	\item $f_{\text{bck}}$ on the CLEVR dataset.
	\begin{itemize}
		\item 32 $\times$ 32 spatial broadcast; Positional Embedding
		\item 5 $\times$ 5 Trans Conv, stride 2, 16 ReLU
		\item 5 $\times$ 5 Trans Conv, stride 2, 16 ReLU
		\item 5 $\times$ 5 Conv, 16 ReLU
		\item 3 $\times$ 3 Conv, 3 Linear
	\end{itemize}
	\item $f_{\text{ord}}$ on all the datasets.
	\begin{itemize}
		\item Fully Connected, 512 ReLU
		\item Fully Connected, 512 ReLU
		\item Fully Connected, 1 Linear
	\end{itemize}
	\item $g_{\text{feat}}$ on the dSprites and Abstract datasets.
	\begin{itemize}
		\item 5 $\times$ 5 Conv, 32 ReLU
		\item 5 $\times$ 5 Conv, 32 ReLU
		\item 5 $\times$ 5 Conv, 32 ReLU
		\item 5 $\times$ 5 Conv, 32 ReLU; Positional Embedding
		\item Layer Norm; Fully Connected, 32 ReLU
		\item Fully Connected, 32 Linear
	\end{itemize}
	\item $g_{\text{feat}}$ on the CLEVR dataset.
	\begin{itemize}
		\item 5 $\times$ 5 Conv, 64 ReLU
		\item 5 $\times$ 5 Conv, 64 ReLU
		\item 5 $\times$ 5 Conv, 64 ReLU
		\item 5 $\times$ 5 Conv, 64 ReLU; Positional Embedding
		\item Layer Norm; Fully Connected, 64 ReLU
		\item Fully Connected, 64 Linear
	\end{itemize}
	\item $g_{\text{key}}$ on all the datasets.
	\begin{itemize}
		\item Layer Norm; Fully Connected, 64 Linear
	\end{itemize}
	\item $g_{\text{qry}}$ on all the datasets.
	\begin{itemize}
		\item Layer Norm; Fully Connected, 64 Linear
	\end{itemize}
	\item $g_{\text{val}}$ on all the datasets.
	\begin{itemize}
		\item Layer Norm; Fully Connected, 65 Linear
	\end{itemize}
	\item $g_{\text{upd}}$ on all the datasets.
	\begin{itemize}
		\item GRU, 65 Tanh
		\item Layer Norm; Fully Connected, 128 ReLU
		\item Fully Connected, 1 + 64 Linear
	\end{itemize}
	\item $g_{\text{bck}}$ on the dSprites and Abstract datasets.
	\begin{itemize}
		\item Fully Connected, 512 ReLU
		\item Fully Connected, 512 ReLU
		\item Fully Connected, 4 + 4 Linear
	\end{itemize}
	\item $g_{\text{bck}}$ on the CLEVR dataset.
	\begin{itemize}
		\item Fully Connected, 512 ReLU
		\item Fully Connected, 512 ReLU
		\item Fully Connected, 8 + 8 Linear
	\end{itemize}
	\item $g_{\text{obj}}$ on the dSprites and Abstract datasets.
	\begin{itemize}
		\item Fully Connected, 512 ReLU
		\item Fully Connected, 512 ReLU
		\item Fully Connected, 32 + 32 + 2 + 1 Linear
	\end{itemize}
	\item $g_{\text{obj}}$ on the CLEVR dataset.
	\begin{itemize}
		\item Fully Connected, 512 ReLU
		\item Fully Connected, 512 ReLU
		\item Fully Connected, 64 + 64 + 2 + 1 Linear
	\end{itemize}
	\item $g_{\text{view}}$ on all the datasets.
	\begin{itemize}
		\item Fully Connected, 512 ReLU
		\item Fully Connected, 512 ReLU
		\item Fully Connected, 1 + 1 Linear
	\end{itemize}
	\item The neural network used by NVIL on all the datasets.
	\begin{itemize}
		\item 3 $\times$ 3 Conv, 16 ReLU
		\item 3 $\times$ 3 Conv, stride 2, 16 ReLU
		\item 3 $\times$ 3 Conv, 32 ReLU
		\item 3 $\times$ 3 Conv, stride 2, 32 ReLU
		\item 3 $\times$ 3 Conv, 64 ReLU
		\item 3 $\times$ 3 Conv, stride 2, 64 ReLU
		\item Fully Connected, 256 ReLU
		\item Fully Connected, 1 Linear
	\end{itemize}
\end{itemize}

\subsubsection{Slot Attention} Slot Attention is trained with the default hyperparameters described in the official code repository\footnote{\url{https://github.com/google-research/google-research/tree/master/slot_attention}}
except: 1) the number of slots $K \!+\! 1$ during training is $7$, $6$, and $8$ on the dSprites, Abstract, and CLEVR datasets; 2) the hyperparameters of neural networks on the dSprites and Abstract datasets are chosen to be same as the default hyperparameters used for the Multi-dSprites dataset by Slot Attention, i.e., Tables 4 and 6 in the supplementary material of \cite{Locatello2020ObjectCentricLW}. 

\subsubsection{GMIOO} GMIOO is trained with the default hyperparameters described in the ``experiments/config.yaml'' file of the official code repository\footnote{\url{https://github.com/jinyangyuan/infinite-occluded-objects}}
except: 1) the number of training steps is $\num[group-separator={,}]{200000}$ and the batch size is $64$; 2) the upper bound $K$ of the number of objects during inference and the hyperparameter $\alpha$ are $6$/$5$/$7$ and $3.5$/$3.0$/$4.5$ on the dSprites/Abstract/CLEVR dataset; 3) the dimensionality of the background latent variable is $4$ on the Abstract and dSprites datasets; 4) the mean parameter of the prior distribution of bounding box scale is $-0.5$ on the dSprites and CLEVR datasets; 5) the standard deviation parameter of the prior distribution of bounding box scale is $0.5$ on all the datasets; 6) the glimpse size is $32$ on the dSprites and Abstract datasets and is $64$ on the CLEVR dataset; 7) some neural networks use different hyperparameters, which are described below.

\begin{itemize}
	\item The decoder of appearance on the Abstract dataset.
	\begin{itemize}
		\item Fully Connected, 256 ReLU
		\item Fully Connected, 8 $\times$ 8 $\times$ 32 ReLU
		\item 2x bilinear upsample; 3 $\times$ 3 Conv, 32 ReLU
		\item 3 $\times$ 3 Conv, 16 ReLU
		\item 2x bilinear upsample; 3 $\times$ 3 Conv, 16 ReLU
		\item 3 $\times$ 3 Conv, 3 Linear
	\end{itemize}
	\item The decoder of appearance on the CLEVR dataset.
	\begin{itemize}
		\item Fully Connected, 256 ReLU
		\item Fully Connected, 8 $\times$ 8 $\times$ 32 ReLU
		\item 2x bilinear upsample; 3 $\times$ 3 Conv, 32 ReLU
		\item 3 $\times$ 3 Conv, 16 ReLU
		\item 2x bilinear upsample; 3 $\times$ 3 Conv, 16 ReLU
		\item 3 $\times$ 3 Conv, 8 ReLU
		\item 2x bilinear upsample; 3 $\times$ 3 Conv, 8 ReLU
		\item 3 $\times$ 3 Conv, 3 Linear
	\end{itemize}
	\item The decoder of background on the CLEVR dataset.
	\begin{itemize}
		\item Fully Connected, 2 $\times$ 2 $\times$ 8 ReLU
		\item 2x bilinear upsample; 3 $\times$ 3 Conv, 8 ReLU
		\item 2x bilinear upsample; 3 $\times$ 3 Conv, 8 ReLU
		\item 2x bilinear upsample; 3 $\times$ 3 Conv, 8 ReLU
		\item 2x bilinear upsample; 3 $\times$ 3 Conv, 8 ReLU
		\item 2x bilinear upsample; 3 $\times$ 3 Conv, 3 linear
	\end{itemize}
	\item The decoder of shape on the dSprites and Abstract datasets.
	\begin{itemize}
		\item Fully Connected, 256 ReLU
		\item Fully Connected, 8 $\times$ 8 $\times$ 32 ReLU
		\item 2x bilinear upsample; 3 $\times$ 3 Conv, 32 ReLU
		\item 3 $\times$ 3 Conv, 16 ReLU
		\item 2x bilinear upsample; 3 $\times$ 3 Conv, 16 ReLU
		\item 3 $\times$ 3 Conv, 1 Linear
	\end{itemize}
	\item The decoder of shape on the CLEVR dataset.
	\begin{itemize}
		\item Fully Connected, 256 ReLU
		\item Fully Connected, 8 $\times$ 8 $\times$ 32 ReLU
		\item 2x bilinear upsample; 3 $\times$ 3 Conv, 32 ReLU
		\item 3 $\times$ 3 Conv, 16 ReLU
		\item 2x bilinear upsample; 3 $\times$ 3 Conv, 16 ReLU
		\item 3 $\times$ 3 Conv, 8 ReLU
		\item 2x bilinear upsample; 3 $\times$ 3 Conv, 8 ReLU
		\item 3 $\times$ 3 Conv, 1 Linear
	\end{itemize}
	\item The convolutional parts of the neural networks used to initialize and update latent variables of background on the CLEVR dataset.
	\begin{itemize}
		\item 3 $\times$ 3 Conv, 4 ReLU
		\item 3 $\times$ 3 Conv, stride 2, 4 ReLU
		\item 3 $\times$ 3 Conv, 8 ReLU
		\item 3 $\times$ 3 Conv, stride 2, 8 ReLU
		\item 3 $\times$ 3 Conv, 16 ReLU
		\item 3 $\times$ 3 Conv, stride 2, 16 ReLU
	\end{itemize}
	\item The convolutional parts of the neural networks used to initialize and update latent variables of objects on the CLEVR dataset.
	\begin{itemize}
		\item 3 $\times$ 3 Conv, 8 ReLU
		\item 3 $\times$ 3 Conv, stride 2, 8 ReLU
		\item 3 $\times$ 3 Conv, 16 ReLU
		\item 3 $\times$ 3 Conv, stride 2, 16 ReLU
		\item 3 $\times$ 3 Conv, 32 ReLU
		\item 3 $\times$ 3 Conv, stride 2, 32 ReLU
	\end{itemize}
\end{itemize}

\subsubsection{SPACE} SPACE is trained with the default hyperparameters described in the ``src/configs/3d\_room\_small.yaml'' file of the official code repository\footnote{\url{https://github.com/zhixuan-lin/SPACE}}
except: 1) the number of training steps is $\num[group-separator={,}]{200000}$; 2) the number of background components is $1$ and ``CompDecoder'' is used as the decoder of background on all the datasets; 3) the dimensionality of the background latent variable is $4$ on the Abstract dataset; 4) the mean parameter of the prior distribution of bounding box scale is decreased from $0$ to $-0.5$ on the dSprites and CLEVR datasets, and from $0.5$ to $0$ on the Abstract dataset; 5) the standard deviation parameter of the prior distribution of bounding box scale is $0.5$ on all the datasets; 6) the standard deviations of foreground and background on the Abstract dataset are $0.2$ and $0.1$, respectively; 7) the glimpse size is $64$ on the CLEVR dataset; 8) some neural networks use different hyperparameters, which are described below.

\begin{itemize}
	\item The convolutional part of the glimpse encoder on the CLEVR dataset.
	\begin{itemize}
		\item 3 $\times$ 3 Conv, 16 CELU; Group Norm
		\item 4 $\times$ 4 Conv, stride 2, 32 CELU; Group Norm
		\item 3 $\times$ 3 Conv, 32 CELU; Group Norm
		\item 4 $\times$ 4 Conv, stride 2, 64 CELU; Group Norm
		\item 4 $\times$ 4 Conv, stride 2, 128 CELU; Group Norm
		\item 8 $\times$ 8 Conv, 256 CELU; Group Norm
	\end{itemize}
	\item The convolutional part of the glimpse decoder on the CLEVR dataset.
	\begin{itemize}
		\item 1 $\times$ 1 Conv, 256 CELU; Group Norm
		\item 1 $\times$ 1 Conv, 128 $\times$ 4 $\times$ 4 CELU
		\item 4x pixel shuffle; Group Norm
		\item 3 $\times$ 3 Conv, 128 CELU; Group Norm
		\item 1 $\times$ 1 Conv, 128 $\times$ 2 $\times$ 2 CELU
		\item 2x pixel shuffle; Group Norm
		\item 3 $\times$ 3 Conv, 128 CELU; Group Norm
		\item 1 $\times$ 1 Conv, 64 $\times$ 2 $\times$ 2 CELU
		\item 2x pixel shuffle; Group Norm
		\item 3 $\times$ 3 Conv, 64 CELU; Group Norm
		\item 1 $\times$ 1 Conv, 32 $\times$ 2 $\times$ 2 CELU
		\item 2x pixel shuffle; Group Norm
		\item 3 $\times$ 3 Conv, 32 CELU; Group Norm
		\item 1 $\times$ 1 Conv, 16 $\times$ 2 $\times$ 2 CELU
		\item 2x pixel shuffle; Group Norm
		\item 3 $\times$ 3 Conv, 16 CELU; Group Norm
	\end{itemize}
\end{itemize}

\section{Computing Infrastructure}

Experiments are conducted on a server with Intel Xeon CPU E5-2678 v3 CPUs, NVIDIA GeForce
RTX 2080 Ti GPUs, 256G memory, and Ubuntu 20.04 operating
system. The code is developed based on the PyTorch framework \cite{Paszke2019PyTorchAI} with version 1.8. In all the experiments, the random seeds are sampled randomly using the built-in python function ``random.randint(0, 0xffffffff)''. On the CLEVR-M, Abstract, and dSprites datasets, the proposed method can be trained and tested with one NVIDIA GeForce
RTX 2080 Ti GPU. On the CLEVR dataset, at least two NVIDIA GeForce
RTX 2080 Ti GPUs are needed.

\section{Extra Experimental Results}

\subsection{Multi-Viewpoint Learning}

\subsubsection{Scene Decomposition}

\begin{figure*}[t]
	\centering
	\includegraphics[width=2.1\columnwidth]{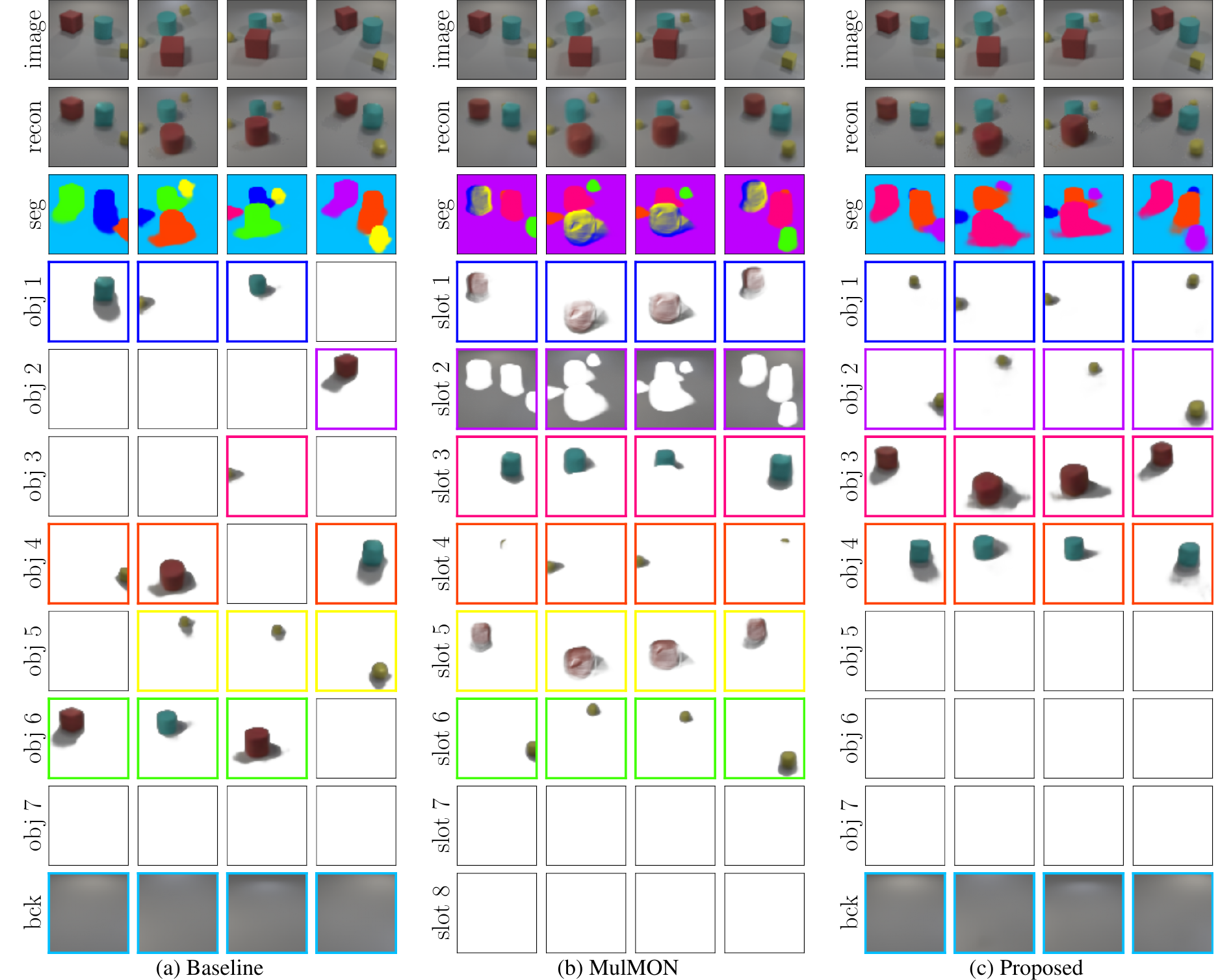}
	\caption{Qualitative comparison on the CLEVR-M1 dataset.}
	\label{fig-supp:multi_1}
\end{figure*}

\begin{figure*}[t]
	\centering
	\includegraphics[width=2.1\columnwidth]{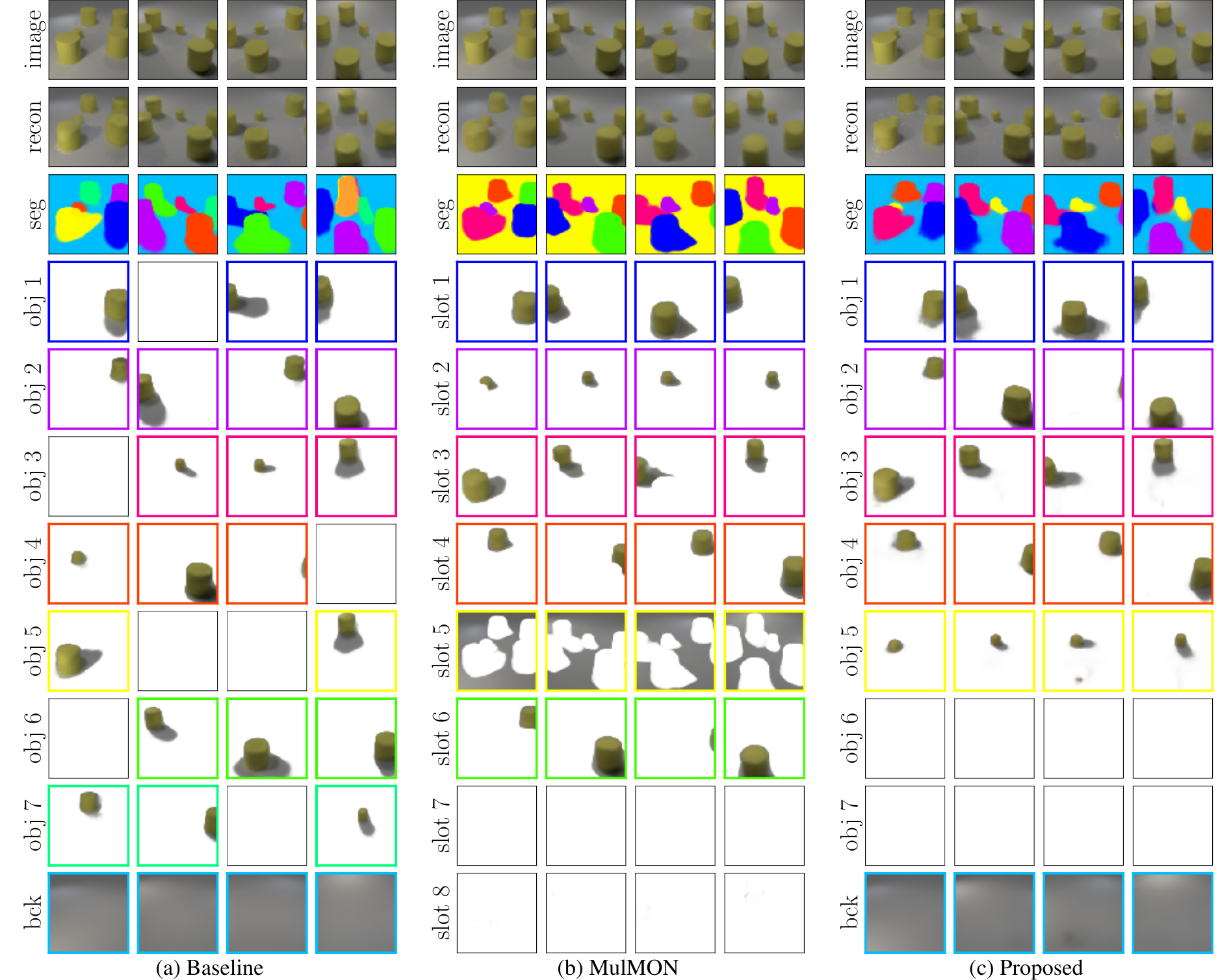}
	\caption{Qualitative comparison on the CLEVR-M2 dataset.}
	\label{fig-supp:multi_2}
\end{figure*}

\begin{figure*}[t]
	\centering
	\includegraphics[width=2.1\columnwidth]{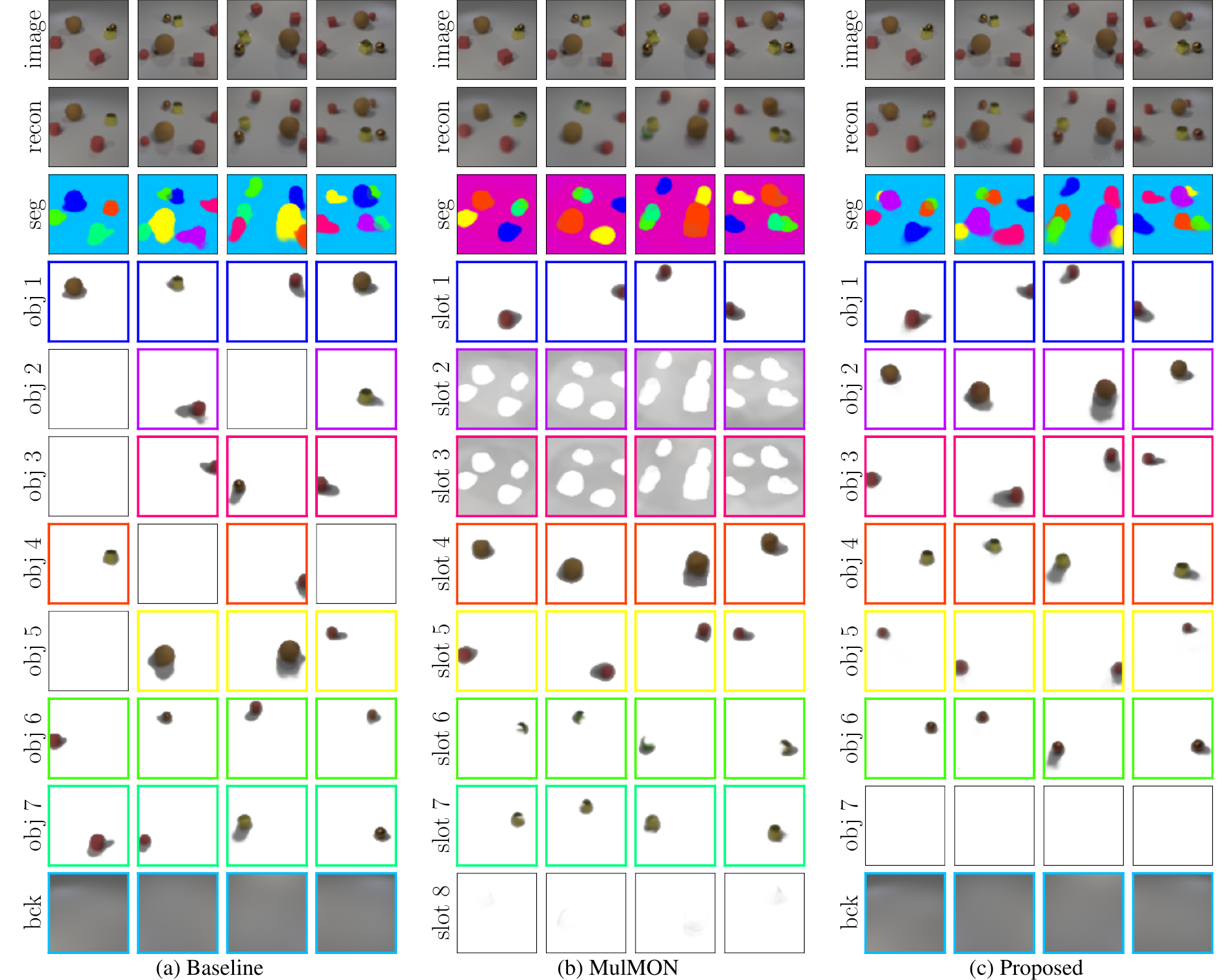}
	\caption{Qualitative comparison on the CLEVR-M3 dataset.}
	\label{fig-supp:multi_3}
\end{figure*}

\begin{figure*}[t]
	\centering
	\includegraphics[width=2.1\columnwidth]{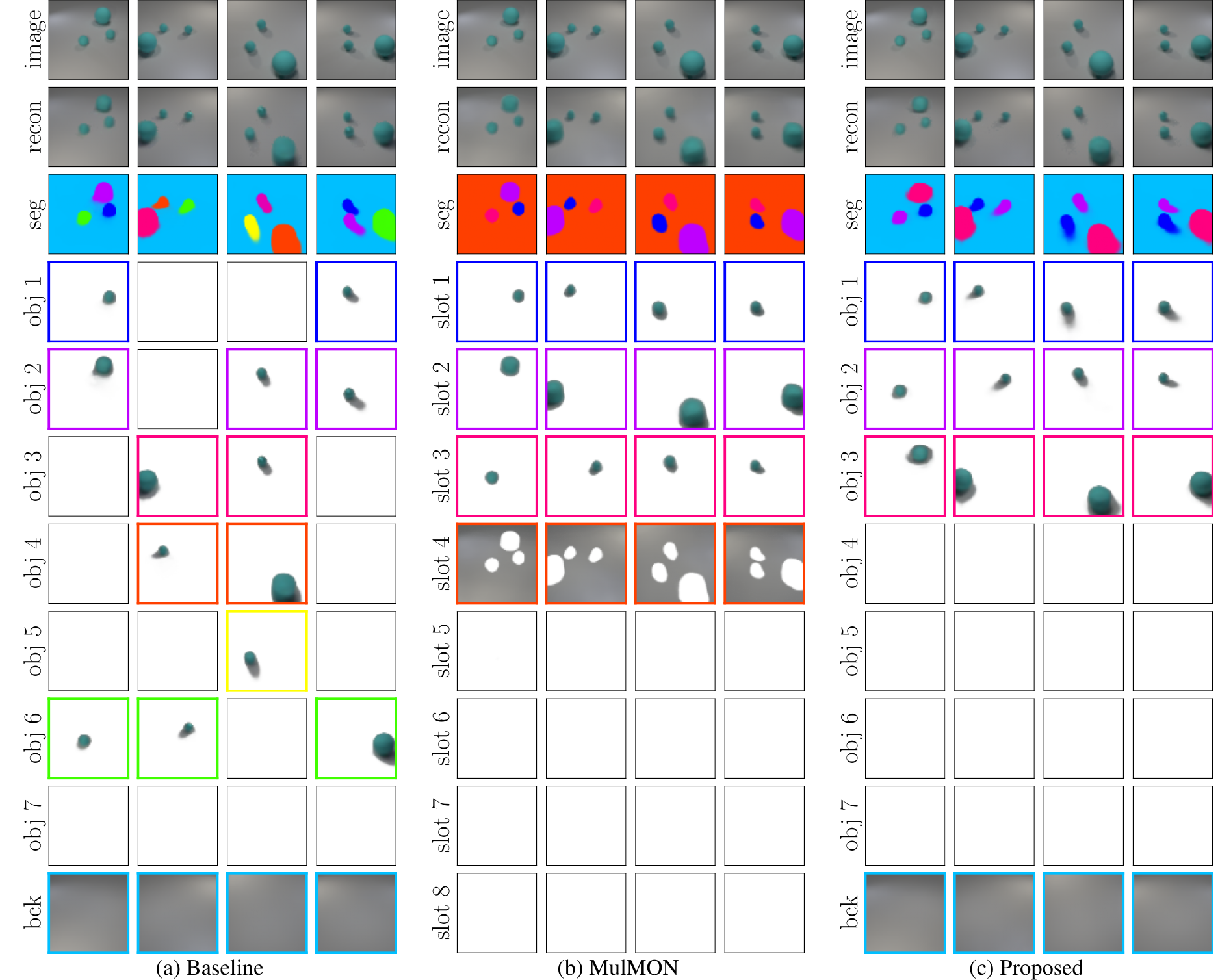}
	\caption{Qualitative comparison on the CLEVR-M4 dataset.}
	\label{fig-supp:multi_4}
\end{figure*}

The qualitative comparison of the baseline method, MulMON, and the proposed method on the CLEVR-M1 to CLEVR-M4 datasets is shown in Figures \ref{fig-supp:multi_1}, \ref{fig-supp:multi_2}, \ref{fig-supp:multi_3}, and \ref{fig-supp:multi_4}, respectively. All the methods tend to estimate shadows as parts of objects. The baseline method is not able to accurately identify the same object across viewpoints, while MulMON and the proposed method are able to achieve object constancy relatively well. The baseline method and the proposed method explicitly model the varying number of objects and distinguish background from objects, while MulMON does not have such abilities. However, it is still possible to estimate the number of objects based on the scene decomposition results of MulMON, in a reasonable though heuristic way. More specifically, let $\boldsymbol{r} \!\in\! {\{0, 1\}^{M \!\times\! N \!\times\! (K + 1)}}$ be the estimated pixel-wise partition of $K \!+\! 1$ slots in $M$ viewpoints. Whether the visual entity represented by the $k$th slot is included in the visual scene can be computed by $\max_{m}{\max_{n}{\,r_{m,n,k}}}$, and the estimated number of objects $\tilde{K}$ is
\begin{equation*}
\tilde{K} = \sum\nolimits_{k=0}^{K}{\Big(\max_{m}{\max_{n}{\,r_{m,n,k}}}\Big)} - 1
\end{equation*}
The proposed method significantly outperforms the baseline that randomly guesses the identities of objects, in all the evaluation metrics except ARI-A. The possible reason is that the baseline is derived from the proposed method and outputs similar background region estimations, which dominates the computation of ARI-A under the circumstance that shadows are incorrectly estimated as objects. Compared with MulMON, the ARI-A and AMI-A scores of the proposed method are slightly lower, and the rest scores are competitive or slightly better.

\subsubsection{Generalizability}

The quantitative results of generalizing the trained models to different number of objects and different number of viewpoints are presented in Tables \ref{tab-supp:multi_1_1}, \ref{tab-supp:multi_1_2}, \ref{tab-supp:multi_2_1}, \ref{tab-supp:multi_2_2}, \ref{tab-supp:multi_4_2}, \ref{tab-supp:multi_8_1}, and \ref{tab-supp:multi_8_2}. Generally speaking, both MulMON and the proposed method perform reasonably well when visual scenes contain more objects and the number of viewpoints varies compared with the ones used for training. MulMON generalizes better than the proposed method when the number of objects is increased (Tables \ref{tab-supp:multi_1_2}, \ref{tab-supp:multi_2_2}, \ref{tab-supp:multi_4_2}, and \ref{tab-supp:multi_8_2}), possibly because MulMON adopts a more powerful but time-consuming inference method, i.e., iterative amortized inference, which iteratively refines parameters of the variational distribution based on the gradients of loss function, while the proposed method does not exploit the information of gradient during inference. The baseline method performs best when the number of viewpoints is $1$ when testing (Tables \ref{tab-supp:multi_1_1} and \ref{tab-supp:multi_1_2}). The possible reason is that the baseline method treats images observed from multiple viewpoints of the same visual scene as images observed from a single viewpoint of different visual scenes, and thus the model does not need to learn how to maintain identities of objects across viewpoints and can better focus on acquiring information of the visual scene from a single viewpoint. When visual scenes are observed from more than one viewpoint, the baseline method does not perform well, while MulMON and the proposed method are both effective.

\subsubsection{Viewpoint Estimation}

Both the viewpoint representations and viewpoint-independent object-centric representations are inferred simultaneously during training, and the model is able to maintain the consistency between these two parts (e.g., using the same global coordinate system to represent both parts). When testing the models, it is possible to only estimate the viewpoints of images, under the condition that the viewpoint-independent attributes of objects and background are known. More specifically, given the intermediate variables $\boldsymbol{y}_{0:K}^{\text{attr}}$ that fully characterize the approximate posteriors of object-centric representations, the proposed method is able to infer the corresponding viewpoint representations of different observations of the same visual scene, while achieving the consistency between the inferred viewpoint representations and the given object-centric representations. This kind of inference can be derived from Algorithm 1 by initializing $\boldsymbol{y}_{k}^{\text{attr}}$ with the given representation instead of sampling from a normal distribution in line 4, and not executing the update operation in line 12. The viewpoints of $M_2 = 4$ images are estimated given the intermediate variables $\boldsymbol{y}_{0:K}^{\text{attr}}$ that are estimated on $M_1 = 1, 2, 4$ images. The viewpoints of the $M_1 + M_2$ images are all different. Experimental results on the Test 1 and Test 2 splits are shown in Tables \ref{tab-supp:cond_1} and \ref{tab-supp:cond_2}, respectively. Generally speaking, the estimated viewpoint representations are consistent with the given object-centric representations, and the scene decomposition performance increases as more viewpoints are used to extract object-centric representations.

\subsection{Single-Viewpoint Learning}

\begin{figure*}[t]
	\centering
	\includegraphics[width=2.1\columnwidth]{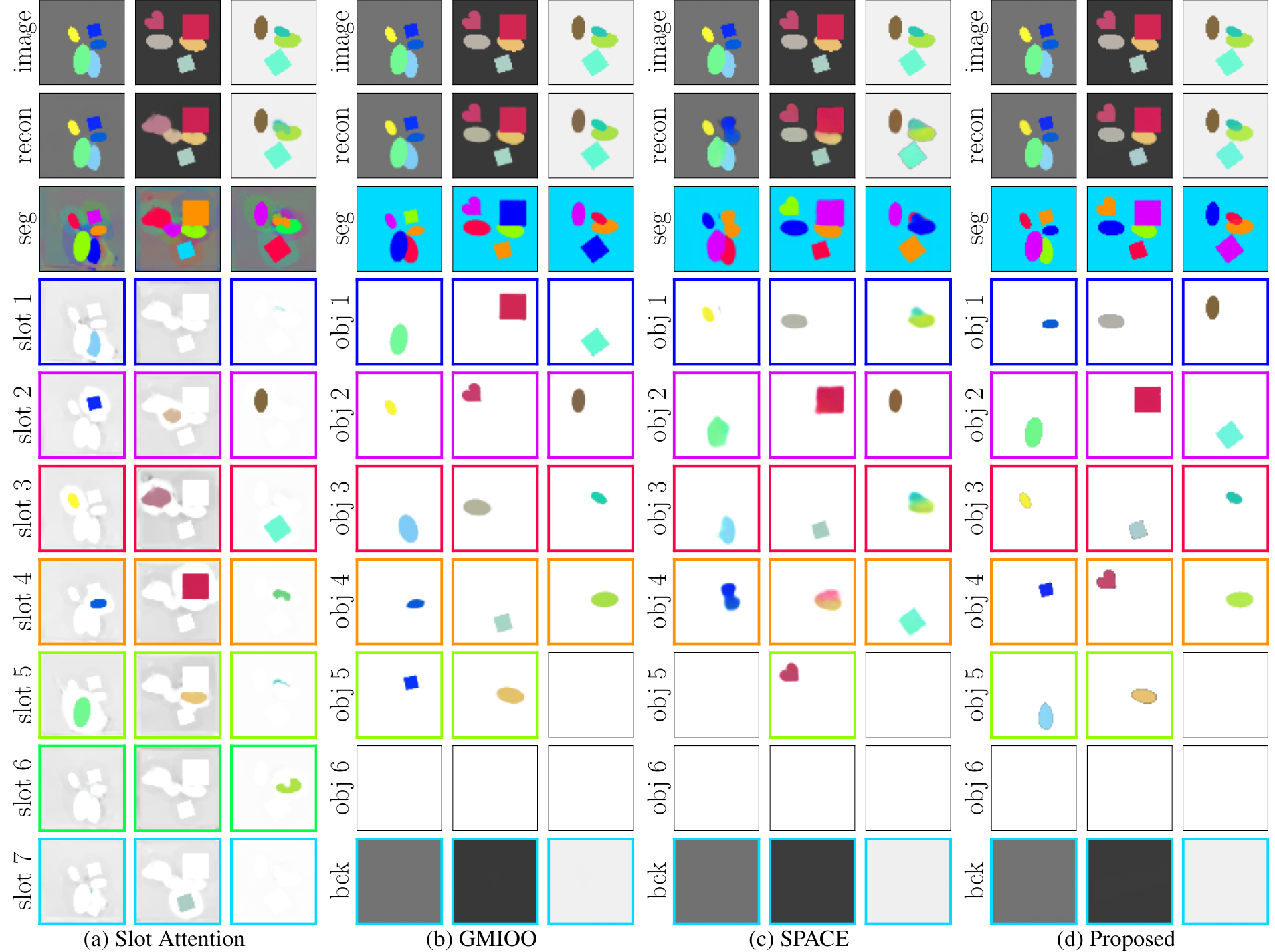}
	\caption{Qualitative comparison on the dSprites dataset.}
	\label{fig-supp:dsprites}
\end{figure*}

\begin{figure*}[t]
	\centering
	\includegraphics[width=2.1\columnwidth]{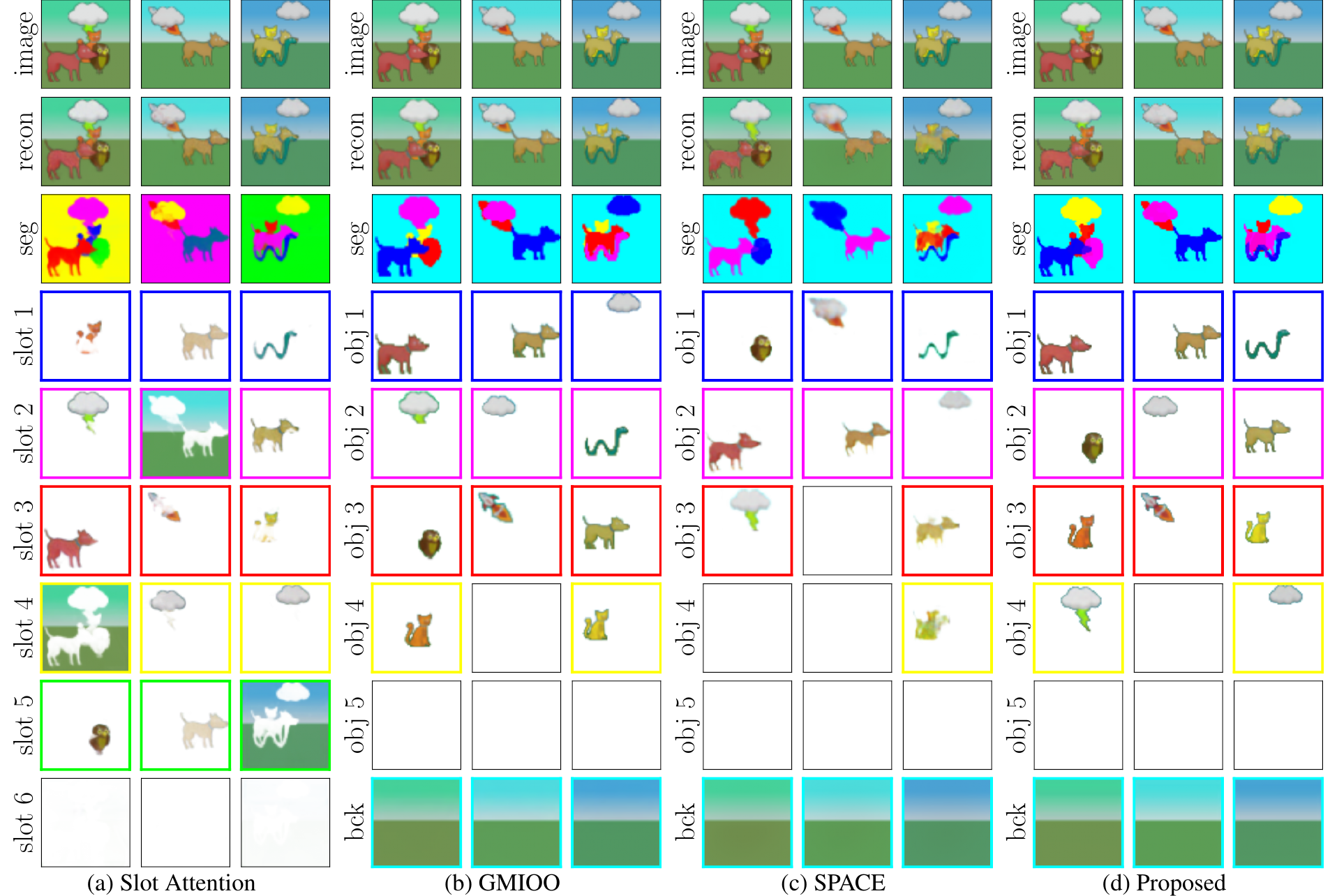}
	\caption{Qualitative comparison on the Abstract dataset.}
	\label{fig-supp:abstract}
\end{figure*}

\begin{figure*}[t]
	\centering
	\includegraphics[width=2.1\columnwidth]{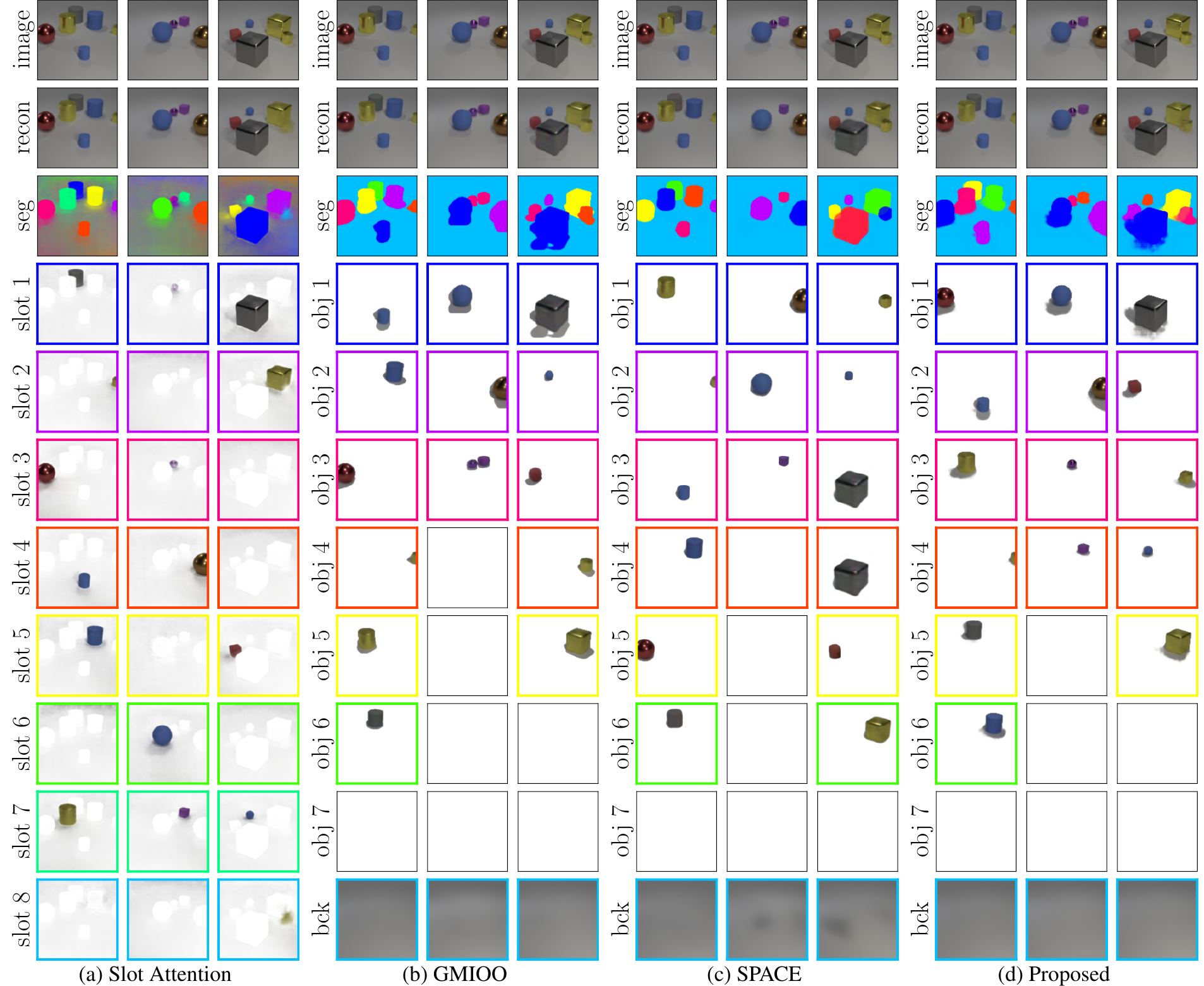}
	\caption{Qualitative comparison on the CLEVR dataset.}
	\label{fig-supp:clevr}
\end{figure*}

The qualitative comparison of Slot Attention, GMIOO, SPACE, and the proposed method on the dSprites, Abstract, and CLEVER datasets is shown in Figures \ref{fig-supp:dsprites}, \ref{fig-supp:abstract}, and \ref{fig-supp:clevr}, respectively. Slot Attention does not distinguish background from objects when modeling visual scenes. Therefore, there is no nature way to determine which slot corresponds to background. The trained model is able to use one random slot to represent background on the Abstract dataset, but fails to do so on the dSprites and CLEVR datasets. GMIOO and the proposed method work well even when objects are heavily occluded on the Abstract dataset. However, GMIOO tends to group nearby small objects as a single object on the CLEVR dataset. SPACE performs well on the CLEVR dataset, but tends to group multiple nearby or occluded objects as one object on the dSprites and Abstract datasets. On the dSprites dataset, GMIOO in general achieves the best performance, possibly because the inferred latent representations are iteratively refined based on both the observed and reconstructed images, which is beneficial when objects are rich in diversity. The performance of the proposed method on this dataset is only slightly lower than GMIOO. On the Abstract dataset, the proposed method in general performs best. The ARI-A and AMI-A scores achieved by Slot Attention is significantly higher than the rest methods, mainly because objects and background are not separately modeled, which leads to better estimations in the boundary regions of objects. GMIOO, SPACE and the proposed method tend to consider the background pixels in the boundary regions as object pixels, possibly because these pixels can be better reconstructed using the object decoders that have higher model capacities than the background decoders.  On the CLEVR dataset, SPACE in general achieves the best results. The possible reason is that SPACE adopts a two-stage framework to infer latent variables, and the attention mechanism with the spatial invariant property in the first stage acts as a strong inductive bias to guide the discovery of objects and to correctly estimate shadows of objects as background. The proposed method can better distinguish different objects from one another than SPACE, but achieves lower ARI-A, AMI-A, IoU, and F1 scores mainly because of the incorrect estimation of shadows. The performance of generalizing the trained models to images containing more numbers of objects is shown in Table \ref{tab-supp:single}. The proposed method achieves reasonable performance. GMIOO and SPACE in general generalize best, possibly because these methods adopt two-stage frameworks that first estimate bounding boxes of objects and then infer latent variables based on the cropped images in the bounding boxes. More specifically, the estimations of bounding boxes are more robust to the increase of number of objects and latent variables are easier to estimate given bounding boxes, which make GMIOO and SPACE more advantageous on images containing more objects than the ones used for training.

\begin{table*}[t]
	\centering
	\begin{small}
		\addtolength{\tabcolsep}{-2.8pt}
		\begin{tabular}{c|c|C{0.62in}C{0.62in}C{0.62in}C{0.62in}C{0.62in}C{0.62in}C{0.62in}C{0.62in}}
			\toprule
			Dataset          &  Method  &          ARI-A          &          AMI-A          &          ARI-O          &          AMI-O          &           IoU           &           F1            &           OCA           &           OOA           \\ \midrule
			\multirow{3}{*}{CLEVR-M1} & Baseline &     0.510$\pm$3e-3      &     0.502$\pm$2e-3      & \textbf{0.957}$\pm$4e-3 & \textbf{0.954}$\pm$3e-3 & \textbf{0.437}$\pm$3e-3 &     0.585$\pm$3e-3      & \textbf{0.688}$\pm$1e-2 &     0.853$\pm$5e-2      \\
			&  MulMON  & \textbf{0.605}$\pm$6e-3 & \textbf{0.555}$\pm$3e-3 &     0.919$\pm$4e-3      &     0.913$\pm$2e-3      &           N/A           &           N/A           &     0.410$\pm$4e-2      &           N/A           \\
			& Proposed &     0.496$\pm$2e-3      &     0.487$\pm$2e-3      &     0.951$\pm$3e-3      &     0.944$\pm$3e-3      &     0.435$\pm$2e-3      & \textbf{0.587}$\pm$3e-3 &     0.598$\pm$4e-2      & \textbf{0.890}$\pm$4e-2 \\ \midrule
			\multirow{3}{*}{CLEVR-M2} & Baseline &     0.508$\pm$2e-3      &     0.495$\pm$2e-3      & \textbf{0.959}$\pm$7e-3 & \textbf{0.954}$\pm$6e-3 & \textbf{0.434}$\pm$2e-3 & \textbf{0.583}$\pm$3e-3 & \textbf{0.666}$\pm$3e-2 &     0.872$\pm$3e-2      \\
			&  MulMON  & \textbf{0.596}$\pm$2e-3 & \textbf{0.544}$\pm$1e-3 &     0.931$\pm$4e-3      &     0.919$\pm$4e-3      &           N/A           &           N/A           &     0.462$\pm$6e-2      &           N/A           \\
			& Proposed &     0.495$\pm$2e-3      &     0.476$\pm$2e-3      &     0.937$\pm$3e-3      &     0.932$\pm$3e-3      &     0.418$\pm$3e-3      &     0.569$\pm$3e-3      &     0.608$\pm$4e-2      & \textbf{0.966}$\pm$1e-2 \\ \midrule
			\multirow{3}{*}{CLEVR-M3} & Baseline &     0.545$\pm$2e-3      &     0.522$\pm$2e-3      & \textbf{0.956}$\pm$2e-3 & \textbf{0.951}$\pm$2e-3 & \textbf{0.468}$\pm$3e-3 & \textbf{0.615}$\pm$4e-3 & \textbf{0.708}$\pm$4e-2 &     0.887$\pm$4e-2      \\
			&  MulMON  & \textbf{0.581}$\pm$9e-3 & \textbf{0.538}$\pm$4e-3 &     0.890$\pm$7e-3      &     0.885$\pm$6e-3      &           N/A           &           N/A           &     0.382$\pm$5e-2      &           N/A           \\
			& Proposed &     0.539$\pm$2e-3      &     0.505$\pm$2e-3      &     0.936$\pm$4e-3      &     0.930$\pm$4e-3      &     0.447$\pm$3e-3      &     0.592$\pm$4e-3      &     0.572$\pm$4e-2      & \textbf{0.937}$\pm$2e-2 \\ \midrule
			\multirow{3}{*}{CLEVR-M4} & Baseline &     0.532$\pm$2e-3      &     0.511$\pm$2e-3      & \textbf{0.959}$\pm$6e-3 & \textbf{0.961}$\pm$4e-3 & \textbf{0.460}$\pm$4e-3 & \textbf{0.612}$\pm$5e-3 & \textbf{0.760}$\pm$4e-2 & \textbf{0.859}$\pm$5e-2 \\
			&  MulMON  & \textbf{0.643}$\pm$3e-3 & \textbf{0.579}$\pm$3e-3 &     0.917$\pm$6e-3      &     0.911$\pm$4e-3      &           N/A           &           N/A           &     0.426$\pm$2e-2      &           N/A           \\
			& Proposed &     0.461$\pm$4e-3      &     0.445$\pm$2e-3      &     0.911$\pm$5e-3      &     0.916$\pm$3e-3      &     0.383$\pm$2e-3      &     0.529$\pm$2e-3      &     0.524$\pm$7e-2      &     0.855$\pm$4e-2      \\ \bottomrule
		\end{tabular}
	\end{small}
	\caption{Comparison of scene decomposition performance when learning from multiple viewpoints. All the methods are trained with $M \!=\! 4$ and $K \!=\! 7$, and tested on the Test 1 splits with $M \!=\! 1$ and $K \!=\! 7$.}
	\label{tab-supp:multi_1_1}
\end{table*}

\begin{table*}[t]
	\centering
	\begin{small}
		\addtolength{\tabcolsep}{-2.8pt}
		\begin{tabular}{c|c|C{0.62in}C{0.62in}C{0.62in}C{0.62in}C{0.62in}C{0.62in}C{0.62in}C{0.62in}}
			\toprule
			Dataset          &  Method  &          ARI-A          &          AMI-A          &          ARI-O          &          AMI-O          &           IoU           &           F1            &           OCA           &           OOA           \\ \midrule
			\multirow{3}{*}{CLEVR-M1} & Baseline &     0.358$\pm$1e-3      &     0.488$\pm$2e-3      & \textbf{0.912}$\pm$5e-3 & \textbf{0.920}$\pm$3e-3 &     0.356$\pm$5e-3      &     0.493$\pm$6e-3      &     0.300$\pm$3e-2      &     0.877$\pm$8e-3      \\
			&  MulMON  & \textbf{0.551}$\pm$5e-3 & \textbf{0.568}$\pm$3e-3 &     0.888$\pm$4e-3      &     0.887$\pm$2e-3      &           N/A           &           N/A           & \textbf{0.312}$\pm$3e-2 &           N/A           \\
			& Proposed &     0.367$\pm$9e-4      &     0.476$\pm$1e-3      &     0.877$\pm$3e-3      &     0.890$\pm$3e-3      & \textbf{0.359}$\pm$2e-3 & \textbf{0.496}$\pm$3e-3 &     0.212$\pm$3e-2      & \textbf{0.901}$\pm$9e-3 \\ \midrule
			\multirow{3}{*}{CLEVR-M2} & Baseline &     0.367$\pm$3e-3      &     0.502$\pm$2e-3      & \textbf{0.914}$\pm$1e-3 & \textbf{0.921}$\pm$2e-3 & \textbf{0.374}$\pm$4e-3 & \textbf{0.513}$\pm$5e-3 &     0.260$\pm$2e-2      &     0.855$\pm$2e-2      \\
			&  MulMON  & \textbf{0.516}$\pm$5e-3 & \textbf{0.555}$\pm$3e-3 &     0.877$\pm$5e-3      &     0.881$\pm$4e-3      &           N/A           &           N/A           & \textbf{0.286}$\pm$2e-2 &           N/A           \\
			& Proposed &     0.363$\pm$4e-3      &     0.477$\pm$2e-3      &     0.841$\pm$3e-3      &     0.865$\pm$2e-3      &     0.355$\pm$2e-3      &     0.490$\pm$3e-3      &     0.188$\pm$4e-2      & \textbf{0.891}$\pm$2e-2 \\ \midrule
			\multirow{3}{*}{CLEVR-M3} & Baseline &     0.325$\pm$3e-3      &     0.476$\pm$1e-3      & \textbf{0.913}$\pm$3e-3 & \textbf{0.919}$\pm$3e-3 & \textbf{0.355}$\pm$2e-3 & \textbf{0.492}$\pm$3e-3 & \textbf{0.280}$\pm$6e-2 &     0.856$\pm$3e-2      \\
			&  MulMON  & \textbf{0.513}$\pm$3e-3 & \textbf{0.533}$\pm$4e-3 &     0.820$\pm$1e-2      &     0.840$\pm$7e-3      &           N/A           &           N/A           &     0.200$\pm$3e-2      &           N/A           \\
			& Proposed &     0.359$\pm$3e-3      &     0.470$\pm$2e-3      &     0.864$\pm$3e-3      &     0.881$\pm$3e-3      &     0.351$\pm$3e-3      &     0.483$\pm$4e-3      &     0.204$\pm$6e-2      & \textbf{0.903}$\pm$7e-3 \\ \midrule
			\multirow{3}{*}{CLEVR-M4} & Baseline &     0.358$\pm$2e-3      &     0.495$\pm$1e-3      & \textbf{0.913}$\pm$4e-3 & \textbf{0.923}$\pm$3e-3 & \textbf{0.374}$\pm$3e-3 & \textbf{0.513}$\pm$4e-3 & \textbf{0.294}$\pm$4e-2 &     0.887$\pm$2e-2      \\
			&  MulMON  & \textbf{0.534}$\pm$5e-3 & \textbf{0.556}$\pm$3e-3 &     0.856$\pm$4e-3      &     0.867$\pm$3e-3      &           N/A           &           N/A           &     0.242$\pm$3e-2      &           N/A           \\
			& Proposed &     0.284$\pm$3e-3      &     0.431$\pm$2e-3      &     0.815$\pm$6e-3      &     0.853$\pm$3e-3      &     0.297$\pm$2e-3      &     0.423$\pm$2e-3      &     0.200$\pm$3e-2      & \textbf{0.928}$\pm$1e-2 \\ \bottomrule
		\end{tabular}
	\end{small}
	\caption{Comparison of scene decomposition performance when learning from multiple viewpoints. All the methods are trained with $M \!=\! 4$ and $K \!=\! 7$, and tested on the Test 2 splits with $M \!=\! 1$ and $K \!=\! 11$.}
	\label{tab-supp:multi_1_2}
\end{table*}

\begin{table*}[t]
	\centering
	\begin{small}
		\addtolength{\tabcolsep}{-2.8pt}
		\begin{tabular}{c|c|C{0.62in}C{0.62in}C{0.62in}C{0.62in}C{0.62in}C{0.62in}C{0.62in}C{0.62in}}
			\toprule
			Dataset          &  Method  &          ARI-A          &          AMI-A          &          ARI-O          &          AMI-O          &           IoU           &           F1            &           OCA           &           OOA           \\ \midrule
			\multirow{3}{*}{CLEVR-M1} & Baseline &     0.510$\pm$2e-3      &     0.431$\pm$2e-3      &     0.540$\pm$1e-2      &     0.669$\pm$7e-3      &     0.248$\pm$3e-3      &     0.376$\pm$4e-3      &     0.108$\pm$3e-2      &     0.721$\pm$7e-2      \\
			&  MulMON  & \textbf{0.609}$\pm$2e-3 & \textbf{0.558}$\pm$2e-3 &     0.930$\pm$7e-3      &     0.920$\pm$4e-3      &           N/A           &           N/A           &     0.424$\pm$3e-2      &           N/A           \\
			& Proposed &     0.502$\pm$7e-4      &     0.485$\pm$1e-3      & \textbf{0.951}$\pm$3e-3 & \textbf{0.940}$\pm$3e-3 & \textbf{0.440}$\pm$3e-3 & \textbf{0.598}$\pm$3e-3 & \textbf{0.686}$\pm$1e-2 & \textbf{0.957}$\pm$2e-2 \\ \midrule
			\multirow{3}{*}{CLEVR-M2} & Baseline &     0.502$\pm$5e-4      &     0.422$\pm$3e-4      &     0.529$\pm$7e-3      &     0.665$\pm$2e-3      &     0.240$\pm$4e-3      &     0.365$\pm$5e-3      &     0.106$\pm$2e-2      &     0.666$\pm$4e-2      \\
			&  MulMON  & \textbf{0.596}$\pm$1e-3 & \textbf{0.548}$\pm$7e-4 & \textbf{0.944}$\pm$3e-3 &     0.931$\pm$2e-3      &           N/A           &           N/A           &     0.512$\pm$2e-2      &           N/A           \\
			& Proposed &     0.499$\pm$2e-3      &     0.477$\pm$2e-3      &     0.943$\pm$3e-3      & \textbf{0.938}$\pm$2e-3 & \textbf{0.424}$\pm$3e-3 & \textbf{0.580}$\pm$4e-3 & \textbf{0.658}$\pm$5e-2 & \textbf{0.935}$\pm$3e-2 \\ \midrule
			\multirow{3}{*}{CLEVR-M3} & Baseline &     0.535$\pm$1e-3      &     0.445$\pm$2e-3      &     0.538$\pm$1e-2      &     0.668$\pm$7e-3      &     0.258$\pm$3e-3      &     0.389$\pm$3e-3      &     0.110$\pm$2e-2      &     0.665$\pm$3e-2      \\
			&  MulMON  & \textbf{0.594}$\pm$4e-3 & \textbf{0.552}$\pm$2e-3 &     0.924$\pm$7e-3      &     0.914$\pm$3e-3      &           N/A           &           N/A           &     0.378$\pm$5e-2      &           N/A           \\
			& Proposed &     0.533$\pm$2e-3      &     0.496$\pm$1e-3      & \textbf{0.935}$\pm$2e-3 & \textbf{0.926}$\pm$3e-3 & \textbf{0.453}$\pm$3e-3 & \textbf{0.607}$\pm$3e-3 & \textbf{0.622}$\pm$5e-2 & \textbf{0.949}$\pm$1e-2 \\ \midrule
			\multirow{3}{*}{CLEVR-M4} & Baseline &     0.524$\pm$2e-3      &     0.435$\pm$4e-3      &     0.534$\pm$3e-2      &     0.673$\pm$2e-2      &     0.249$\pm$6e-3      &     0.379$\pm$6e-3      &     0.090$\pm$3e-2      &     0.680$\pm$5e-2      \\
			&  MulMON  & \textbf{0.645}$\pm$8e-4 & \textbf{0.582}$\pm$7e-4 & \textbf{0.936}$\pm$3e-3 & \textbf{0.926}$\pm$2e-3 &           N/A           &           N/A           &     0.462$\pm$6e-2      &           N/A           \\
			& Proposed &     0.471$\pm$3e-3      &     0.451$\pm$2e-3      &     0.919$\pm$8e-3      &     0.921$\pm$5e-3      & \textbf{0.392}$\pm$2e-3 & \textbf{0.543}$\pm$2e-3 & \textbf{0.542}$\pm$5e-2 & \textbf{0.869}$\pm$4e-2 \\ \bottomrule
		\end{tabular}
	\end{small}
	\caption{Comparison of scene decomposition performance when learning from multiple viewpoints. All the methods are trained with $M \!=\! 4$ and $K \!=\! 7$, and tested on the Test 1 splits with $M \!=\! 2$ and $K \!=\! 7$.}
	\label{tab-supp:multi_2_1}
\end{table*}

\begin{table*}[t]
	\centering
	\begin{small}
		\addtolength{\tabcolsep}{-2.8pt}
		\begin{tabular}{c|c|C{0.62in}C{0.62in}C{0.62in}C{0.62in}C{0.62in}C{0.62in}C{0.62in}C{0.62in}}
			\toprule
			Dataset          &  Method  &          ARI-A          &          AMI-A          &          ARI-O          &          AMI-O          &           IoU           &           F1            &           OCA           &           OOA           \\ \midrule
			\multirow{3}{*}{CLEVR-M1} & Baseline &     0.342$\pm$1e-3      &     0.403$\pm$2e-3      &     0.529$\pm$5e-3      &     0.692$\pm$3e-3      &     0.201$\pm$3e-3      &     0.316$\pm$4e-3      &     0.060$\pm$6e-3      &     0.683$\pm$1e-2      \\
			&  MulMON  & \textbf{0.557}$\pm$2e-3 & \textbf{0.579}$\pm$1e-3 & \textbf{0.910}$\pm$2e-3 & \textbf{0.907}$\pm$1e-3 &           N/A           &           N/A           & \textbf{0.396}$\pm$4e-2 &           N/A           \\
			& Proposed &     0.362$\pm$3e-3      &     0.466$\pm$1e-3      &     0.858$\pm$6e-3      &     0.870$\pm$2e-3      & \textbf{0.358}$\pm$3e-3 & \textbf{0.498}$\pm$4e-3 &     0.210$\pm$3e-2      & \textbf{0.925}$\pm$9e-3 \\ \midrule
			\multirow{3}{*}{CLEVR-M2} & Baseline &     0.341$\pm$6e-4      &     0.408$\pm$1e-3      &     0.525$\pm$6e-3      &     0.688$\pm$3e-3      &     0.207$\pm$2e-3      &     0.323$\pm$2e-3      &     0.064$\pm$3e-2      &     0.673$\pm$2e-2      \\
			&  MulMON  & \textbf{0.526}$\pm$3e-3 & \textbf{0.566}$\pm$2e-3 & \textbf{0.891}$\pm$5e-3 & \textbf{0.896}$\pm$3e-3 &           N/A           &           N/A           & \textbf{0.420}$\pm$2e-2 &           N/A           \\
			& Proposed &     0.356$\pm$6e-4      &     0.463$\pm$2e-3      &     0.820$\pm$5e-3      &     0.847$\pm$4e-3      & \textbf{0.354}$\pm$3e-3 & \textbf{0.494}$\pm$4e-3 &     0.242$\pm$5e-2      & \textbf{0.905}$\pm$9e-3 \\ \midrule
			\multirow{3}{*}{CLEVR-M3} & Baseline &     0.314$\pm$3e-3      &     0.394$\pm$2e-3      &     0.531$\pm$9e-3      &     0.691$\pm$4e-3      &     0.203$\pm$2e-3      &     0.318$\pm$3e-3      &     0.070$\pm$1e-2      &     0.680$\pm$3e-2      \\
			&  MulMON  & \textbf{0.540}$\pm$2e-3 & \textbf{0.566}$\pm$8e-4 & \textbf{0.891}$\pm$3e-3 & \textbf{0.890}$\pm$1e-3 &           N/A           &           N/A           & \textbf{0.330}$\pm$5e-2 &           N/A           \\
			& Proposed &     0.357$\pm$3e-3      &     0.461$\pm$1e-3      &     0.845$\pm$6e-3      &     0.862$\pm$3e-3      & \textbf{0.356}$\pm$1e-3 & \textbf{0.494}$\pm$2e-3 &     0.220$\pm$5e-2      & \textbf{0.893}$\pm$1e-2 \\ \midrule
			\multirow{3}{*}{CLEVR-M4} & Baseline &     0.340$\pm$1e-3      &     0.409$\pm$2e-3      &     0.537$\pm$5e-3      &     0.700$\pm$4e-3      &     0.215$\pm$1e-3      &     0.334$\pm$2e-3      &     0.060$\pm$2e-2      &     0.720$\pm$1e-2      \\
			&  MulMON  & \textbf{0.551}$\pm$2e-3 & \textbf{0.579}$\pm$2e-3 & \textbf{0.901}$\pm$4e-3 & \textbf{0.903}$\pm$2e-3 &           N/A           &           N/A           & \textbf{0.426}$\pm$5e-2 &           N/A           \\
			& Proposed &     0.289$\pm$2e-3      &     0.425$\pm$2e-3      &     0.804$\pm$7e-3      &     0.841$\pm$4e-3      & \textbf{0.307}$\pm$2e-3 & \textbf{0.441}$\pm$3e-3 &     0.212$\pm$3e-2      & \textbf{0.908}$\pm$3e-3 \\ \bottomrule
		\end{tabular}
	\end{small}
	\caption{Comparison of scene decomposition performance when learning from multiple viewpoints. All the methods are trained with $M \!=\! 4$ and $K \!=\! 7$, and tested on the Test 2 splits with $M \!=\! 2$ and $K \!=\! 11$.}
	\label{tab-supp:multi_2_2}
\end{table*}

\begin{table*}[t]
	\centering
	\begin{small}
		\addtolength{\tabcolsep}{-2.8pt}
		\begin{tabular}{c|c|C{0.62in}C{0.62in}C{0.62in}C{0.62in}C{0.62in}C{0.62in}C{0.62in}C{0.62in}}
			\toprule
			Dataset          &  Method  &          ARI-A          &          AMI-A          &          ARI-O          &          AMI-O          &           IoU           &           F1            &           OCA           &           OOA           \\ \midrule
			\multirow{3}{*}{CLEVR-M1} & Baseline &     0.333$\pm$2e-3      &     0.318$\pm$5e-4      &     0.289$\pm$4e-3      &     0.478$\pm$2e-3      &     0.136$\pm$9e-4      &     0.228$\pm$1e-3      &     0.002$\pm$4e-3      &     0.611$\pm$1e-2      \\
			&  MulMON  & \textbf{0.557}$\pm$1e-3 & \textbf{0.579}$\pm$1e-3 & \textbf{0.916}$\pm$4e-3 & \textbf{0.910}$\pm$2e-3 &           N/A           &           N/A           & \textbf{0.426}$\pm$6e-2 &           N/A           \\
			& Proposed &     0.365$\pm$3e-3      &     0.464$\pm$2e-3      &     0.856$\pm$4e-3      &     0.864$\pm$2e-3      & \textbf{0.365}$\pm$2e-3 & \textbf{0.508}$\pm$2e-3 &     0.196$\pm$3e-2      & \textbf{0.922}$\pm$7e-3 \\ \midrule
			\multirow{3}{*}{CLEVR-M2} & Baseline &     0.332$\pm$1e-3      &     0.324$\pm$1e-3      &     0.294$\pm$4e-3      &     0.482$\pm$2e-3      &     0.140$\pm$1e-3      &     0.235$\pm$2e-3      &     0.002$\pm$4e-3      &     0.602$\pm$2e-2      \\
			&  MulMON  & \textbf{0.529}$\pm$8e-4 & \textbf{0.570}$\pm$9e-4 & \textbf{0.903}$\pm$1e-3 & \textbf{0.903}$\pm$1e-3 &           N/A           &           N/A           & \textbf{0.546}$\pm$5e-2 &           N/A           \\
			& Proposed &     0.355$\pm$6e-4      &     0.456$\pm$6e-4      &     0.817$\pm$5e-3      &     0.838$\pm$1e-3      & \textbf{0.356}$\pm$2e-3 & \textbf{0.499}$\pm$3e-3 &     0.214$\pm$2e-2      & \textbf{0.903}$\pm$9e-3 \\ \midrule
			\multirow{3}{*}{CLEVR-M3} & Baseline &     0.300$\pm$2e-3      &     0.307$\pm$2e-3      &     0.291$\pm$7e-3      &     0.478$\pm$4e-3      &     0.134$\pm$7e-4      &     0.225$\pm$9e-4      &     0.002$\pm$4e-3      &     0.625$\pm$2e-2      \\
			&  MulMON  & \textbf{0.531}$\pm$5e-3 & \textbf{0.566}$\pm$2e-3 & \textbf{0.899}$\pm$5e-3 & \textbf{0.897}$\pm$3e-3 &           N/A           &           N/A           & \textbf{0.426}$\pm$6e-2 &           N/A           \\
			& Proposed &     0.352$\pm$2e-3      &     0.452$\pm$1e-3      &     0.839$\pm$3e-3      &     0.852$\pm$2e-3      & \textbf{0.355}$\pm$4e-3 & \textbf{0.498}$\pm$5e-3 &     0.182$\pm$5e-2      & \textbf{0.897}$\pm$4e-3 \\ \midrule
			\multirow{3}{*}{CLEVR-M4} & Baseline &     0.325$\pm$1e-3      &     0.320$\pm$2e-3      &     0.294$\pm$5e-3      &     0.484$\pm$4e-3      &     0.143$\pm$8e-4      &     0.239$\pm$1e-3      &     0.000$\pm$0e-0      &     0.632$\pm$1e-2      \\
			&  MulMON  & \textbf{0.552}$\pm$9e-4 & \textbf{0.581}$\pm$6e-4 & \textbf{0.902}$\pm$1e-3 & \textbf{0.902}$\pm$5e-4 &           N/A           &           N/A           & \textbf{0.462}$\pm$4e-2 &           N/A           \\
			& Proposed &     0.284$\pm$2e-3      &     0.418$\pm$2e-3      &     0.789$\pm$5e-3      &     0.827$\pm$3e-3      & \textbf{0.314}$\pm$2e-3 & \textbf{0.453}$\pm$3e-3 &     0.180$\pm$2e-2      & \textbf{0.875}$\pm$2e-2 \\ \bottomrule
		\end{tabular}
	\end{small}
	\caption{Comparison of scene decomposition performance when learning from multiple viewpoints. All the methods are trained with $M \!=\! 4$ and $K \!=\! 7$, and tested on the Test 2 splits with $M \!=\! 4$ and $K \!=\! 11$.}
	\label{tab-supp:multi_4_2}
\end{table*}

\begin{table*}[t]
	\centering
	\begin{small}
		\addtolength{\tabcolsep}{-2.8pt}
		\begin{tabular}{c|c|C{0.62in}C{0.62in}C{0.62in}C{0.62in}C{0.62in}C{0.62in}C{0.62in}C{0.62in}}
			\toprule
			Dataset          &  Method  &          ARI-A          &          AMI-A          &          ARI-O          &          AMI-O          &           IoU           &           F1            &           OCA           &           OOA           \\ \midrule
			\multirow{3}{*}{CLEVR-M1} & Baseline &     0.511$\pm$7e-4      &     0.304$\pm$1e-3      &     0.140$\pm$2e-3      &     0.230$\pm$2e-3      &     0.130$\pm$2e-3      &     0.221$\pm$2e-3      &     0.000$\pm$0e-0      &     0.579$\pm$2e-2      \\
			&  MulMON  & \textbf{0.611}$\pm$3e-3 & \textbf{0.557}$\pm$2e-3 &     0.928$\pm$2e-3      &     0.916$\pm$2e-3      &           N/A           &           N/A           &     0.386$\pm$4e-2      &           N/A           \\
			& Proposed &     0.512$\pm$2e-3      &     0.489$\pm$1e-3      & \textbf{0.952}$\pm$2e-3 & \textbf{0.935}$\pm$2e-3 & \textbf{0.449}$\pm$2e-3 & \textbf{0.611}$\pm$3e-3 & \textbf{0.728}$\pm$4e-2 & \textbf{0.975}$\pm$4e-3 \\ \midrule
			\multirow{3}{*}{CLEVR-M2} & Baseline &     0.505$\pm$6e-4      &     0.301$\pm$2e-3      &     0.146$\pm$3e-3      &     0.237$\pm$5e-3      &     0.129$\pm$2e-3      &     0.221$\pm$3e-3      &     0.000$\pm$0e-0      &     0.587$\pm$9e-3      \\
			&  MulMON  & \textbf{0.607}$\pm$5e-4 & \textbf{0.552}$\pm$4e-4 &     0.937$\pm$3e-3      &     0.920$\pm$2e-3      &           N/A           &           N/A           &     0.578$\pm$5e-2      &           N/A           \\
			& Proposed &     0.509$\pm$1e-3      &     0.480$\pm$1e-3      & \textbf{0.946}$\pm$3e-3 & \textbf{0.936}$\pm$3e-3 & \textbf{0.437}$\pm$2e-3 & \textbf{0.598}$\pm$2e-3 & \textbf{0.718}$\pm$4e-2 & \textbf{0.945}$\pm$5e-3 \\ \midrule
			\multirow{3}{*}{CLEVR-M3} & Baseline &     0.530$\pm$3e-4      &     0.314$\pm$1e-3      &     0.142$\pm$3e-3      &     0.231$\pm$4e-3      &     0.131$\pm$1e-3      &     0.225$\pm$2e-3      &     0.000$\pm$0e-0      &     0.589$\pm$4e-2      \\
			&  MulMON  & \textbf{0.590}$\pm$9e-3 & \textbf{0.549}$\pm$3e-3 &     0.937$\pm$3e-3      &     0.922$\pm$3e-3      &           N/A           &           N/A           &     0.410$\pm$6e-2      &           N/A           \\
			& Proposed &     0.533$\pm$3e-4      &     0.497$\pm$7e-4      & \textbf{0.942}$\pm$5e-3 & \textbf{0.928}$\pm$2e-3 & \textbf{0.453}$\pm$3e-3 & \textbf{0.611}$\pm$4e-3 & \textbf{0.696}$\pm$4e-2 & \textbf{0.959}$\pm$2e-2 \\ \midrule
			\multirow{3}{*}{CLEVR-M4} & Baseline &     0.520$\pm$4e-4      &     0.310$\pm$1e-3      &     0.146$\pm$5e-3      &     0.237$\pm$5e-3      &     0.130$\pm$2e-3      &     0.222$\pm$2e-3      &     0.000$\pm$0e-0      &     0.600$\pm$1e-2      \\
			&  MulMON  & \textbf{0.644}$\pm$8e-4 & \textbf{0.580}$\pm$8e-4 & \textbf{0.936}$\pm$3e-3 &     0.922$\pm$1e-3      &           N/A           &           N/A           &     0.498$\pm$5e-2      &           N/A           \\
			& Proposed &     0.479$\pm$8e-4      &     0.456$\pm$2e-3      &     0.930$\pm$8e-3      & \textbf{0.924}$\pm$5e-3 & \textbf{0.407}$\pm$3e-3 & \textbf{0.567}$\pm$4e-3 & \textbf{0.616}$\pm$4e-2 & \textbf{0.880}$\pm$2e-2 \\ \bottomrule
		\end{tabular}
	\end{small}
	\caption{Comparison of scene decomposition performance when learning from multiple viewpoints. All the methods are trained with $M \!=\! 4$ and $K \!=\! 7$, and tested on the Test 1 splits with $M \!=\! 8$ and $K \!=\! 7$.}
	\label{tab-supp:multi_8_1}
\end{table*}

\begin{table*}[t]
	\centering
	\begin{small}
		\addtolength{\tabcolsep}{-2.8pt}
		\begin{tabular}{c|c|C{0.62in}C{0.62in}C{0.62in}C{0.62in}C{0.62in}C{0.62in}C{0.62in}C{0.62in}}
			\toprule
			Dataset          &  Method  &          ARI-A          &          AMI-A          &          ARI-O          &          AMI-O          &           IoU           &           F1            &           OCA           &           OOA           \\ \midrule
			\multirow{3}{*}{CLEVR-M1} & Baseline &     0.329$\pm$1e-3      &     0.245$\pm$2e-3      &     0.153$\pm$4e-3      &     0.294$\pm$5e-3      &     0.103$\pm$2e-3      &     0.180$\pm$2e-3      &     0.000$\pm$0e-0      &     0.570$\pm$9e-3      \\
			&  MulMON  & \textbf{0.552}$\pm$8e-4 & \textbf{0.575}$\pm$1e-3 & \textbf{0.913}$\pm$1e-3 & \textbf{0.906}$\pm$1e-3 &           N/A           &           N/A           & \textbf{0.442}$\pm$4e-2 &           N/A           \\
			& Proposed &     0.366$\pm$8e-4      &     0.463$\pm$2e-3      &     0.855$\pm$6e-3      &     0.859$\pm$3e-3      & \textbf{0.367}$\pm$2e-3 & \textbf{0.513}$\pm$3e-3 &     0.190$\pm$4e-2      & \textbf{0.923}$\pm$4e-3 \\ \midrule
			\multirow{3}{*}{CLEVR-M2} & Baseline &     0.327$\pm$6e-4      &     0.248$\pm$1e-3      &     0.155$\pm$4e-3      &     0.297$\pm$3e-3      &     0.105$\pm$1e-3      &     0.184$\pm$2e-3      &     0.000$\pm$0e-0      &     0.559$\pm$1e-2      \\
			&  MulMON  & \textbf{0.534}$\pm$6e-4 & \textbf{0.570}$\pm$8e-4 & \textbf{0.900}$\pm$2e-3 & \textbf{0.895}$\pm$1e-3 &           N/A           &           N/A           & \textbf{0.538}$\pm$6e-2 &           N/A           \\
			& Proposed &     0.353$\pm$6e-4      &     0.450$\pm$6e-4      &     0.810$\pm$3e-3      &     0.828$\pm$2e-3      & \textbf{0.354}$\pm$2e-3 & \textbf{0.498}$\pm$2e-3 &     0.170$\pm$2e-2      & \textbf{0.880}$\pm$1e-2 \\ \midrule
			\multirow{3}{*}{CLEVR-M3} & Baseline &     0.301$\pm$7e-4      &     0.236$\pm$1e-3      &     0.156$\pm$4e-3      &     0.295$\pm$4e-3      &     0.101$\pm$7e-4      &     0.177$\pm$1e-3      &     0.000$\pm$0e-0      &     0.591$\pm$2e-2      \\
			&  MulMON  & \textbf{0.531}$\pm$2e-3 & \textbf{0.566}$\pm$9e-4 & \textbf{0.906}$\pm$2e-3 & \textbf{0.899}$\pm$8e-4 &           N/A           &           N/A           & \textbf{0.394}$\pm$7e-2 &           N/A           \\
			& Proposed &     0.357$\pm$4e-3      &     0.452$\pm$2e-3      &     0.838$\pm$5e-3      &     0.847$\pm$3e-3      & \textbf{0.360}$\pm$3e-3 & \textbf{0.504}$\pm$4e-3 &     0.178$\pm$3e-2      & \textbf{0.912}$\pm$1e-2 \\ \midrule
			\multirow{3}{*}{CLEVR-M4} & Baseline &     0.322$\pm$7e-4      &     0.245$\pm$2e-3      &     0.156$\pm$5e-3      &     0.299$\pm$5e-3      &     0.106$\pm$2e-3      &     0.186$\pm$2e-3      &     0.000$\pm$0e-0      &     0.592$\pm$2e-2      \\
			&  MulMON  & \textbf{0.556}$\pm$5e-4 & \textbf{0.579}$\pm$5e-4 & \textbf{0.892}$\pm$1e-3 & \textbf{0.893}$\pm$8e-4 &           N/A           &           N/A           & \textbf{0.508}$\pm$3e-2 &           N/A           \\
			& Proposed &     0.278$\pm$1e-3      &     0.409$\pm$2e-3      &     0.783$\pm$6e-3      &     0.813$\pm$4e-3      & \textbf{0.310}$\pm$3e-3 & \textbf{0.451}$\pm$4e-3 &     0.134$\pm$3e-2      & \textbf{0.873}$\pm$1e-2 \\ \bottomrule
		\end{tabular}
	\end{small}
	\caption{Comparison of scene decomposition performance when learning from multiple viewpoints. All the methods are trained with $M \!=\! 4$ and $K \!=\! 7$, and tested on the Test 2 splits with $M \!=\! 8$ and $K \!=\! 11$.}
	\label{tab-supp:multi_8_2}
\end{table*}

\begin{table*}[t]
	\centering
	\begin{small}
		\addtolength{\tabcolsep}{-1.8pt}
		\begin{tabular}{c|c|C{0.62in}C{0.62in}C{0.62in}C{0.62in}C{0.62in}C{0.62in}C{0.62in}C{0.62in}}
			\toprule
			Dataset          & $M_1$ &          ARI-A          &          AMI-A          &          ARI-O          &          AMI-O          &           IoU           &           F1            &           OCA           &           OOA           \\ \midrule
			\multirow{3}{*}{CLEVR-M1} &   1   &     0.507$\pm$2e-3      &     0.475$\pm$2e-3      &     0.916$\pm$4e-3      &     0.905$\pm$3e-3      &     0.426$\pm$1e-3      &     0.581$\pm$1e-3      &     0.540$\pm$2e-2      &     0.935$\pm$8e-3      \\
			&   2   &     0.515$\pm$1e-3      &     0.487$\pm$1e-3      &     0.939$\pm$4e-3      &     0.925$\pm$2e-3      &     0.445$\pm$2e-3      &     0.603$\pm$3e-3      &     0.676$\pm$6e-2      & \textbf{0.965}$\pm$8e-3 \\
			&   4   & \textbf{0.518}$\pm$2e-3 & \textbf{0.491}$\pm$1e-3 & \textbf{0.943}$\pm$4e-3 & \textbf{0.930}$\pm$3e-3 & \textbf{0.451}$\pm$1e-3 & \textbf{0.611}$\pm$2e-3 & \textbf{0.692}$\pm$4e-2 & \textbf{0.965}$\pm$1e-2 \\ \midrule
			\multirow{3}{*}{CLEVR-M2} &   1   &     0.492$\pm$1e-3      &     0.454$\pm$2e-3      &     0.889$\pm$6e-3      &     0.886$\pm$5e-3      &     0.400$\pm$1e-3      &     0.550$\pm$2e-3      &     0.554$\pm$3e-2      &     0.903$\pm$2e-2      \\
			&   2   &     0.502$\pm$1e-3      &     0.469$\pm$2e-3      &     0.925$\pm$6e-3      &     0.917$\pm$5e-3      &     0.423$\pm$1e-3      &     0.578$\pm$2e-3      &     0.626$\pm$4e-2      &     0.937$\pm$1e-2      \\
			&   4   & \textbf{0.506}$\pm$1e-3 & \textbf{0.475}$\pm$2e-3 & \textbf{0.938}$\pm$3e-3 & \textbf{0.929}$\pm$3e-3 & \textbf{0.430}$\pm$5e-3 & \textbf{0.589}$\pm$6e-3 & \textbf{0.676}$\pm$5e-2 & \textbf{0.953}$\pm$1e-2 \\ \midrule
			\multirow{3}{*}{CLEVR-M3} &   1   &     0.507$\pm$2e-3      &     0.462$\pm$3e-3      &     0.890$\pm$3e-3      &     0.880$\pm$3e-3      &     0.421$\pm$4e-3      &     0.573$\pm$5e-3      &     0.576$\pm$3e-2      &     0.944$\pm$1e-2      \\
			&   2   &     0.517$\pm$1e-3      &     0.478$\pm$2e-3      &     0.914$\pm$5e-3      &     0.903$\pm$4e-3      &     0.436$\pm$3e-3      &     0.591$\pm$4e-3      &     0.602$\pm$3e-2      &     0.942$\pm$1e-2      \\
			&   4   & \textbf{0.523}$\pm$8e-4 & \textbf{0.487}$\pm$7e-4 & \textbf{0.926}$\pm$3e-3 & \textbf{0.916}$\pm$2e-3 & \textbf{0.449}$\pm$2e-3 & \textbf{0.607}$\pm$2e-3 & \textbf{0.630}$\pm$1e-2 & \textbf{0.954}$\pm$1e-2 \\ \midrule
			\multirow{3}{*}{CLEVR-M4} &   1   &     0.448$\pm$2e-3      &     0.412$\pm$2e-3      &     0.856$\pm$1e-2      &     0.856$\pm$7e-3      &     0.360$\pm$3e-3      &     0.506$\pm$4e-3      &     0.498$\pm$2e-2      &     0.796$\pm$2e-2      \\
			&   2   &     0.463$\pm$2e-3      &     0.433$\pm$1e-3      &     0.889$\pm$4e-3      &     0.888$\pm$3e-3      &     0.383$\pm$1e-3      &     0.536$\pm$2e-3      &     0.562$\pm$2e-2      &     0.849$\pm$2e-2      \\
			&   4   & \textbf{0.472}$\pm$9e-4 & \textbf{0.446}$\pm$1e-3 & \textbf{0.919}$\pm$4e-3 & \textbf{0.912}$\pm$3e-3 & \textbf{0.400}$\pm$4e-3 & \textbf{0.558}$\pm$5e-3 & \textbf{0.596}$\pm$3e-2 & \textbf{0.889}$\pm$2e-2 \\ \bottomrule
		\end{tabular}
	\end{small}
	\caption{Results of scene decomposition performance when estimating viewpoints given $\boldsymbol{y}_{0:K}^{\text{attr}}$. All the models are trained with $M \!=\! 4$ and $K \!=\! 7$, and tested on the Test 1 splits with $K \!=\! 7$. The models first estimate $\boldsymbol{y}_{0:K}^{\text{attr}}$ based on $M_1 = 1, 2, 4$ viewpoints, and then estimate viewpoint latent variables of $M_2 = 4$ novel viewpoints.}
	\label{tab-supp:cond_1}
\end{table*}

\begin{table*}[t]
	\centering
	\begin{small}
		\addtolength{\tabcolsep}{-1.8pt}
		\begin{tabular}{c|c|C{0.62in}C{0.62in}C{0.62in}C{0.62in}C{0.62in}C{0.62in}C{0.62in}C{0.62in}}
			\toprule
			Dataset          & $M_1$ &          ARI-A          &          AMI-A          &          ARI-O          &          AMI-O          &           IoU           &           F1            &           OCA           &           OOA           \\ \midrule
			\multirow{3}{*}{CLEVR-M1} &   1   &     0.346$\pm$2e-3      &     0.437$\pm$1e-3      &     0.800$\pm$4e-3      &     0.821$\pm$2e-3      &     0.338$\pm$1e-3      &     0.477$\pm$1e-3      & \textbf{0.266}$\pm$3e-2 &     0.886$\pm$4e-3      \\
			&   2   &     0.350$\pm$4e-3      &     0.447$\pm$3e-3      &     0.822$\pm$4e-3      &     0.838$\pm$3e-3      &     0.351$\pm$3e-3      &     0.493$\pm$4e-3      &     0.232$\pm$2e-2      &     0.910$\pm$5e-3      \\
			&   4   & \textbf{0.358}$\pm$2e-3 & \textbf{0.457}$\pm$9e-4 & \textbf{0.845}$\pm$2e-3 & \textbf{0.854}$\pm$1e-3 & \textbf{0.365}$\pm$1e-3 & \textbf{0.510}$\pm$2e-3 &     0.200$\pm$4e-2      & \textbf{0.919}$\pm$9e-3 \\ \midrule
			\multirow{3}{*}{CLEVR-M2} &   1   &     0.333$\pm$2e-3      &     0.423$\pm$3e-3      &     0.751$\pm$9e-3      &     0.790$\pm$5e-3      &     0.327$\pm$2e-3      &     0.463$\pm$4e-3      & \textbf{0.256}$\pm$3e-2 &     0.873$\pm$1e-2      \\
			&   2   &     0.336$\pm$2e-3      &     0.434$\pm$1e-3      &     0.778$\pm$4e-3      &     0.810$\pm$1e-3      &     0.339$\pm$3e-3      &     0.478$\pm$5e-3      &     0.210$\pm$2e-2      &     0.871$\pm$6e-3      \\
			&   4   & \textbf{0.344}$\pm$2e-3 & \textbf{0.445}$\pm$2e-3 & \textbf{0.799}$\pm$5e-3 & \textbf{0.827}$\pm$3e-3 & \textbf{0.350}$\pm$2e-3 & \textbf{0.493}$\pm$2e-3 &     0.192$\pm$3e-2      & \textbf{0.884}$\pm$1e-2 \\ \midrule
			\multirow{3}{*}{CLEVR-M3} &   1   &     0.343$\pm$2e-3      &     0.429$\pm$2e-3      &     0.787$\pm$9e-3      &     0.812$\pm$5e-3      &     0.329$\pm$3e-3      &     0.464$\pm$4e-3      & \textbf{0.270}$\pm$4e-2 &     0.897$\pm$5e-3      \\
			&   2   & \textbf{0.350}$\pm$2e-3 &     0.440$\pm$2e-3      &     0.810$\pm$1e-2      &     0.829$\pm$5e-3      &     0.344$\pm$4e-3      &     0.483$\pm$5e-3      &     0.204$\pm$3e-2      & \textbf{0.903}$\pm$2e-2 \\
			&   4   &     0.349$\pm$2e-3      & \textbf{0.445}$\pm$2e-3 & \textbf{0.820}$\pm$4e-3 & \textbf{0.840}$\pm$2e-3 & \textbf{0.352}$\pm$2e-3 & \textbf{0.494}$\pm$2e-3 &     0.152$\pm$4e-2      &     0.902$\pm$1e-2      \\ \midrule
			\multirow{3}{*}{CLEVR-M4} &   1   &     0.256$\pm$4e-3      &     0.376$\pm$2e-3      &     0.705$\pm$3e-3      &     0.762$\pm$1e-3      &     0.272$\pm$2e-3      &     0.398$\pm$2e-3      & \textbf{0.248}$\pm$4e-2 &     0.838$\pm$2e-2      \\
			&   2   &     0.265$\pm$1e-3      &     0.394$\pm$2e-3      &     0.746$\pm$4e-3      &     0.794$\pm$4e-3      &     0.291$\pm$3e-3      &     0.425$\pm$3e-3      &     0.222$\pm$4e-2      &     0.843$\pm$2e-2      \\
			&   4   & \textbf{0.268}$\pm$2e-3 & \textbf{0.402}$\pm$2e-3 & \textbf{0.765}$\pm$9e-3 & \textbf{0.808}$\pm$5e-3 & \textbf{0.302}$\pm$3e-3 & \textbf{0.440}$\pm$4e-3 &     0.178$\pm$3e-2      & \textbf{0.859}$\pm$1e-2 \\ \bottomrule
		\end{tabular}
	\end{small}
	\caption{Results of scene decomposition performance when estimating viewpoints given $\boldsymbol{y}_{0:K}^{\text{attr}}$. All the models are trained with $M \!=\! 4$ and $K \!=\! 7$, and tested on the Test 2 splits with $K \!=\! 11$. The models first estimate $\boldsymbol{y}_{0:K}^{\text{attr}}$ based on $M_1 = 1, 2, 4$ viewpoints, and then estimate viewpoint latent variables of $M_2 = 4$ novel viewpoints.}
	\label{tab-supp:cond_2}
\end{table*}

\begin{table*}[t]
	\centering
	\begin{small}
		\addtolength{\tabcolsep}{-2.2pt}
		\begin{tabular}{c|c|C{0.62in}C{0.62in}C{0.62in}C{0.62in}C{0.62in}C{0.62in}C{0.62in}C{0.62in}}
			\toprule
			Dataset          &  Method   &          ARI-A          &          AMI-A          &          ARI-O          &          AMI-O          &           IoU           &           F1            &           OCA           &           OOA           \\ \midrule
			\multirow{4}{*}{dSprites} & Slot Attn &     0.143$\pm$1e-3      &     0.348$\pm$8e-4      & \textit{0.862}$\pm$1e-3 & \textit{0.869}$\pm$7e-4 &           N/A           &           N/A           &     0.000$\pm$0e-0      &           N/A           \\
			&   GMIOO   & \textbf{0.958}$\pm$4e-4 & \textbf{0.902}$\pm$7e-4 & \textbf{0.920}$\pm$1e-3 & \textbf{0.927}$\pm$9e-4 & \textbf{0.780}$\pm$2e-3 & \textbf{0.843}$\pm$2e-3 & \textbf{0.558}$\pm$8e-3 & \textbf{0.871}$\pm$3e-3 \\
			&   SPACE   & \textit{0.914}$\pm$3e-4 & \textit{0.818}$\pm$1e-4 &     0.779$\pm$8e-4      &     0.828$\pm$3e-4      &     0.562$\pm$6e-4      &     0.653$\pm$8e-4      &     0.204$\pm$8e-3      &     0.654$\pm$9e-3      \\
			& Proposed  &     0.779$\pm$5e-4      &     0.692$\pm$4e-4      &     0.797$\pm$2e-3      &     0.824$\pm$1e-3      & \textit{0.615}$\pm$1e-3 & \textit{0.740}$\pm$1e-3 & \textit{0.330}$\pm$1e-2 & \textit{0.677}$\pm$6e-3 \\ \midrule
			\multirow{4}{*}{Abstract} & Slot Attn & \textbf{0.909}$\pm$1e-3 & \textbf{0.843}$\pm$1e-3 & \textit{0.904}$\pm$1e-3 &     0.881$\pm$8e-4      &           N/A           &           N/A           &     0.632$\pm$1e-2      &           N/A           \\
			&   GMIOO   &     0.816$\pm$3e-4      &     0.758$\pm$4e-4      & \textbf{0.905}$\pm$1e-3 & \textbf{0.902}$\pm$7e-4 & \textit{0.736}$\pm$1e-3 & \textit{0.829}$\pm$1e-3 & \textbf{0.796}$\pm$4e-3 & \textbf{0.936}$\pm$1e-3 \\
			&   SPACE   & \textit{0.849}$\pm$2e-4 &     0.758$\pm$3e-4      &     0.761$\pm$8e-4      &     0.799$\pm$7e-4      &     0.614$\pm$7e-4      &     0.693$\pm$8e-4      &     0.267$\pm$2e-3      &     0.789$\pm$4e-3      \\
			& Proposed  &     0.844$\pm$3e-4      & \textit{0.776}$\pm$2e-4 &     0.901$\pm$1e-3      & \textit{0.894}$\pm$8e-4 & \textbf{0.770}$\pm$9e-4 & \textbf{0.861}$\pm$1e-3 & \textit{0.664}$\pm$1e-2 & \textit{0.926}$\pm$2e-3 \\ \midrule
			\multirow{4}{*}{CLEVR}   & Slot Attn &     0.078$\pm$8e-5      &     0.378$\pm$2e-4      & \textbf{0.939}$\pm$8e-4 & \textbf{0.945}$\pm$5e-4 &           N/A           &           N/A           &     0.011$\pm$2e-3      &           N/A           \\
			&   GMIOO   & \textit{0.666}$\pm$3e-4 & \textit{0.670}$\pm$4e-4 &     0.884$\pm$1e-3      &     0.923$\pm$8e-4      & \textit{0.515}$\pm$1e-3 & \textit{0.624}$\pm$1e-3 &     0.160$\pm$5e-3      &     0.855$\pm$2e-3      \\
			&   SPACE   & \textbf{0.823}$\pm$3e-4 & \textbf{0.781}$\pm$2e-4 & \textit{0.932}$\pm$5e-4 & \textit{0.939}$\pm$4e-4 & \textbf{0.684}$\pm$4e-4 & \textbf{0.773}$\pm$5e-4 & \textbf{0.417}$\pm$3e-3 & \textit{0.895}$\pm$4e-3 \\
			& Proposed  &     0.478$\pm$8e-4      &     0.560$\pm$7e-4      &     0.916$\pm$1e-3      &     0.925$\pm$1e-3      &     0.473$\pm$2e-3      &     0.621$\pm$2e-3      & \textit{0.309}$\pm$1e-2 & \textbf{0.932}$\pm$3e-3 \\ \bottomrule
		\end{tabular}
	\end{small}
	\caption{Comparison of scene decomposition performance when learning from a single viewpoint. All the methods are trained with $K \!=\! 6$, $K \!=\! 5$, and $K \!=\! 7$, and tested on the Test 2 splits with $K \!=\! 9$, $K \!=\! 7$, and $K \!=\! 11$ for the dSprites, Abstract, and CLEVR datasets, respectively.}
	\label{tab-supp:single}
\end{table*}

\end{document}